\documentclass{article}
\usepackage{iclr2027_conference,times}

\usepackage{amsmath,amsfonts,bm}

\def\eqref#1{equation~\ref{#1}}
\def\Eqref#1{Equation~\ref{#1}}

\def\1{\bm{1}}

\DeclareMathAlphabet{\mathsfit}{\encodingdefault}{\sfdefault}{m}{sl}
\SetMathAlphabet{\mathsfit}{bold}{\encodingdefault}{\sfdefault}{bx}{n}

\newcommand{\Var}{\mathrm{Var}}

\usepackage{graphicx}
\usepackage{placeins}
\usepackage{booktabs}
\usepackage{multirow}
\usepackage{colortbl}
\usepackage{amssymb}
\usepackage{tcolorbox}
\usepackage{hyperref}
\usepackage{url}
\hypersetup{
  pdftitle={DisKO: Deep Koopman Learning in Distribution Space from Unpaired Snapshots},
  pdfauthor={He Ma, Xiaochen Liu, Wanfeng Lu, Ying Wang, Wei Lin, Qunxi Zhu},
  colorlinks=false,
  pdfborder={0 0 1},
  linkbordercolor={1 0 0},
  citebordercolor={0 1 0}
}

\newcounter{illustrativeexample}[section]
\renewcommand{\theillustrativeexample}{\thesection.\arabic{illustrativeexample}}
\newenvironment{illustrativeexample}[1]{%
  \refstepcounter{illustrativeexample}%
  \begin{tcolorbox}[colback=black!3,colframe=black!28,
    boxrule=0.4pt,arc=2mm,boxsep=0pt,
    left=8pt,right=8pt,top=8pt,bottom=8pt,
    before skip=10pt,after skip=10pt]%
  \textbf{Example~\theillustrativeexample\ (#1).}\enspace
}{\end{tcolorbox}}

\graphicspath{{figures/generated/}{figures/assets/}}
\title{DisKO: Deep Koopman Learning in\\
Distribution Space from Unpaired Snapshots}
\author{%
  \begin{minipage}[t]{\dimexpr\textwidth-2\tabcolsep\relax}
    \centering\normalfont
    \textbf{He Ma}\textsuperscript{1,2}\qquad
    \textbf{Xiaochen Liu}\textsuperscript{2,5}\qquad
    \textbf{Wanfeng Lu}\textsuperscript{1,2}
    \endgraf\vspace{4pt}
    \textbf{Ying Wang}\textsuperscript{2}\qquad
    \textbf{Wei Lin}\textsuperscript{1,2,3,4,5}\qquad
    \textbf{Qunxi Zhu}\textsuperscript{2,}\thanks{Correspondence to: Qunxi Zhu
      (\href{mailto:qxzhu@fudan.edu.cn}{\nolinkurl{qxzhu@fudan.edu.cn}}).}
    \endgraf\vspace{8pt}
    {\small
      \textsuperscript{1}School of Mathematical Sciences, Fudan University, China.\\[2pt]
      \textsuperscript{2}Research Institute of Intelligent Complex Systems, Fudan University, China.\\[2pt]
      \textsuperscript{3}Shanghai Artificial Intelligence Laboratory, China.\\[2pt]
      \textsuperscript{4}State Key Laboratory of Medical Neurobiology and MOE Frontiers Center for Brain Science,
        Institutes of Brain Science, Fudan University, China.\\[2pt]
      \textsuperscript{5}Shanghai Center for Mathematical Sciences, Fudan University, China.
      \endgraf
    }
  \end{minipage}%
}
\iclrfinalcopy

\begin{document}
\maketitle
\lhead{Preprint}

\begin{abstract}
Many complex systems are observed only through temporally unpaired distribution snapshots, making trajectory-based dynamical learning difficult without additional assumptions.
We therefore formulate the problem directly in distribution space, treating the distribution itself as the dynamical state.
The challenge is that distribution space is infinite-dimensional, making compact and approximately closed representations difficult to learn from finite snapshots.
We introduce \textbf{DisKO}, which extends deep Koopman learning to distribution dynamics by jointly learning predictive distributional observables, a finite-dimensional Koopman representation, and a generative map back to the full distribution.
Across seven diverse benchmarks, DisKO achieves \textbf{state-of-the-art} extrapolation performance, with \textbf{substantially slower error accumulation} on long-horizon prediction tasks.
DisKO further recovers leading Koopman eigenvalues and eigenfunctions on systems with analytic spectra, revealing meaningful dynamical structure in the learned representation.

\end{abstract}

\section{Introduction}
\label{sec:introduction}

\looseness=-1
Reconstructing the dynamics of complex systems from observations is a central problem in science \citep{chen2022automated}. Many established methods rely on temporally paired state observations to estimate state transitions or temporal derivatives \citep{brunton2016discovering,lusch2018deep}. In many settings, however, destructive measurements, acquisition costs, or the spatial and temporal scales of the process make it infeasible to track the same entities across time \citep{weinreb2020lineage,bunne2023learning}. Instead, observations consist of finite samples from the system's state distribution at a limited number of time points, without information linking individual observations across time. We refer to such observations as \emph{unpaired distribution snapshots}, or simply \emph{distribution snapshots}. Such data arise in diverse settings, including single-cell sequencing, particle ensembles in collective motion and transport systems, and halo populations in cosmological simulations \citep{schiebinger2019optimal,klein2025moscot,weiler2024cellrank,reynolds1987flocks,villaescusa2021camels}. The problem is therefore to learn dynamics directly from the temporal evolution of distributions.

\looseness=-1
When individual correspondences across time are unavailable, many methods infer latent trajectories between snapshots using optimal transport, stochastic dynamics, or path regularization \citep{schiebinger2019optimal,tong2020trajectorynet,zhang2025deepruot,sha2024reconstructing,zhang2025inferring}. Yet marginal distributions do not uniquely determine individual trajectories: different couplings or stochastic processes can produce the same snapshots \citep{gu2025partially,petrovic2025curly}. Trajectory reconstruction therefore requires additional assumptions to recover information that is not observed. If the goal is to predict population distributions, a more direct approach is to model dynamics in distribution space itself \citep{atanackovic2025meta,haviv2025wasserstein}.

\looseness=-1
This better matches the form of snapshot data, but distribution space is infinite-dimensional, and directly learning nonlinear dynamics there does not naturally yield a compact, approximately closed representation for long-horizon prediction. Koopman methods offer a different perspective by lifting nonlinear state dynamics to linear evolution of observables, enabling structured propagation and spectral analysis \citep{brunton2022modern,froyland2021spectral}. Deep Koopman learning further learns these observables from data rather than prescribing them \citep{li2017dictionary,lusch2018deep,mardt2018vampnets}.

\looseness=-1
For distributional dynamics, however, observables become functionals of entire distributions. The Distributional Koopman Operator extends Koopman theory to this setting, but existing data-driven methods estimate finite-dimensional operators over prescribed observable dictionaries \citep{oprea2025distributional}. They neither learn predictive distributional observables nor provide a generative map from those observables back to the full distribution.

\begin{figure}[!htbp]
    \centering
    \includegraphics[width=.85\linewidth]{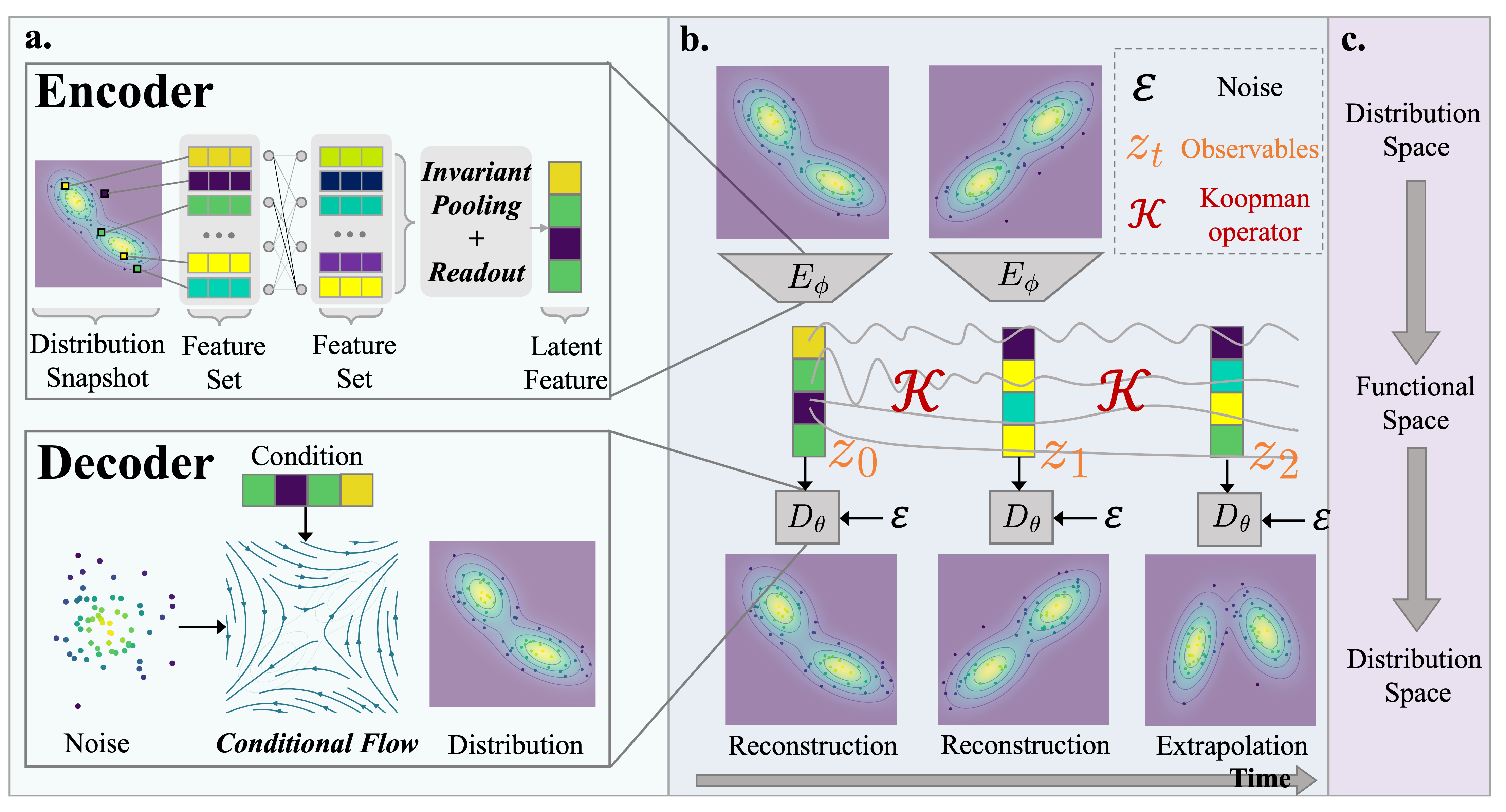}
    \caption{Overview of DisKO. Distribution snapshots are encoded into learned observables, evolved under Koopman dynamics, and decoded back to distributions for reconstruction and extrapolation.}
    \label{fig:framework}
\end{figure}

\looseness=-1
To close this gap, we introduce \textbf{DisKO}, a deep \textbf{Dis}tributional \textbf{KO}opman learning framework that extends deep Koopman learning to distribution-valued dynamics. Rather than prescribing distributional observables, DisKO jointly learns predictive observables, a finite-dimensional representation of the corresponding Koopman operator, and a generative map back to the full probability distribution. Concretely, a permutation-invariant encoder maps empirical distribution snapshots to finite-dimensional latent coordinates, a shared affine generator advances these coordinates in continuous time, and a conditional generative decoder maps encoded or propagated coordinates back to full distributions (Figure~\ref{fig:framework}). Table~\ref{tab:method-capabilities} compares method capabilities.

\looseness=-1
Our contributions are threefold: \textbf{(1)} We \textbf{extend deep Koopman learning to distribution dynamics} and introduce DisKO, an end-to-end framework that learns directly from unpaired distribution snapshots. \textbf{(2)} Across seven benchmarks spanning stochastic dynamics, collective motion, transport, cosmology, and single-cell dynamics, DisKO achieves \textbf{state-of-the-art} extrapolation while exhibiting \textbf{substantially slower error accumulation} over long prediction horizons. \textbf{(3)} On systems with analytic Koopman spectra, DisKO recovers leading eigenvalues and eigenfunctions, showing that its learned representation is not only predictive but also \textbf{captures interpretable decay and oscillatory modes}.

\begin{table}[htbp]
    \centering
    \scriptsize
    \setlength{\tabcolsep}{1.5pt}
    \renewcommand{\arraystretch}{0.9}
    \setlength{\aboverulesep}{0.5pt}
    \setlength{\belowrulesep}{0.7pt}
    \newcommand{\capabilityyes}{\textcolor{red}{$\checkmark$}}
    \newcommand{\capabilityno}{\textcolor{green!60!black}{$\times$}}
    \resizebox{\linewidth}{!}{%
    \begin{tabular}{@{}lccccc@{}}
        \toprule
        Method & Distribution snapshots
        & Learned observables
        & Distribution space
        & Full-distribution prediction
        & Spectral access \\
        \midrule
        Deep Koopman \citep{lusch2018deep}
        & \capabilityno & \capabilityyes & \capabilityno & \capabilityno & \capabilityyes \\
        \midrule[0.3pt]
        DeepRUOT \citep{zhang2025deepruot}
        & \capabilityyes & \capabilityno & \capabilityno & \capabilityyes & \capabilityno \\
        \midrule[0.3pt]
        WLM \citep{guan2026lagrangian}
        & \capabilityyes & \capabilityno & \capabilityyes & \capabilityyes & \capabilityno \\
        \midrule[0.3pt]
        \textbf{Distributional DMD} \citep{oprea2025distributional}
        & \capabilityyes & \capabilityno & \capabilityyes & \capabilityno & \capabilityyes \\
        \midrule[0.3pt]
        \textbf{DisKO}
        & \capabilityyes & \capabilityyes & \capabilityyes & \capabilityyes & \capabilityyes \\
        \bottomrule
    \end{tabular}%
    }
    \caption{Comparison of methods for learning distribution dynamics.}
    \label{tab:method-capabilities}
\end{table}

\section{Related Work}
\label{sec:related-work}

\noindent\textbf{Learning distribution dynamics from snapshots.}\enspace
\looseness=-1
Learning population dynamics from unpaired snapshots requires assumptions about the evolution between observations \citep{lamanno2018velocity,bergen2020generalizing,gayoso2024deep,rohbeck2025modeling,kviman2026multimarginal}.
TrajectoryNet constrains the paths between snapshots through transport regularization \citep{tong2020trajectorynet}, while DeepRUOT and VGFM couple transport with population growth \citep{zhang2025deepruot,wang2025vgfm}.
This evolution can also be modeled in learned latent coordinates, as in MIOFlow and scNODE \citep{huguet2022mioflow,zhang2024scnode}, or constrained by energy and mechanical principles \citep{persiianov2026learning}, as in PRESCIENT, JKOnet*, and WLM \citep{yeo2021prescient,terpin2024lightspeed,guan2026lagrangian}.
By contrast, DisKO brings a Koopman perspective to this problem, learning distribution observables whose evolution supports both prediction and spectral analysis.

\noindent\textbf{Koopman theory and learning.}\enspace
\looseness=-1
Koopman methods obtain such spectral descriptions by approximating the linear evolution of observables \citep{ryu2024operator,jeong2025efficient,colbrook2026adversarial}.
Dynamic mode decomposition and its extended form estimate this evolution from paired state observations using a prescribed representation \citep{schmid2010dynamic,williams2015edmd}.
Dictionary learning and neural embeddings make that representation trainable \citep{li2017dictionary,lusch2018deep,naiman2024generative,xu2025reskoopnet,selim2025metakoopman}.
For population prediction, spectral expansions have been used to forecast probability densities \citep{wang2026functional}, and distributional Koopman theory extends the observables themselves to functionals of distributions \citep{oprea2025distributional}.
The latter formulation provides the basis for DisKO, which learns these functionals jointly with a continuous generator and a generative decoder.

\noindent\textbf{Set representations and conditional generation.}\enspace
\looseness=-1
Learning distribution observables from finite snapshots requires representations that respect sample permutations \citep{fishman2025generative}.
Deep Sets and Set Transformer provide such encoders \citep{zaheer2017deep,lee2019set}.
To recover a distribution from its evolved representation, we pair these encoders with conditional generative flows trained by flow matching \citep{lipman2023flow,tong2024improving}.

\section{Theoretical Framework}
\label{sec:theory}

We introduce the snapshot prediction problem and Koopman representations of distribution dynamics, with details in Appendix~\ref{app:revised-theory}.

\subsection{Problem Formulation}
\label{sec:problem}

\looseness=-1
We learn from \emph{distribution snapshots}, each consisting of an unordered set of state samples collected at a given time.
Such data may be generated by deterministic or stochastic dynamics \citep{qiu2022mapping,vinyard2025learning}.
A representative model is
\begin{equation}
    dX_t=b(X_t)\,dt+\sigma(X_t)\,dW_t,
    \qquad X_0\sim\nu_0,
    \label{eq:state-dynamics}
\end{equation}
where $X_t\in\mathcal X\subseteq\mathbb R^p$ and $W_t$ is a standard Brownian motion independent of $X_0$.
We assume a unique nonexplosive solution and write $\nu_t=\operatorname{Law}(X_t)$ for its distribution.
Setting $\sigma=0$ recovers an ODE.

\looseness=-1
Sequences of distribution snapshots, indexed by $r$, share the dynamics and may have different initial distributions $\nu_0^{(r)}$.
At time $t_k$, sequence $r$ provides an unordered sample $\{y_{k,i}^{(r)}\}_{i=1}^{N_{r,k}}$ from $\nu_{t_k}^{(r)}$, represented by
\begingroup
\setlength{\abovedisplayskip}{3pt plus1pt minus1pt}
\setlength{\belowdisplayskip}{3pt plus1pt minus1pt}
\setlength{\abovedisplayshortskip}{0pt}
\setlength{\belowdisplayshortskip}{3pt plus1pt minus1pt}
\begin{equation}
    \widehat\nu_{t_k}^{(r)}
    =\frac{1}{N_{r,k}}\sum\nolimits_{i=1}^{N_{r,k}}\delta_{y_{k,i}^{(r)}}.
    \label{eq:empirical-snapshot}
\end{equation}
\endgroup
Observation times and sequence membership are known, while individual samples are unpaired across time.

\looseness=-1
Our goal is to extrapolate distribution evolution beyond the training interval $[0,\tau]$.
After training on snapshots within this interval, the model receives a source snapshot $\widehat\nu_s^{(r)}$ at $s\leq\tau$ and predicts $\nu_t^{(r)}$ for $t>\tau$ without further observations.
Our experiments use the first snapshot as input, with future snapshots providing ground truth distributions for evaluation (Appendix~\ref{app:prediction-protocol}).

\subsection{The Distributional Koopman Operator}
\label{sec:distributional-koopman}

\looseness=-1
Let $\mathfrak M\subseteq\mathcal P(\mathcal X)$ be a family of distributions equipped with a deterministic evolution $T_t:\mathfrak M\to\mathfrak M$ satisfying
$T_0=\operatorname{Id}, \; T_{t+s}=T_t\circ T_s, \; t,s\geq0.$
This formulation describes evolution on the space of distributions independently of a particular state equation.
Well-posed autonomous ODEs and time-homogeneous Markov SDEs induce such evolutions through deterministic pushforward and Markov transition probabilities, respectively.

\looseness=-1
Let $\mathcal H$ be a linear space of real-valued distribution observables $h:\mathfrak M\to\mathbb R$ that is preserved under composition with $T_t$.
Following \citet{oprea2025distributional}, the distributional Koopman operator is
\begin{equation}
    (\mathcal K_t h)(\mu)=h(T_t\mu).
    \label{eq:distributional-koopman}
\end{equation}
We use $\mathcal K_t$ for this operator and reserve $D$ for the decoder.
The input to $h$ is a distribution, and its output is a scalar statistic of that distribution.
This extends the usual composition of a state observable with a state evolution map to the space of distributions.
Composition is linear in the observable, so
$\mathcal K_t(\alpha h+\beta g)=\alpha\mathcal K_t h+\beta\mathcal K_t g, \; \mathcal K_{t+s}=\mathcal K_t\mathcal K_s.$
On a function space where this semigroup is strongly continuous, its generator $\mathcal L$ is defined for observables in its domain by
\begin{equation}
    \mathcal Lh=\lim_{\Delta t\downarrow0}
    \frac{\mathcal K_{\Delta t}h-h}{\Delta t}.
    \label{eq:distributional-generator}
\end{equation}
The generator describes the instantaneous evolution of distribution observables and provides the continuous-time object approximated below.

\subsection{Learning Finite-Dimensional Distributional Representations}
\label{sec:learned-distribution-observables}

\looseness=-1
To model distribution evolution in finite dimensions, we jointly learn observables, their dynamics, and a decoder.
The components of an encoder $E:\mathcal P(\mathcal X)\to\mathbb R^m$ define the distribution observables,
$h_j(\mu)=E_j(\mu), \; z=E(\nu)=\mathbf h(\nu)\in\mathbb R^m, \; \mathbf h=(h_1,\ldots,h_m)^\top.$
The $h_j$ are potentially nonlinear functionals on $\mathfrak M$, and $z$ collects their values at $\nu$.
In practice, the encoder is evaluated on empirical snapshots.
To propagate these coordinates, we learn an affine system
\begin{equation}
    \dot z=Az+c,
    \qquad A\in\mathbb R^{m\times m},\quad c\in\mathbb R^m,
    \label{eq:latent-generator}
\end{equation}
with flow $F_t$.
For observables in the domain of $\mathcal L$, we seek the generator and flow approximations
\begin{equation}
    \mathcal L\mathbf h\approx A\mathbf h+c,
    \qquad
    \mathbf h(T_t\mu)\approx F_t(\mathbf h(\mu)).
    \label{eq:observable-propagation}
\end{equation}
Adding the constant observable makes the model linear in $(z,1)$, approximating the action of $\mathcal L$ on the span of $1,h_1,\ldots,h_m$.
Exact propagation requires this space to be invariant under Koopman evolution \citep{brunton2016invariant,cenedese2022datadriven,fathi2024course,lu2026interpretable}.
In practice, we seek approximate closure and assess it through future predictions and spectral diagnostics. Appendix~\ref{app:revised-observables} provides further analysis of observable expressiveness, dynamical closure, and spectral interpretation.

\looseness=-1
To recover distributions from these coordinates, we learn a decoder $D:\mathbb R^m\to\mathcal P(\mathcal X)$ satisfying $D(E(\nu))\approx\nu$.
Reconstruction encourages the encoder to retain distributional information.
Exact recovery on a modeled family requires $E$ to distinguish its members and $D$ to invert $E$ on its image.
For $t>s$, we predict $\widetilde\nu_{t\mid s}^{(r)}=D(F_{t-s}(E(\widehat\nu_s^{(r)})))$ by encoding, propagating, and decoding the source snapshot.
In practice, $D$ generates samples conditioned on the propagated coordinates.
Section~\ref{sec:method} specifies the parameterizations and joint training objectives. The following OU example illustrates exact finite-dimensional propagation and distribution reconstruction.

\begin{illustrativeexample}{Distribution observables for an OU system}
\label{ex:ou-distribution-observables}
Consider $dX_t=-aX_t\,dt+\sigma\,dW_t$ with $a>0$, $\sigma\geq0$, and $X_0\sim\mathcal N(m_0,v_0)$ independent of the Brownian motion $W_t$.
Its distribution remains Gaussian and is determined by the values of the mean and variance observables
$h_1(\nu)=\int x\,d\nu(x)$, $h_2(\nu)=\int x^2\,d\nu(x)-h_1(\nu)^2$.
Their Koopman evolution is explicit, with $v_\infty=\sigma^2/(2a)$,
$(\mathcal K_t h_1)(\nu_0)=h_1(\nu_t)=m_t=e^{-at}m_0$, $(\mathcal K_t h_2)(\nu_0)=h_2(\nu_t)=v_t=e^{-2at}v_0+v_\infty(1-e^{-2at})$.
The encoder $E(\nu)=(h_1(\nu),h_2(\nu))^\top$ therefore gives coordinates $z_t=(m_t,v_t)^\top$ with exact affine dynamics
$\dot z_t=(-am_t,-2av_t+\sigma^2)^\top$.
The span of $1,h_1,h_2$ is invariant under $\mathcal K_t$.
Within the Gaussian family, $D(m,v)=\mathcal N(m,v)$ for $v\geq0$ then gives exact reconstruction,
$D(F_t(E(\nu_0)))=\mathcal N(e^{-at}m_0,e^{-2at}v_0+v_\infty(1-e^{-2at}))=\nu_t$.
\end{illustrativeexample}

\section{DisKO Model and Learning Algorithm}
\label{sec:method}

\looseness=-1
DisKO instantiates the distribution encoder $E$, latent  flow $F_t$, and distribution decoder $D$ from Section~\ref{sec:theory} with a set network, affine dynamics, and conditional flow matching.
We describe these components and their joint training, followed by a discrete closed-form alternative to continuous generator learning.

\subsection{Distribution Encoding with Set Networks}
\label{sec:set-encoder}

\looseness=-1
For a snapshot $S=\{y_i\}_{i=1}^N$, $E_\theta(S)$ denotes the encoding of its empirical distribution.
We use permutation-invariant networks based on Deep Sets \citep{zaheer2017deep} and Set Transformer \citep{lee2019set}, combining a learned aggregate $g_\theta(S)$ with snapshot statistics,
$E_\theta(S)=\rho_\theta\!\left( [g_\theta(S),\bar y_S,s_S]\right)\in\mathbb R^m,$
where $\bar y_S$ and $s_S$ are the coordinatewise mean and standard deviation with normalization $1/N$, brackets denote concatenation, and $\rho_\theta$ is a multilayer perceptron.
These statistics give the encoder direct access to the distribution's location and marginal spread.
The pooling variant concatenates the mean and coordinatewise maximum of point features from a shared network $\phi_\theta$.
The attention variant processes these features with self-attention blocks and pools them through multihead attention with a learned query \citep{lee2019set}.
Its blocks use residual connections, feedforward layers, and layer normalization.
Both variants produce a fixed-dimensional vector invariant to sample order; the choice is specified for each dataset.
The interface also supports other pooling networks or Galerkin attention \citep{cao2021galerkin} with permutation-invariant pooling.

\subsection{Continuous-Time Latent Dynamics}
\label{sec:latent-implementation}

\looseness=-1
We propagate the encoded coordinates with the affine generator in Equation~\ref{eq:latent-generator}, learning $A,c$ through an augmented matrix exponential,
\begin{equation}
    B=\begin{pmatrix}A&c\\0&0\end{pmatrix},
    \qquad
    \begin{pmatrix}F_{\Delta t}(z)\\1\end{pmatrix}
    =\exp(\Delta t B)\begin{pmatrix}z\\1\end{pmatrix}.
    \label{eq:augmented-latent-flow}
\end{equation}
A single parameter set governs all intervals and gives $F_{t+s}=F_t\circ F_s$, supporting irregular observation times and prediction at arbitrary future times.
The affine term accommodates nonzero offsets in observable evolution, as in Example~\ref{ex:ou-distribution-observables}.
Gradients pass through the matrix exponential to the generator and encoder.
The eigenvalues of $A$ describe latent decay, growth, and oscillation rates; their agreement with the underlying system is assessed experimentally.

\subsection{Distribution Decoding via Conditional Flow Matching}
\label{sec:cfm-decoder}

\looseness=-1
To recover a distribution from the propagated coordinates, we implement $D_\psi$ using conditional flow matching \citep{lipman2023flow,tong2024improving}.
Given a condition $z$ and target snapshot $S$, draw $y$ uniformly from $S$, independent noise $\epsilon\sim\mathcal N(0,I_p)$, and $\alpha\sim\mathcal U(0,1)$, where $p$ is the data dimension.
The linear path and its velocity are
$\xi_\alpha=(1-\alpha)\epsilon+\alpha y, \; u=y-\epsilon.$
A multilayer perceptron $v_\psi$ receives the concatenation of $\xi_\alpha$, $z$, and $\alpha$ and minimizes
\begin{equation}
    \ell_{\mathrm{CFM}}(z,S)
    =\mathbb E_{\alpha,\epsilon,y}\!\left[
        \frac{1}{p}\left\|v_\psi(\xi_\alpha,\alpha,z)-(y-\epsilon)\right\|_2^2
    \right].
    \label{eq:cfm-loss}
\end{equation}
To sample from $D_\psi(z)$, we independently draw $U_0\sim\mathcal N(0,I_p)$ for each sample and numerically integrate
$\frac{dU_\alpha}{d\alpha}=v_\psi(U_\alpha,\alpha,z), \; \alpha\in[0,1], \; D_\psi(z)=\operatorname{Law}(U_1).$
Here, $\alpha$ parametrizes generation at a fixed physical time.
Physical evolution enters through the shared condition $z=F_{t-s}(E_\theta(S_s))$ when predicting from time $s$ to $t$.
This decoder accommodates high-dimensional distributions, including the 30-dimensional Pancreas and 50-dimensional WOT iPSC representations in our experiments.
For low-dimensional data with suitable structure, conditional Gaussian mixtures or other parametric families provide alternative decoders with interpretable parameters, trained with an appropriate likelihood or distribution loss.

\subsection{Training Objectives and Algorithm}
\label{sec:training}

\looseness=-1
To couple representation learning with temporal prediction, we draw pairs $(S_s,S_t)$ from the same snapshot sequence, with $s<t$ in the training window.
Points are sampled independently from each snapshot.
Define $z_s=E_\theta(S_s)$, $z_t=E_\theta(S_t)$, and $\widehat z_t=F_{t-s}(z_s)$.
The prediction, reconstruction, and latent consistency objectives are
\begin{equation}
    \begin{aligned}
        \mathcal J_{\mathrm{pred}}=\mathbb E\,\ell_{\mathrm{CFM}}(\widehat z_t,S_t),\ 
        \mathcal J_{\mathrm{rec}}=\mathbb E\,\ell_{\mathrm{CFM}}(z_t,S_t),\
        \mathcal J_{\mathrm{lat}}=\mathbb E\!\left[
            \frac{1}{m}\|\widehat z_t-\operatorname{sg}(z_t)\|_2^2
        \right],
    \end{aligned}
    \label{eq:training-components}
\end{equation}
where expectations are over training pairs, $m$ is the latent dimension, and $\operatorname{sg}$ stops the gradient.
Reconstruction encourages the encoding to retain distributional information, while prediction trains the decoder with the propagated condition used at inference.
Latent consistency aligns the propagated coordinates with the target encoding, passing its encoder gradient through the source branch.
We further compare a generated set $\widetilde S_t$ from $D_\psi(\widehat z_t)$ with an equally sized target subset $S_t^*$,
\begin{equation}
    \mathcal J_{\mathrm{dist}}=\mathbb E\!\left[
        \operatorname{SW}_1(\widetilde S_t,S_t^*)
        +\beta\operatorname{MMD}^2(\widetilde S_t,S_t^*)
        +\gamma d_{\mathrm{mom}}(\widetilde S_t,S_t^*)
    \right].
    \label{eq:distribution-training-loss}
\end{equation}
Here, $\operatorname{SW}_1$ averages one-dimensional Wasserstein distances over random projections \citep{bonneel2015sliced}, $\operatorname{MMD}^2$ uses a mixture of Gaussian kernels \citep{gretton2012kernel}, and $d_{\mathrm{mom}}$ sums mean squared differences between sample means and raw second-moment matrices.
Gradients pass through the numerical sampler to supervise the generated distribution.
We combine the four terms with nonnegative weights,
\begin{equation}
    \mathcal J=
    \lambda_{\mathrm{pred}}\mathcal J_{\mathrm{pred}}
    +\lambda_{\mathrm{rec}}\mathcal J_{\mathrm{rec}}
    +\lambda_{\mathrm{lat}}\mathcal J_{\mathrm{lat}}
    +\lambda_{\mathrm{dist}}\mathcal J_{\mathrm{dist}}.
    \label{eq:joint-objective}
\end{equation}

\looseness=-1
Continuous generator learning proceeds in three stages.
We first pretrain $E_\theta,D_\psi$ on individual snapshots with $\ell_{\mathrm{CFM}}(E_\theta(S),S)$, then fix them and fit $A,c$ using latent prediction error on training pairs.
Finally, we jointly optimize all components with Equation~\ref{eq:joint-objective}.
This initializes reconstruction and propagation before coupling them through the full objective.
We use AdamW with gradient clipping; the full training procedure is detailed in Appendix~\ref{app:training-procedure}, with hyperparameter settings in Table~\ref{tab:extrapolation-training-settings}.

\noindent\textbf{Discrete closed-form updates.}\enspace
\label{sec:discrete-closed-form}
\looseness=-1
For uniformly spaced snapshots, an alternative is to learn the action of $\mathcal K_{\Delta t}$ directly through $F_{\Delta t}(z)=Kz+b$.
Let $X$ contain augmented source encodings $[z_s^\top,1]^\top$ as columns and $Y$ the corresponding target encodings at $s+\Delta t$, using all adjacent pairs from training sequences within the training window.
With $\Theta=[K\ b]$, the ridge update is
\begin{equation}
    \widehat\Theta
    =\underset{\Theta}{\operatorname{argmin}}\,
    \|Y-\Theta X\|_F^2+\eta\|\Theta\|_F^2
    =YX^\top(XX^\top+\eta I)^{-1},
    \qquad \eta>0.
    \label{eq:discrete-ridge}
\end{equation}
The fit uses unstandardized latent coordinates and a sum of squared residuals over pairs, with a penalty on both $K$ and $b$.
We evaluate the solution by singular value decomposition.
After the same pretraining, this fit initializes the discrete dynamics.
Joint training alternates encoder and decoder updates under Equation~\ref{eq:joint-objective} with re-encoding the training pairs and refitting $K,b$.
The ridge solve is detached from the gradient computation; $K,b$ are fixed during each gradient step, while propagation retains gradients to the source encoder.
The spectral experiments refit after every step with $\eta=10^{-3}$ (Appendix~\ref{app:spectral-parameterizations}).
Repeated application of $F_{\Delta t}$ yields $F_{q\Delta t}$ for nonnegative integers $q$, supporting the same prediction losses over multiple horizons.
For either variant, inference encodes the source snapshot once, propagates its coordinates to each requested time, and decodes the corresponding distribution.

\section{Results}
\label{sec:experiments}

\subsection{Experimental Setup}
\label{sec:experimental-setup}

\looseness=-1
\noindent\textbf{Datasets.}\enspace
We evaluate DisKO's ability to extrapolate distribution dynamics on the following seven datasets.
OU and Duffing test long-horizon extrapolation across multiple initial distributions (\textit{50 and 80 future steps}, respectively), while Boids tests extrapolation from a single sequence (\textit{50 future steps}).
Ocean tests prediction from a short sequence (one future step).
CAMELS tests \textit{generalization across initial distributions} in 10 dimensions.
Pancreas and WOT iPSC assess prediction on real single-cell data (\textit{30 and 50 dimensions}, respectively), with WOT iPSC spanning 39 snapshots.
Circle and Torus provide analytic spectra for spectral validation.
Dataset details appear in Appendix~\ref{app:datasets}.

\looseness=-1
\noindent\textbf{Experimental protocol.}\enspace
We divide the observation times into training and extrapolation windows.
For datasets with multiple initial distributions (OU, Duffing, and CAMELS), we also split the initial distributions into training and test sets.
Detailed splits are provided in Table~\ref{tab:dataset-protocol} of Appendix~\ref{app:datasets}.

\looseness=-1
\noindent\textbf{Evaluation.}\enspace
We quantify prediction accuracy using $W_1$ and $\mathrm{SW}_1$ between predicted and ground truth distributions; $W_2$ and $\mathrm{MMD}^2$ are additionally reported in the appendix.

\looseness=-1
\noindent\textbf{Baselines.}\enspace
We compare DisKO with eight population dynamics models to assess its predictive advantage.
TrajectoryNet learns regularized transport flows \citep{tong2020trajectorynet}.
DeepRUOT models stochasticity and growth through unbalanced transport \citep{zhang2025deepruot}.
VGFM jointly estimates velocity and growth by flow matching \citep{wang2025vgfm}.
MIOFlow learns latent differential dynamics with a geodesic autoencoder \citep{huguet2022mioflow}.
scNODE couples a variational autoencoder to a latent Neural ODE \citep{zhang2024scnode}.
PRESCIENT models stochastic evolution through a potential landscape \citep{yeo2021prescient}.
JKOnet* identifies energy terms using Wasserstein gradient-flow optimality \citep{terpin2024lightspeed}.
WLM represents inertial population dynamics with a Wasserstein Lagrangian \citep{guan2026lagrangian}.

\subsection{Distribution Extrapolation}
\label{sec:extrapolation-results}

\looseness=-1
Table~\ref{tab:test-extrapolation} reports distribution extrapolation results for DisKO and eight baselines across seven datasets.
DisKO achieves the lowest $W_1$ on all seven datasets and the lowest $\mathrm{SW}_1$ on six.
Relative to the strongest baseline on each dataset, DisKO reduces $W_1$ by $58.6\%$ on OU, $53.9\%$ on Duffing, and $36.2\%$ on Ocean.
The corresponding reductions on the real single-cell datasets Pancreas and WOT iPSC are $18.0\%$ and $18.7\%$, respectively.
Complete numerical results across time windows and initial distribution splits are provided in Tables~\ref{tab:train-time-train-initial}--\ref{tab:test-time-test-initial} of Appendix~\ref{app:distribution-results}.

\begin{table}[!htbp]
\centering
\begingroup
\definecolor{resultFirst}{RGB}{190,25,35}
\definecolor{resultSecond}{RGB}{25,80,185}
\caption{Results over the extrapolation window for test initial distributions in datasets with multiple initial distributions, and for single sequences.
\textcolor{resultFirst}{Red} and \textcolor{resultSecond}{blue}
indicate the best and second best results, respectively.}
\label{tab:test-extrapolation}
\fontsize{8}{10.5}\selectfont
\setlength{\tabcolsep}{1pt}
\renewcommand{\arraystretch}{0.86}
\setlength{\aboverulesep}{0.8pt}
\setlength{\belowrulesep}{1.1pt}
\newcommand{\resultmetric}[2]{\mbox{#1\textsubscript{\fontsize{6}{7}\selectfont$\pm$#2}}}
\begin{tabular*}{\linewidth}{@{}l@{\hspace{4pt}}c@{\hspace{4pt}\extracolsep{\fill}}ccccccc@{}}
\toprule
Method & Metric & OU & Duffing & Boids & Ocean & CAMELS & Pancreas & WOT iPSC \\
\midrule
\multirow{2}{*}[-1pt]{TrajectoryNet} & $W_1$ & \resultmetric{1.369}{0.071} & \resultmetric{9.222}{0.783} & \resultmetric{11.843}{0.560} & \resultmetric{2.198}{0.900} & \resultmetric{143.767}{2.170} & \resultmetric{47.376}{2.950} & \resultmetric{118.524}{2.509} \\
\arrayrulecolor{black!45}
\cmidrule[0.2pt](l{1pt}r{1pt}){3-9}
\arrayrulecolor{black}
 & SW1 & \resultmetric{0.857}{0.047} & \resultmetric{5.815}{0.712} & \resultmetric{7.267}{0.365} & \resultmetric{1.426}{0.556} & \resultmetric{35.364}{1.128} & \resultmetric{6.765}{0.826} & \resultmetric{14.502}{0.647} \\
\specialrule{0.25pt}{0.7pt}{0.7pt}
\multirow{2}{*}[-1pt]{DeepRUOT} & $W_1$ & \resultmetric{0.846}{0.044} & \resultmetric{1.032}{0.287} & \resultmetric{5.962}{0.283} & \resultmetric{0.214}{0.020} & \resultmetric{79.989}{1.252} & \resultmetric{17.286}{1.572} & \resultmetric{23.635}{1.623} \\
\arrayrulecolor{black!45}
\cmidrule[0.2pt](l{1pt}r{1pt}){3-9}
\arrayrulecolor{black}
 & SW1 & \resultmetric{0.554}{0.030} & \resultmetric{0.650}{0.186} & \resultmetric{3.796}{0.172} & \resultmetric{0.149}{0.014} & \resultmetric{14.362}{0.466} & \resultmetric{2.113}{0.221} & \resultmetric{2.108}{0.220} \\
\specialrule{0.25pt}{0.7pt}{0.7pt}
\multirow{2}{*}[-1pt]{VGFM} & $W_1$ & \resultmetric{1.290}{0.486} & \resultmetric{12.834}{0.976} & \resultmetric{31.008}{1.709} & \resultmetric{0.195}{0.047} & \resultmetric{123.071}{2.075} & \resultmetric{27.497}{1.337} & \resultmetric{187.154}{2.894} \\
\arrayrulecolor{black!45}
\cmidrule[0.2pt](l{1pt}r{1pt}){3-9}
\arrayrulecolor{black}
 & SW1 & \resultmetric{0.807}{0.315} & \resultmetric{9.358}{0.352} & \resultmetric{19.604}{1.428} & \resultmetric{0.128}{0.034} & \resultmetric{24.995}{17.602} & \resultmetric{2.978}{2.126} & \resultmetric{27.389}{1.762} \\
\specialrule{0.25pt}{0.7pt}{0.7pt}
\multirow{2}{*}[-1pt]{MIOFlow} & $W_1$ & \resultmetric{1.514}{0.135} & \resultmetric{0.692}{0.109} & \resultmetric{9.642}{0.390} & \resultmetric{0.187}{0.044} & \resultmetric{98.173}{1.421} & \textcolor{resultSecond}{\resultmetric{11.111}{0.902}} & \resultmetric{34.854}{0.627} \\
\arrayrulecolor{black!45}
\cmidrule[0.2pt](l{1pt}r{1pt}){3-9}
\arrayrulecolor{black}
 & SW1 & \resultmetric{0.950}{0.087} & \resultmetric{0.432}{0.070} & \resultmetric{5.879}{1.216} & \resultmetric{0.130}{0.033} & \resultmetric{20.381}{1.104} & \textcolor{resultSecond}{\resultmetric{0.895}{0.171}} & \resultmetric{3.931}{0.053} \\
\specialrule{0.25pt}{0.7pt}{0.7pt}
\multirow{2}{*}[-1pt]{scNODE} & $W_1$ & \resultmetric{0.759}{0.010} & \resultmetric{0.887}{0.075} & \resultmetric{7.822}{0.359} & \resultmetric{0.783}{0.246} & \resultmetric{143.751}{1.692} & \resultmetric{43.047}{1.030} & \resultmetric{94.268}{1.536} \\
\arrayrulecolor{black!45}
\cmidrule[0.2pt](l{1pt}r{1pt}){3-9}
\arrayrulecolor{black}
 & SW1 & \resultmetric{0.481}{0.008} & \resultmetric{0.541}{0.070} & \resultmetric{4.381}{0.233} & \resultmetric{0.515}{0.154} & \resultmetric{34.896}{1.137} & \resultmetric{6.168}{0.831} & \resultmetric{10.544}{1.150} \\
\specialrule{0.25pt}{0.7pt}{0.7pt}
\multirow{2}{*}[-1pt]{PRESCIENT} & $W_1$ & \resultmetric{0.849}{0.034} & \resultmetric{0.532}{0.040} & \resultmetric{3.498}{0.136} & \textcolor{resultSecond}{\resultmetric{0.153}{0.009}} & \resultmetric{143.797}{0.232} & \resultmetric{12.169}{0.325} & \textcolor{resultSecond}{\resultmetric{23.328}{0.086}} \\
\arrayrulecolor{black!45}
\cmidrule[0.2pt](l{1pt}r{1pt}){3-9}
\arrayrulecolor{black}
 & SW1 & \resultmetric{0.525}{0.020} & \resultmetric{0.280}{0.025} & \resultmetric{1.951}{0.087} & \textcolor{resultSecond}{\resultmetric{0.097}{0.006}} & \resultmetric{36.244}{0.062} & \resultmetric{0.965}{0.068} & \textcolor{resultSecond}{\resultmetric{1.883}{0.013}} \\
\specialrule{0.25pt}{0.7pt}{0.7pt}
\multirow{2}{*}[-1pt]{JKOnet*} & $W_1$ & \resultmetric{1.288}{0.067} & \resultmetric{5.246}{0.447} & \resultmetric{3.282}{0.187} & \resultmetric{2.332}{0.876} & \resultmetric{134.105}{2.073} & \resultmetric{52.395}{1.119} & \resultmetric{63.051}{1.367} \\
\arrayrulecolor{black!45}
\cmidrule[0.2pt](l{1pt}r{1pt}){3-9}
\arrayrulecolor{black}
 & SW1 & \resultmetric{0.797}{0.044} & \resultmetric{3.290}{0.413} & \resultmetric{1.770}{0.086} & \resultmetric{1.561}{0.587} & \resultmetric{30.058}{0.976} & \resultmetric{8.221}{0.713} & \resultmetric{7.076}{0.325} \\
\specialrule{0.25pt}{0.7pt}{0.7pt}
\multirow{2}{*}[-1pt]{WLM} & $W_1$ & \textcolor{resultSecond}{\resultmetric{0.602}{0.032}} & \textcolor{resultSecond}{\resultmetric{0.318}{0.028}} & \textcolor{resultSecond}{\resultmetric{1.595}{0.047}} & \resultmetric{0.941}{0.264} & \textcolor{resultSecond}{\resultmetric{55.019}{0.370}} & \resultmetric{13.345}{0.517} & \resultmetric{25.805}{0.638} \\
\arrayrulecolor{black!45}
\cmidrule[0.2pt](l{1pt}r{1pt}){3-9}
\arrayrulecolor{black}
 & SW1 & \textcolor{resultSecond}{\resultmetric{0.362}{0.020}} & \textcolor{resultSecond}{\resultmetric{0.160}{0.019}} & \textcolor{resultSecond}{\resultmetric{0.656}{0.050}} & \resultmetric{0.520}{0.246} & \textcolor{resultFirst}{\resultmetric{6.699}{0.206}} & \resultmetric{0.988}{0.140} & \resultmetric{1.929}{0.093} \\
\midrule
\multirow{2}{*}[-1pt]{\textbf{DisKO}} & $W_1$ & \textcolor{resultFirst}{\resultmetric{0.249}{0.013}} & \textcolor{resultFirst}{\resultmetric{0.146}{0.005}} & \textcolor{resultFirst}{\resultmetric{1.454}{0.110}} & \textcolor{resultFirst}{\resultmetric{0.097}{0.024}} & \textcolor{resultFirst}{\resultmetric{54.241}{0.830}} & \textcolor{resultFirst}{\resultmetric{9.114}{0.143}} & \textcolor{resultFirst}{\resultmetric{18.973}{0.330}} \\
\arrayrulecolor{black!45}
\cmidrule[0.2pt](l{1pt}r{1pt}){3-9}
\arrayrulecolor{black}
 & SW1 & \textcolor{resultFirst}{\resultmetric{0.135}{0.010}} & \textcolor{resultFirst}{\resultmetric{0.056}{0.004}} & \textcolor{resultFirst}{\resultmetric{0.656}{0.087}} & \textcolor{resultFirst}{\resultmetric{0.062}{0.019}} & \textcolor{resultSecond}{\resultmetric{7.039}{0.500}} & \textcolor{resultFirst}{\resultmetric{0.559}{0.007}} & \textcolor{resultFirst}{\resultmetric{0.940}{0.039}} \\
\bottomrule
\end{tabular*}
\endgroup
\end{table}

\begin{figure}[!htbp]
 \centering
 \begin{minipage}[c]{.72\linewidth}
  \includegraphics[width=\linewidth]{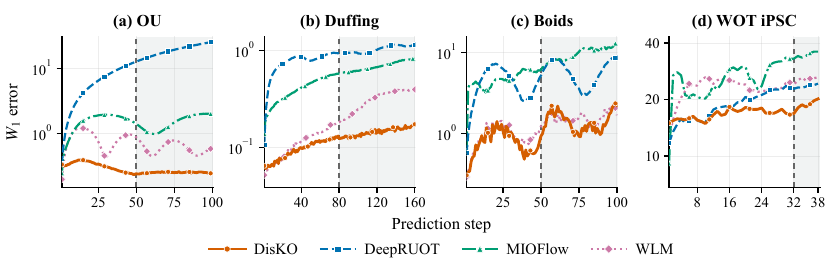}
 \end{minipage}\hfill
 \begin{minipage}[c]{.26\linewidth}
  \caption{\raggedright $W_1$ prediction errors on four benchmarks. Vertical dashed lines mark the end of training, with the extrapolation window shaded. All error axes use a logarithmic scale.}
    \label{fig:results-extrapolation-curves}
 \end{minipage}
\end{figure}

\looseness=-1
The prediction error curves in Figure~\ref{fig:results-extrapolation-curves} further characterize DisKO's long-horizon extrapolation performance.
On OU, prediction error \textbf{remains approximately constant} over 50 extrapolation steps.
On Duffing, error grows slowly over 80 extrapolation steps and remains below that of the baselines shown in the figure.
These results indicate that the learned Koopman representation supports sustained distribution prediction beyond the training window.
We further examine the contribution of the shared linear evolution structure to extrapolation stability through an ablation against a nonlinear latent Neural ODE in Section~\ref{sec:further-discussions}.
The complete error curves for all seven datasets are shown in Figure~\ref{fig:extrapolation-w1-overview} in Appendix~\ref{app:distribution-results}.

\looseness=-1
We obtain the reported performance with limited hyperparameter search, evaluating only one to three configurations per dataset.
The main table reports the results of this initial search.
For some datasets, further hyperparameter tuning can yield better results.
These results demonstrate that DisKO achieves competitive performance across datasets with a modest tuning budget.

\subsection{Koopman Spectral Approximation}
\label{sec:spectral-approximation-results}

\begin{figure}[!htbp]
 \centering
 \includegraphics[width=.84\linewidth]{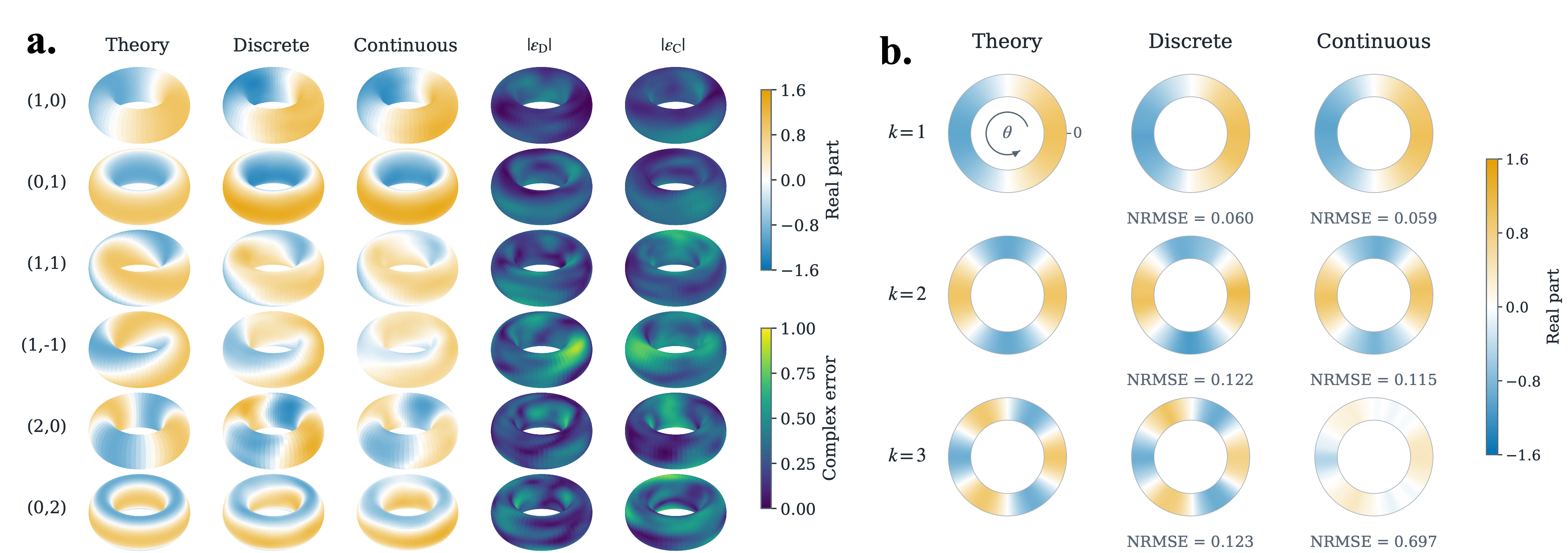}
 \caption{Distributional eigenfunctions on wrapped Gaussian probes ($\sigma=0.15$), indexed by center. (a) Torus. (b) Circle. Plots show real components and complex errors after amplitude and phase calibration on training snapshots.}
 \label{fig:results-spectrum}
\end{figure}

\looseness=-1
Recovering distributional Koopman spectra from finite, unpaired snapshots is challenging because the eigenfunctions are defined on an infinite-dimensional space of probability measures.
We assess distributional Koopman recovery on Circle ($q=1$) and Torus ($q=2$), governed by angular drift and diffusion,
$d\vartheta_{j,t}=\omega_j\,dt+\sqrt{2\kappa_j}\,dW_{j,t} \pmod{2\pi},\; j=1,\ldots,q,$
with independent Brownian motions.
Circle uses $(\omega_1,\kappa_1)=(1,0.01)$; Torus uses $\omega=(1,1.7)$ and $\kappa=(0.01,0.014)$.
The analytical eigenpairs for $k\in\mathbb Z^q$ satisfy
\begin{equation}
 \begin{aligned}
 h_k(\mu)&=\int e^{\mathrm{i}k\cdot\vartheta}\,d\mu(\vartheta),
 &\lambda_k&=-\sum_{j=1}^q\kappa_j k_j^2+\mathrm{i}\,\omega\cdot k,\\
 \mathcal K_t h_k&=e^{t\lambda_k}h_k,
 &\rho_k&=e^{\Delta t\lambda_k}.
 \end{aligned}
 \label{eq:results-fourier-eigenpairs}
\end{equation}
Here $h_k$ is a distributional Koopman eigenfunction with continuous-time generator eigenvalue $\lambda_k$, finite-time eigenvalue $\rho_k$, decay rate $-\operatorname{Re}(\lambda_k)$, and angular frequency $\operatorname{Im}(\lambda_k)$.

\looseness=-1
Uniformly spaced snapshots permit direct, closed-form fitting of the finite-time Koopman representation, providing a complementary route to spectral recovery.
We therefore evaluate discrete DisKO (Section~\ref{sec:discrete-closed-form}) alongside the continuous generator model (Section~\ref{sec:latent-implementation}).
Table~\ref{tab:results-spectrum} shows close agreement between analytical and recovered finite-time eigenvalues at $\Delta t=0.1$ for both parameterizations.
Absolute errors are below $0.006$ for Circle's first three harmonics and $0.003$ for the Torus fundamental modes.

\begin{table}[!htbp]
\centering
\begin{minipage}[c]{.60\linewidth}
\small
\setlength{\tabcolsep}{2.5pt}
\renewcommand{\arraystretch}{0.9}
\setlength{\aboverulesep}{0.6pt}
\setlength{\belowrulesep}{0.8pt}
\begin{tabular*}{\linewidth}{@{}l@{\extracolsep{\fill}}ccccc@{}}
\toprule
 & \multicolumn{3}{c}{Circle} & \multicolumn{2}{c}{Torus} \\
\cmidrule(lr){2-4}\cmidrule(lr){5-6}
DisKO & $k=1$ & $k=2$ & $k=3$ & $k=(1,0)$ & $k=(0,1)$ \\
\midrule
Discrete & 1.315 & 3.191 & 5.928 & 2.404 & 2.892 \\
Continuous & 1.314 & 2.410 & 5.868 & 1.781 & 1.377 \\
\bottomrule
\end{tabular*}
\end{minipage}\hfill
\begin{minipage}[c]{.38\linewidth}
\setlength{\abovecaptionskip}{0pt}
\caption{\raggedright Absolute eigenvalue errors $|\widehat\rho_k-\rho_k|$ ($\times10^{-3}$) for $\mathcal K_{0.1}$. Full eigenvalues and higher Circle harmonics appear in Appendix~\ref{app:eigenvalue-approximation}.}
\label{tab:results-spectrum}
\end{minipage}
\end{table}

\looseness=-1
The nonlinear set encoder coordinates and the constant observable span a finite-dimensional observable space.
Left eigenvectors of the augmented operator or generator define candidate eigenfunctions through linear combinations $u_k(\nu)=w_k^\top[E(\nu)^\top,1]^\top$ (Appendix~\ref{app:spectral-parameterizations}).
We evaluate their shapes on wrapped Gaussian probes with center $a$ and covariance $\sigma^2 I$, where $h_k(\mu_{a,\sigma})=e^{-\sigma^2\|k\|^2/2}e^{\mathrm{i}k\cdot a}$ (Figure~\ref{fig:results-spectrum}).
On future test snapshots, complex correlations reach $0.999$, $0.994$, and $0.976$ for the first three Circle harmonics with discrete DisKO and exceed $0.995$ for both Torus fundamental modes with either parameterization (Appendix~\ref{app:eigenfunction-evaluation}).

\looseness=-1
DisKO thus learns an observable space containing accurate approximations of analytical Fourier distribution functionals \textbf{directly from snapshots, without a prescribed Fourier dictionary.}
Discrete DisKO better resolves higher Circle harmonics, while continuous DisKO gives more accurate Torus eigenvalues.
Eigenvalue accuracy and eigenfunction fidelity are therefore distinct tests.

\subsection{Dynamical Closure and Extrapolation Stability}
\label{sec:further-discussions}

\looseness=-1
On OU, we test approximate closure of the learned observables under Koopman evolution and the contribution of this structure to extrapolation stability.

\begin{figure}[!htbp]
 \centering
 \begin{minipage}[c]{.62\linewidth}
  \includegraphics[width=\linewidth]{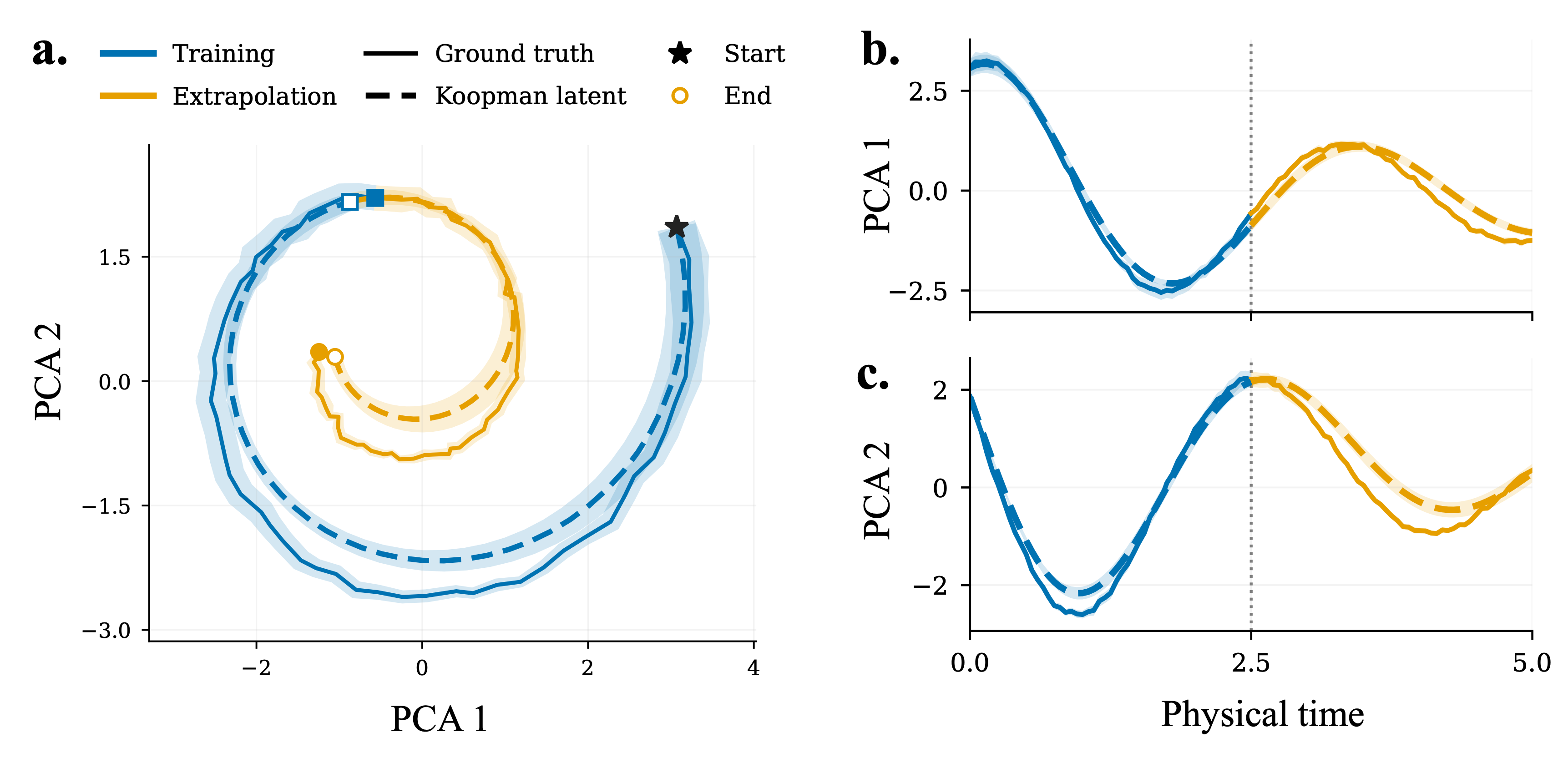}
 \end{minipage}\hfill
 \begin{minipage}[c]{.36\linewidth}
  \caption{\raggedright Reference encodings and forecasts for one OU test sequence, with one standard deviation across five seeds. PCA and rigid alignment use all times after forecasting.}
 \label{fig:results-latent}
 \end{minipage}
\end{figure}

\noindent\textbf{Latent dynamical consistency.}\enspace
\label{sec:latent-dynamics-results}
\looseness=-1
For a nearly closed observable subspace, propagating the initial encoding should agree with directly encoding later distributions.
With the trained encoder fixed, Figure~\ref{fig:results-latent} compares these trajectories, $F_t(E(\widehat\nu_0))$ and $E(\widehat\nu_t)$.
Their agreement in rotation and contraction beyond the training interval \textbf{supports approximate closure} along this OU trajectory in the PCA projection (Appendix~\ref{app:latent-dynamics}).

\begin{figure}[!htbp]
 \centering
 \begin{minipage}[c]{.62\linewidth}
  \includegraphics[width=\linewidth]{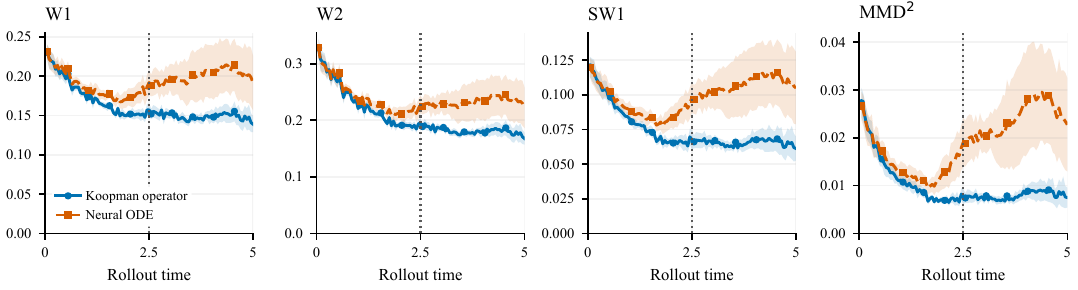}
 \end{minipage}\hfill
 \begin{minipage}[c]{.36\linewidth}
  \caption{\raggedright OU errors averaged over 64 test initial distributions and five seeds. Shading shows one seed standard deviation; the dotted line marks the training boundary. Forecasts start at $t=0$.}
 \label{fig:results-node}
 \end{minipage}
\end{figure}

\noindent\textbf{Ablation of Koopman latent dynamics.}\enspace
\label{sec:ablation-results}
\looseness=-1
To isolate Koopman latent dynamics, we replace only the affine generator with a Neural ODE, keeping encoder and decoder architectures, losses, and training budgets fixed.
Within each of five seeds, both models start from identical pretrained encoder and decoder weights, with comparable dynamics parameter counts ($4{,}160$ and $4{,}192$).
Figure~\ref{fig:results-node} shows lower future errors and smaller seed variation for Koopman dynamics in all four metrics.
Future $W_1$ decreases from $0.201\pm0.026$ to $0.149\pm0.007$, and $\mathrm{SW}_1$ from $0.107\pm0.018$ to $0.065\pm0.005$.
The OU system has a stable stationary distribution, so long-horizon errors do not necessarily increase monotonically: as both models approach the attracting distribution, distribution discrepancies can decrease at later times. The Koopman parameterization achieves lower future errors and reduced variability across seeds, \textit{supporting its role as an effective inductive bias for extrapolation.} (Appendix~\ref{app:ablation}).

\FloatBarrier
\clearpage
\section{Conclusion and Limitations}
\label{sec:conclusion}

DisKO extends deep Koopman learning to population evolution, showing that a shared representation learned from unpaired snapshots can support both accurate distribution extrapolation and the recovery of interpretable decay and oscillation.

Two important theoretical questions remain open. First, under what conditions are the Koopman dynamics identifiable from unpaired distribution snapshots? Second, when identifiability holds, what guarantees can be established for convergence and stability of the proposed learning procedure toward that operator?

\section*{AI Use Statement}
\label{sec:ai-use-statement}
We used generative AI tools to assist with code implementation and debugging, including support for baseline implementations, experimental visualization, and language polishing. AI tools were not used to generate research ideas, design the core methodology, formulate the theoretical framework, or produce experimental results. All AI-assisted content was reviewed and verified by the authors, who take full responsibility for the final content.

\section*{Ethics Statement}
\label{sec:ethics-statement}
This work does not involve new human-subject data collection.
Our experiments use synthetic simulations and previously released benchmark datasets, including publicly available single-cell datasets.
We do not use participant identifiers or personally identifiable information.
Data sources, preprocessing, experimental splits, and evaluation protocols are documented in the paper and appendices.
The proposed method is intended for scientific modeling of distribution dynamics; applications in sensitive or high-stakes settings should require appropriate domain-specific validation and safeguards.

\section*{Reproducibility Statement}
\label{sec:reproducibility-statement}
Dataset sources, preprocessing procedures, model architectures, training settings, evaluation protocols, and hyperparameters are documented in the paper and appendix.
Additional ablations and supporting results are provided in the appendix.

\clearpage
\bibliography{references/references}
\bibliographystyle{iclr2027_conference}

\clearpage
\appendix
This appendix provides theoretical foundations, experimental details, and supplementary analyses for DisKO.
Appendix~\ref{app:experimental-setup} describes the datasets, preprocessing, prediction protocols, and evaluation metrics.
Appendix~\ref{app:distribution-results} presents model configurations, training settings, and complete distribution extrapolation results, including error curves and distribution visualizations.
Appendix~\ref{app:spectral-approximation} details the analytical spectral references, the discrete and continuous DisKO implementations, and the evaluation of recovered eigenvalues and eigenfunctions.
Appendices~\ref{app:latent-dynamics}, \ref{app:ablation}, and~\ref{app:data-efficiency} examine latent dynamical consistency, the contribution of Koopman latent dynamics through a Neural ODE comparison, and data efficiency on the OU benchmark, respectively.
Appendix~\ref{app:revised-theory} develops the distributional Koopman representation, learned observable spaces and closure, and observation sufficiency and identifiability.

\medskip
\section{Experimental Setup}
\label{app:experimental-setup}
\setcounter{table}{0}
\renewcommand{\thetable}{\thesection.\arabic{table}}

\subsection{Datasets and Preprocessing}
\label{app:datasets}

Table~\ref{tab:dataset-protocol} summarizes the seven distribution prediction benchmarks.
Each sequence comprises temporally ordered empirical distribution snapshots, which DisKO receives as sample sets without individual correspondences between times.

\begin{table}[!ht]
\centering
\caption{Distribution prediction datasets and temporal partitions. The observation dimension is $p$. Sequence counts are training/test counts for OU, Duffing, and CAMELS; the other datasets contain one sequence. Snapshot indices start at zero. Sample counts refer to the available data before evaluation subsampling.}
\label{tab:dataset-protocol}
\begin{tabular}{lrrrrcc}
\toprule
Dataset & $p$ & Sequences & Snapshots & Samples/snapshot & Training & Future \\
\midrule
OU        & 2  & 192/64 & 101 & 1,024 & $0$--$50$ & $51$--$100$ \\
Duffing   & 2  & 192/64 & 161 & 1,024 & $0$--$80$ & $81$--$160$ \\
Boids     & 2  & 1      & 101 & 1,000 & $0$--$50$ & $51$--$100$ \\
Ocean     & 2  & 1      & 11  & 400   & $0$--$9$  & $10$ \\
CAMELS    & 10 & 21/6   & 9   & 4,096 & $0$--$5$  & $6$--$8$ \\
Pancreas  & 30 & 1      & 8   & 3,562--12,477 & $0$--$5$ & $6$--$7$ \\
WOT iPSC  & 50 & 1      & 39  & 1,956--12,416 & $0$--$32$ & $33$--$38$ \\
\bottomrule
\end{tabular}
\end{table}

\paragraph{Ornstein--Uhlenbeck (OU).}
The two-dimensional process satisfies
\begin{equation}
    dX_t=F X_t\,dt+G\,dW_t,
    \qquad
    F=\begin{pmatrix}-0.25&-2\\2&-0.25\end{pmatrix},
    \qquad
    G=\begin{pmatrix}0.35&0\\0&0.15\end{pmatrix}.
    \label{eq:benchmark-ou}
\end{equation}
Initial distributions are Gaussian mixtures with one, two, or three components.
Component means are sampled uniformly from $[-2,2]^2$, principal standard deviations from $[0.10,0.35]$, and covariance orientations uniformly over a full rotation; mixture weights follow a symmetric Dirichlet distribution with concentration one.
Each Gaussian component is propagated using its exact mean and covariance evolution, and 1,024 samples are drawn independently at each observation time.
Snapshots are spaced by $0.05$, with training through $2.5$ and extrapolation over $[2.55,5]$.

\paragraph{Duffing.}
The stochastic double-well oscillator is
\begin{equation}
    dx_t=v_t\,dt,
    \qquad
    dv_t=(x_t-x_t^3-0.4v_t)\,dt+0.35\,dW_t,
    \label{eq:benchmark-duffing}
\end{equation}
with observed state $(x_t,v_t)$.
We set mutilple initial laws, each corresponding to a different Gaussian distribution. For each law, 64 independent population simulations are generated with 5,000 particles using Euler--Maruyama integration at step size $0.005$.
Each saved snapshot contains 1,024 randomly selected particles, with sample order shuffled across times.
Observations are spaced by $0.05$ over $[0,8]$; training ends at $4$ and extrapolation covers $4.05$ to $8$.
The training and test sequence split assesses predictions for independently sampled populations from these initial laws.

\paragraph{Boids and Ocean.}
Boids contains the planar positions of 1,000 interacting agents governed by cohesion, separation, and alignment rules with a confining boundary, following the flocking model of \citet{reynolds1987flocks}.
Snapshots are spaced by $0.5$, with training over $[0,25]$ and extrapolation over $[25.5,50]$.
Ocean uses the large-vortex Gulf of Mexico benchmark distributed with WLM \citep{guan2026lagrangian}, based on a HYCOM ocean velocity field.
Its time coordinate is the snapshot index, with observations $0$ through $9$ used for training and observation $10$ for extrapolation.
Both benchmarks are evaluated in planar position space.

\paragraph{CAMELS.}
We use 27 IllustrisTNG simulations from the cosmic variance set of CAMELS \citep{villaescusa2021camels}, with fixed cosmological parameters and different random initial conditions.
Each sequence contains nine halo population snapshots, parameterized by the cosmological scale factor $a$.
The observed ten-dimensional state comprises six periodic position coordinates $\{\sin(2\pi x_j/L),\cos(2\pi x_j/L)\}_{j=1}^3$, three velocity components, and $\log_{10}M_{200c}$, with box size $L=25\,\mathrm{Mpc}/h$ and halo mass in $M_\odot/h$.
Halos with positive mass and finite position and velocity are retained, and 4,096 are sampled without replacement per snapshot.
The first six scale factors, approximately $0.143$, $0.200$, $0.263$, $0.333$, $0.423$, and $0.537$, form the training interval; predictions are evaluated at approximately $0.682$, $0.866$, and $1$.
The sequence split contains 21 training simulations and six test simulations, with no validation partition.

\paragraph{Pancreas.}
This benchmark describes human pancreatic $\beta$-cell differentiation \citep{veres2019charting}.
We use the preprocessed 30-dimensional representation released with DeepRUOTv2.\footnote{\url{https://github.com/zhenyiizhang/DeepRUOTv2}}
The resulting dataset contains 51,274 cells across eight ordered observations, with 6,190, 6,177, 3,562, 12,477, 6,480, 6,404, 6,013, and 3,971 cells, respectively.
The supplied representation is used directly, and the observation index provides the time coordinate.
The first six observations are used for training, and the final two for extrapolation.

\paragraph{WOT iPSC.}
The WOT reprogramming dataset tracks population changes during the induction of pluripotent stem cells \citep{schiebinger2019optimal}.
Our processed dataset contains 236,285 cells at 39 observation times between day $0$ and day $18$.
The 50-dimensional representation is obtained by PCA or truncated singular value decomposition fitted to cells pooled across observation times in training set.
Training uses the 33 observations through day $15$, and extrapolation uses the remaining six observations from day $15.5$ through day $18$.
The recorded observation times are used in propagation, including the irregular sampling times within the training interval.

\paragraph{Circle and Torus.}
These synthetic systems provide analytical reference spectra for dynamical validation.
On the circle, the angular process follows $d\theta_t=\omega\,dt+\sqrt{2\kappa}\,dW_t$ modulo $2\pi$, with $\omega=1$ and $\kappa=0.01$, and observations are $(\cos\theta_t,\sin\theta_t)$.
On the torus, two independent angular processes have $(\omega_1,\omega_2)=(1,1.7)$ and $(\kappa_1,\kappa_2)=(0.01,0.014)$, with observations given by the sine and cosine of each angle.
Initial laws are mixtures of wrapped Gaussians.
Both systems are observed every $0.1$ time units over $[0,12]$, with training through time $8$.
For $h_k(\mu)=\int e^{\mathrm{i}k\cdot\theta}\,d\mu(\theta)$, the distributional generator eigenvalue is
\begin{equation}
    \lambda_k=-\sum_j\kappa_j k_j^2+\mathrm{i}\sum_j\omega_j k_j,
    \label{eq:reference-fourier-spectrum}
\end{equation}
where $k$ is an integer on the circle and an integer pair on the torus.
These reference modes permit direct assessment of learned oscillation frequencies and decay rates.

\subsection{Prediction Protocol}
\label{app:prediction-protocol}

The initial distribution and time partitions follow Section~\ref{sec:problem}.
OU, Duffing, and CAMELS have distinct training and test sequences, allowing evaluation of all four combinations of initial distribution split and time window.
Boids, Ocean, Pancreas, and WOT iPSC contain a single sequence and use a temporal partition.
For these datasets, Table~\ref{tab:train-time-train-initial} reports predictions within the training window, and Table~\ref{tab:test-time-test-initial} reports temporal extrapolation.

Every prediction is conditioned on the earliest snapshot of the sequence being evaluated.
For a target time $t_k$, DisKO encodes $\widehat\nu^{(r)}_{t_0}$ and generates samples from $D(F_{t_k-t_0}(E(\widehat\nu^{(r)}_{t_0})))$.
Training-window scores also measure prediction from this initial snapshot.
For a test sequence, its first snapshot is available as input; subsequent observations are used only for evaluation, with no updates from intervening targets.
Time increments use the dataset's stated coordinate, including the source scale factor $a\approx0.143$ in CAMELS.

For DisKO, coordinatewise means and standard deviations are estimated using only training sequences within the training time window and then fixed for prediction.
Both single-cell datasets follow this standardization procedure.
Evaluation uses predictions transformed back to the input coordinates, which are the reduced representations for Pancreas and WOT iPSC.
Errors are computed separately for each sequence and target time and then averaged over the relevant sequence and temporal partition.

\subsection{Evaluation Metrics}
\label{app:evaluation-metrics}

W1, SW1, W2, and squared maximum mean discrepancy quantify prediction errors, with smaller values indicating closer agreement.
Table~\ref{tab:evaluation-budgets} gives the standard sampling budgets.
Sampling uses replacement when a snapshot contains fewer points than required.
Evaluation sampling and random projections use seed zero, with separate random streams for each dataset, initial distribution split, population, prediction horizon, and sampling operation, independently of training randomness.

\begin{table}[!ht]
\centering
\caption{Standard evaluation sample counts per distribution. SW1 uses the full evaluation sample and 128 projection directions; MMD$^2$ uses at most 512 samples per distribution. The DeepRUOT evaluations on Pancreas and CAMELS use the weighted configuration described below.}
\label{tab:evaluation-budgets}
\begin{tabular}{lrr}
\toprule
Dataset & Evaluation samples & W1/W2 sample limit \\
\midrule
OU, Duffing, Ocean & 512 & 256 \\
Boids & 1,000 & 256 \\
CAMELS, Pancreas, WOT iPSC & 1,024 & 512 \\
\bottomrule
\end{tabular}
\end{table}

\paragraph{Wasserstein distances.}
Let $\{\widetilde y_i\}_{i=1}^n$ and $\{y_i\}_{i=1}^n$ denote equally sized samples from the predicted and target distributions.
For $q\in\{1,2\}$, we compute
\begin{equation}
    \widehat W_q=
    \left(\frac{1}{n}\min_{\pi\in\mathfrak S_n}
    \sum_{i=1}^n\|\widetilde y_i-y_{\pi(i)}\|_2^q\right)^{1/q},
    \label{eq:evaluation-wasserstein}
\end{equation}
where $\mathfrak S_n$ is the set of permutations of $n$ elements.
The sample count is the minimum of the prescribed transport limit and the available counts on both sides.
The two matching problems use Euclidean and squared Euclidean costs, respectively, and are solved without entropic regularization.

\paragraph{Sliced Wasserstein distance.}
Following \citet{bonneel2015sliced}, we approximate the first-order sliced distance using $L=128$ independent directions $u_\ell$ obtained by normalizing standard Gaussian vectors in $\mathbb R^p$.
If $\widetilde s_{\ell,(i)}$ and $s_{\ell,(i)}$ are the sorted projections of the predicted and target samples onto $u_\ell$, then
\begin{equation}
    \widehat{\mathrm{SW}}_1=
    \frac{1}{Ln}\sum_{\ell=1}^{L}\sum_{i=1}^{n}
    |\widetilde s_{\ell,(i)}-s_{\ell,(i)}|.
    \label{eq:evaluation-sliced}
\end{equation}
Here $n$ is the smaller of the two evaluation sample counts, using the full balanced sample before the optimal transport limit is applied.

\paragraph{Maximum mean discrepancy.}
We use the Gaussian kernel $k_\sigma(y,y')=\exp(-\|y-y'\|_2^2/(2\sigma^2))$ and the biased squared estimator \citep{gretton2012kernel}
\begin{equation}
    \widehat{\mathrm{MMD}}^2=
    \frac{1}{n^2}\sum_{i,j=1}^{n}k_\sigma(\widetilde y_i,\widetilde y_j)
    +\frac{1}{n^2}\sum_{i,j=1}^{n}k_\sigma(y_i,y_j)
    -\frac{2}{n^2}\sum_{i,j=1}^{n}k_\sigma(\widetilde y_i,y_j).
    \label{eq:evaluation-mmd}
\end{equation}
This estimator includes diagonal terms and uses the first $n=\min(N_{\mathrm{pred}},N_{\mathrm{target}},512)$ points of each evaluation sample, where $N_{\mathrm{pred}}$ and $N_{\mathrm{target}}$ are the two available sample counts.
The bandwidth $\sigma$ is the median pairwise Euclidean distance exceeding $10^{-12}$ in the pooled predicted and target sample, with $\sigma=1$ when no such distance exists.
Entries labeled MMD in the result tables report this squared quantity.

\paragraph{Weighted evaluation for DeepRUOT.}
On Pancreas and CAMELS, DeepRUOT retains particle masses estimated by its growth model, normalized to total mass one.
Evaluation uses 1,024 samples per distribution, weighted optimal transport with all samples for W1/W2, and 128 directions for SW1.
MMD$^2$ also uses all 1,024 samples, with bandwidth determined by the target sample alone.
These entries therefore use a larger transport budget and a different kernel bandwidth convention than the standard configuration.

\FloatBarrier

\section{Distribution Extrapolation Experiments}
\label{app:distribution-results}
\setcounter{table}{0}
\renewcommand{\thetable}{\thesection.\arabic{table}}

\setcounter{figure}{0}
\renewcommand{\thefigure}{\thesection.\arabic{figure}}
\renewcommand{\theHfigure}{\thesection.\arabic{figure}}
\renewcommand{\theHtable}{\thesection.\arabic{table}}

\label{app:extrapolation-settings}
\label{app:extrapolation-curves}

\paragraph{Model architecture and implementation settings.}
All datasets use a DeepSets encoder, continuous affine latent dynamics, and a conditional flow matching decoder.
Table~\ref{tab:extrapolation-model-architecture} summarizes the state dimension, encoder depth, network width, latent dimension, and decoder depth for each dataset.

\begin{table}[!htbp]
    \centering
    \caption{DisKO model configurations for the distribution extrapolation benchmarks. Enc. and Dec. denote encoder and decoder, respectively.}
    \label{tab:extrapolation-model-architecture}
    \begin{tabular*}{\linewidth}{@{\extracolsep{\fill}}lccccc@{}}
        \toprule
        Dataset & State dim. & Enc. depth & Width & Latent dim. & Dec. depth \\
        \midrule
        OU            & 2  & 3 & 128 & 32 & 4 \\
        Duffing       & 2  & 3 & 256 & 64 & 8 \\
        Boids         & 2  & 4 & 512 & 32 & 4 \\
        Ocean         & 2  & 3 & 128 & 32 & 4 \\
        CAMELS        & 10 & 3 & 256 & 64 & 8 \\
        Pancreas          & 30 & 3 & 256 & 96 & 8 \\
        WOT iPSC          & 50 & 3 & 128 & 64 & 4 \\
        \bottomrule
    \end{tabular*}
\end{table}
\FloatBarrier

\paragraph{Training procedure.}
\label{app:training-procedure}
Each training pair comprises two snapshots $(S_s,S_t)$ from the same sequence.
We sample a valid source index uniformly, then a positive snapshot lag uniformly among those whose target remains in the training window.
Particle subsets are sampled independently, and propagation uses the actual time difference $t-s$.
Training uses varying source times; evaluation starts from the initial snapshot (Appendix~\ref{app:prediction-protocol}).
We first pretrain $E_\theta,D_\psi$ on individual training snapshots using the conditional flow matching reconstruction loss in \Eqref{eq:cfm-loss}, with dynamics fixed.
We then freeze both networks and update only $A,c$, minimizing the mean squared difference between $F_{t-s}(E_\theta(S_s))$ and $E_\theta(S_t)$.
Finally, we jointly update all three modules using the prediction, reconstruction, latent consistency, and distribution losses in \Eqref{eq:joint-objective}.
The prediction branch conditions the decoder on the propagated encoding; the reconstruction branch uses the directly encoded target.
Target encodings are detached only in the latent consistency term, so reconstruction continues to train the encoder.
Each stage uses a fresh AdamW optimizer with gradient norm clipping.
Table~\ref{tab:extrapolation-training-settings} specifies stage budgets, learning rates, batch sizes, snapshot sample counts, and loss weights.
The discrete variant initializes its operator by ridge regression and alternates encoder and decoder updates with operator refitting, as detailed in Appendix~\ref{app:spectral-parameterizations}.

Table~\ref{tab:extrapolation-training-settings} reports the training and inference settings.
The stage order pre/dyn/joint denotes encoder and decoder pretraining, latent dynamics fitting with the encoder and decoder fixed, and joint optimization, following Section~\ref{sec:training}.
Batch denotes the training batch size, and samples/snapshot is the number of samples used from each snapshot during training.
The endpoint/inference entries give the numbers of decoder integration steps used for endpoint supervision during training and for sample generation at inference, respectively.

Loss weights are listed in the order prediction FM, oracle FM, latent consistency, and endpoint distribution loss, followed after the semicolon by the MMD and moment weights,
\[
    \lambda_{\mathrm{pred}}/\lambda_{\mathrm{rec}}/\lambda_{\mathrm{lat}}/\lambda_{\mathrm{dist}};\,\beta/\gamma.
\]
Here oracle FM is the reconstruction objective $\mathcal J_{\mathrm{rec}}$ in \Eqref{eq:training-components}, which conditions the decoder on the encoded target snapshot.
The endpoint loss $\mathcal J_{\mathrm{dist}}$ uses $\operatorname{SW}_1+\beta\operatorname{MMD}^2+\gamma d_{\mathrm{mom}}$, as defined in \Eqref{eq:distribution-training-loss}; its contribution to the joint objective is weighted by $\lambda_{\mathrm{dist}}$.
Duffing and Pancreas use $W_1$ for the endpoint discrepancy, with the same outer weight $\lambda_{\mathrm{dist}}$.
The dashes in their MMD and moment entries indicate that these two terms are not used.

\begin{table}[!htbp]
    \centering
    \caption{DisKO training and inference settings. Stage updates and learning rates follow the order pre/dyn/joint. The two panels report complementary settings for the same seven datasets. Loss weights follow the order specified in the text.}
    \label{tab:extrapolation-training-settings}
    \textbf{(a) Optimization and training samples}\par\smallskip
    \begin{tabular*}{\linewidth}{@{\extracolsep{\fill}}lcccc@{}}
        \toprule
        Dataset & \shortstack{Stage updates\\($\times 10^3$)} & \shortstack{Stage learning rates\\($\times 10^{-4}$)} & Batch & \shortstack{Samples/\\snapshot} \\
        \midrule
        OU            & $10/1/2.5$ & $3/3/3$ & 4 & 512 \\
        Duffing       & $30/3/15$  & $3/3/3$ & 4 & 512 \\
        Boids    & $30/30/75$ & $3/1/1$ & 4 & 512 \\
        Ocean    & $10/10/20$ & $3/1/1$ & 4 & 400 \\
        CAMELS & $40/3/15$  & $3/3/3$ & 4 & 1024 \\
        Pancreas          & $60/3/24$  & $3/3/3$ & 8 & 512 \\
        WOT iPSC          & $15/1.5/4$ & $3/3/3$ & 8 & 512 \\
        \bottomrule
    \end{tabular*}

    \medskip
    \textbf{(b) Loss weights and decoder integration}\par\smallskip
    \begin{tabular*}{\linewidth}{@{\extracolsep{\fill}}lccc@{}}
        \toprule
        Dataset & Loss weights & Endpoint loss & \shortstack{Endpoint/\\inference steps} \\
        \midrule
        OU            & $1/.5/.05/.2;\,.25/.5$ & $\mathcal J_{\mathrm{dist}}$ & $8/32$ \\
        Duffing       & $1/.5/.05/.2;$\,--/--   & $W_1$ & $8/32$ \\
        Boids    & $1/.25/.05/.5;\,.1/1$  & $\mathcal J_{\mathrm{dist}}$ & $16/32$ \\
        Ocean     & $1/.5/.2/.2;\,.25/.5$  & $\mathcal J_{\mathrm{dist}}$ & $8/32$ \\
        CAMELS & $1/.5/.05/.2;\,.25/.5$ & $\mathcal J_{\mathrm{dist}}$ & $8/32$ \\
        Pancreas          & $1/.5/.05/.2;$\,--/--   & $W_1$ & $8/32$ \\
        WOT iPSC          & $1/.5/.05/.2;\,.25/.5$ & $\mathcal J_{\mathrm{dist}}$ & $8/32$ \\
        \bottomrule
    \end{tabular*}
\end{table}
\FloatBarrier

\FloatBarrier

\begin{figure}[!htbp]
    \centering
    \includegraphics[width=\linewidth,height=0.80\textheight,keepaspectratio]{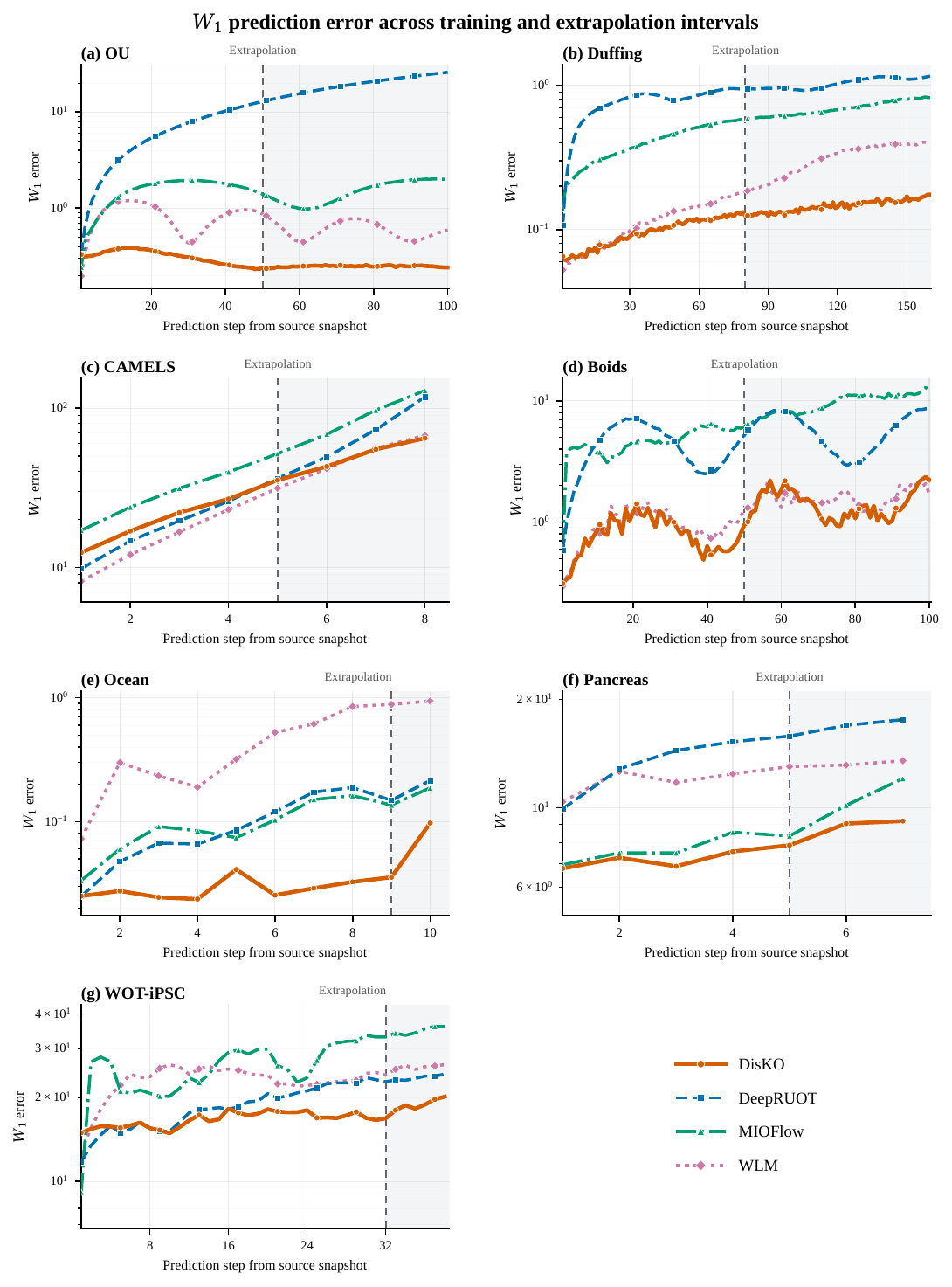}
    \caption{$W_1$ prediction error across the training and extrapolation intervals for (a) OU, (b) Duffing, (c) CAMELS, (d) Boids, (e) Ocean, (f) Pancreas, and (g) WOT iPSC. Curves compare DisKO, DeepRUOT, MIOFlow, and WLM. Vertical dashed lines mark the training boundary, and shaded regions indicate extrapolation.}
    \label{fig:extrapolation-w1-overview}
\end{figure}

\paragraph{Prediction error across the training and extrapolation intervals.}
Figure~\ref{fig:extrapolation-w1-overview} compares the $W_1$ prediction errors of DisKO, DeepRUOT, MIOFlow, and WLM across the training and extrapolation intervals for all seven datasets.
Each panel uses prediction step on the horizontal axis and a logarithmic error scale on the vertical axis.
Vertical dashed lines mark the final training step, and the shaded regions identify extrapolation.
DisKO has the lowest displayed extrapolation errors among these four methods on OU, Duffing, Ocean, Pancreas, and WOT iPSC.
Its OU error remains approximately level throughout the future interval, while its Duffing error increases gradually with the prediction horizon.
For Ocean, the error increases from the final training step at step 9 to the single future target at step 10, while remaining below the three comparison methods.
The Pancreas and WOT iPSC panels likewise show lower future errors for DisKO despite some increase beyond the training boundary.

CAMELS and Boids exhibit closer comparisons between DisKO and WLM.
On CAMELS, errors increase with the prediction horizon for all four methods; DisKO and WLM remain close near the end of the interval and finish below DeepRUOT and MIOFlow.

\clearpage
\FloatBarrier

Tables~\ref{tab:train-time-train-initial} through~\ref{tab:test-time-test-initial} provide the complete aggregate numerical results for W1, SW1, W2, and MMD across the four combinations of time window and initial distribution split.
These supporting tables retain all eight baselines and DisKO.

\AddToHookNext{file/test_time_test_initial.tex/after}{%
\clearpage

\paragraph{Prediction distribution visualizations.}
Figures~\ref{fig:extrapolation-distribution-ou} through~\ref{fig:extrapolation-distribution-wot-ipsc} compare the reference and predicted distributions for OU, Duffing, Boids, Ocean, CAMELS, Pancreas, and WOT iPSC, respectively.
The grid figures show the reference, DisKO, DeepRUOT, MIOFlow, and WLM in successive rows, with columns covering selected training and extrapolation times and the source snapshot repeated in each row.
Reference contours provide a common spatial comparison at each target time.
Ocean uses one panel per method, with time encoded by color.
OU, Duffing, Boids, and Ocean are shown in their two-dimensional observation coordinates; CAMELS is shown through halo mass and speed, and the two single-cell datasets through their first two principal component coordinates.

For OU (Figure~\ref{fig:extrapolation-distribution-ou}), DisKO follows the location and spread of the compact reference cloud over the displayed future times.
DeepRUOT predictions move outside the plotting range, while MIOFlow and WLM produce displaced or deformed distributions.
For Duffing (Figure~\ref{fig:extrapolation-distribution-duffing}), DisKO retains the two prominent lobes of the future reference distribution.
DeepRUOT concentrates near the center, MIOFlow produces a narrow band, and WLM increasingly concentrates along thin curved structures.

For Boids (Figure~\ref{fig:extrapolation-distribution-boids}), DisKO broadly reproduces the changes between ring, multiple-lobe, and compact configurations, with visible blurring and displacement at later times. DisKO also recovers substantial spatial structure.
DeepRUOT and MIOFlow miss much of the spatial extent or arrangement of the future reference populations.
For Ocean (Figure~\ref{fig:extrapolation-distribution-ocean}), DisKO follows the sequence of reference locations and places the future population near the reference endpoint, with a residual offset and differences in spread.
For CAMELS (Figure~\ref{fig:extrapolation-distribution-camels}), both DisKO and WLM reproduce much of the increasing spread in the mass and speed coordinates, while differences in the tails and local density remain visible.

For Pancreas (Figure~\ref{fig:extrapolation-distribution-pancreas}), DisKO captures the dominant compact population at the two future observations, with remaining differences in orientation and diffuse tails.
For WOT iPSC (Figure~\ref{fig:extrapolation-distribution-wot-ipsc}), DisKO follows much of the changing population shape, while the predictions at the latest times remain displaced toward lower values of the second principal component relative to the reference.

\begin{figure}[p]
    \centering
    \includegraphics[width=\linewidth,height=0.86\textheight,keepaspectratio]{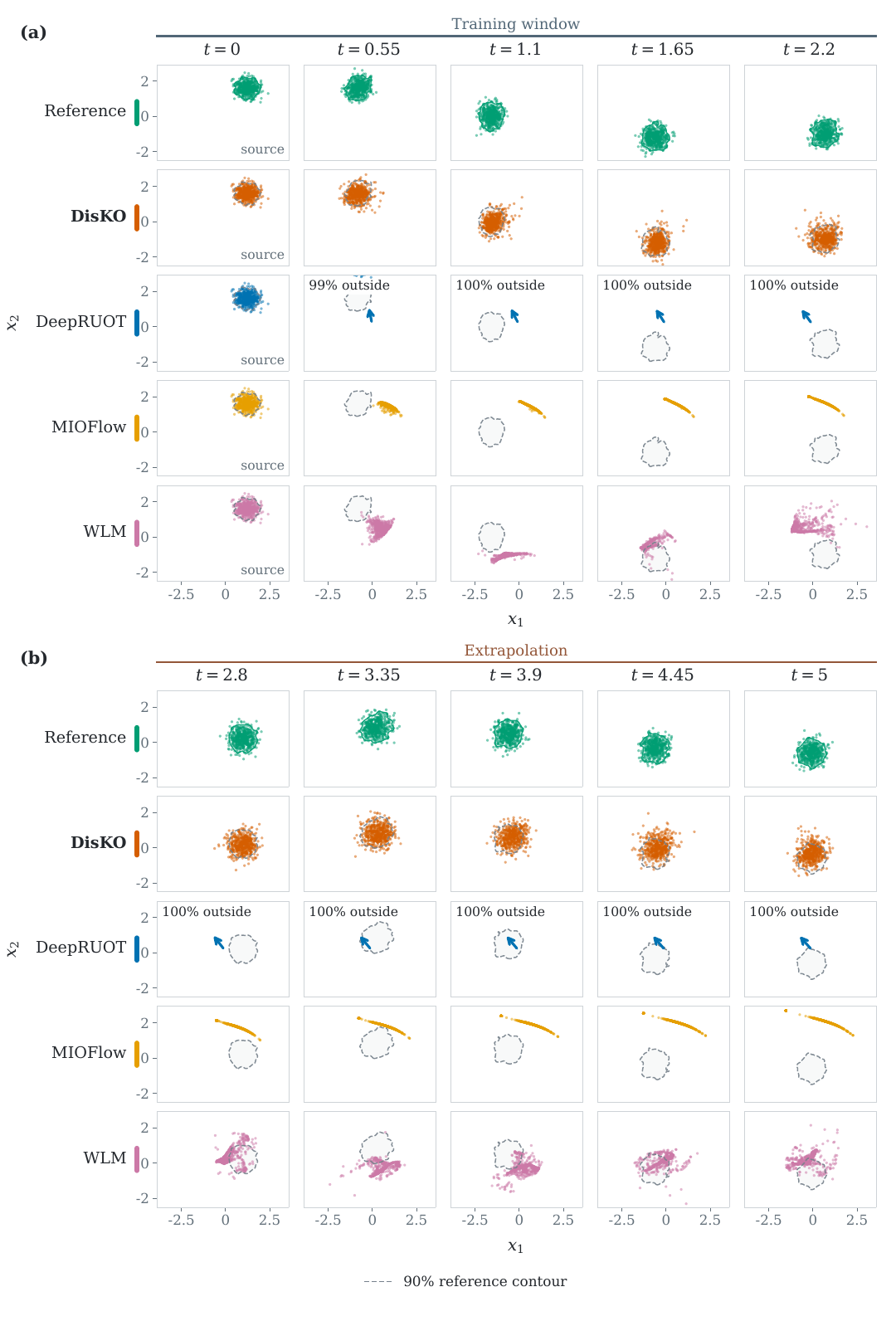}
    \caption{OU prediction distributions in the observation coordinates $x_1$ and $x_2$. Rows show the reference distribution, DisKO, DeepRUOT, MIOFlow, and WLM; columns show the times printed above each panel. Panel (a) covers the training window and panel (b) covers extrapolation. Dashed contours indicate the 90\% reference contour.}
    \label{fig:extrapolation-distribution-ou}
\end{figure}

\begin{figure}[p]
    \centering
    \includegraphics[width=\linewidth,height=0.86\textheight,keepaspectratio]{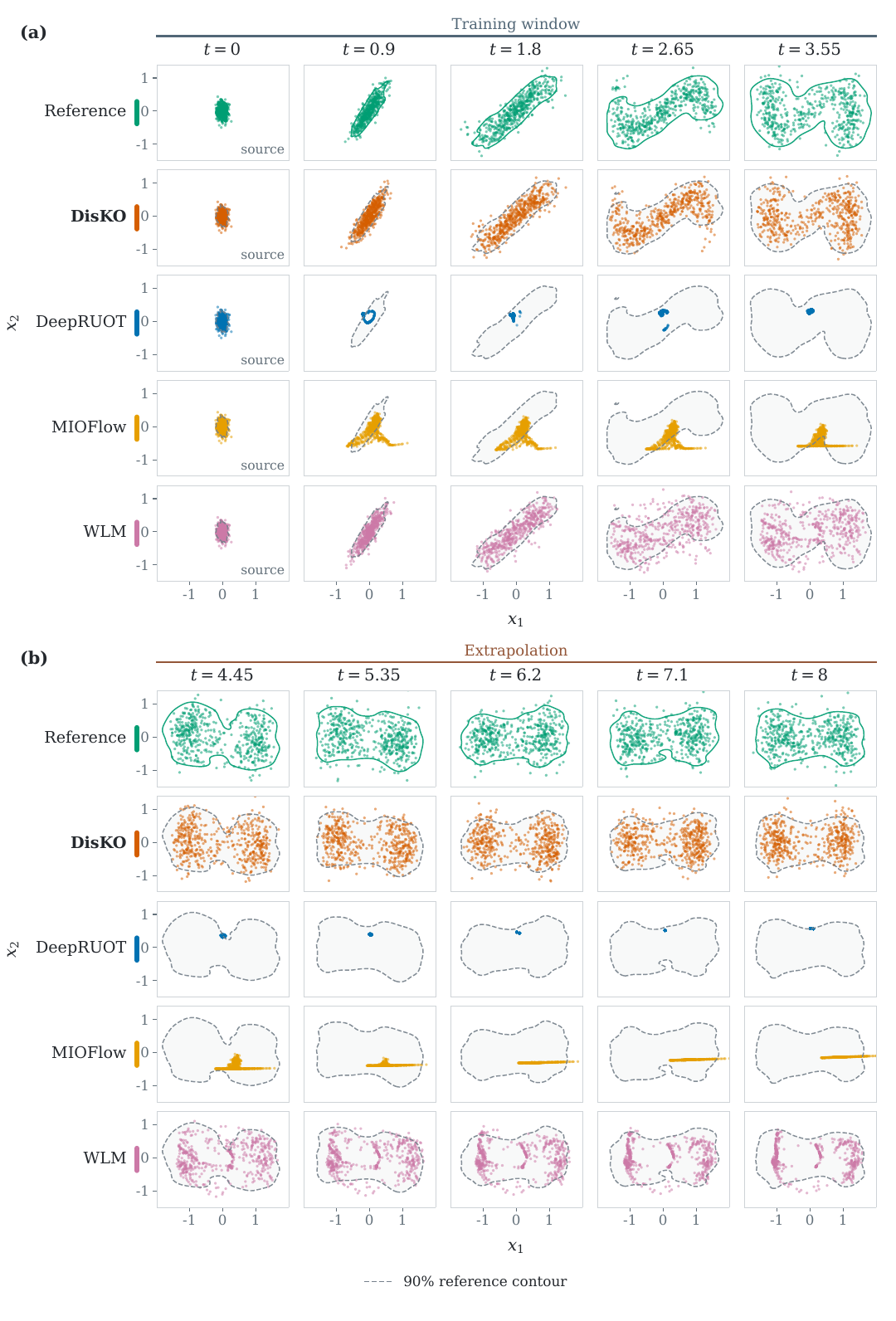}
    \caption{Duffing prediction distributions in the observation coordinates $x_1$ and $x_2$. Rows show the reference distribution, DisKO, DeepRUOT, MIOFlow, and WLM. Panel (a) covers the training window and panel (b) covers extrapolation, at the times shown above the columns. Dashed contours indicate the 90\% reference contour.}
    \label{fig:extrapolation-distribution-duffing}
\end{figure}

\begin{figure}[p]
    \centering
    \includegraphics[width=\linewidth,height=0.86\textheight,keepaspectratio]{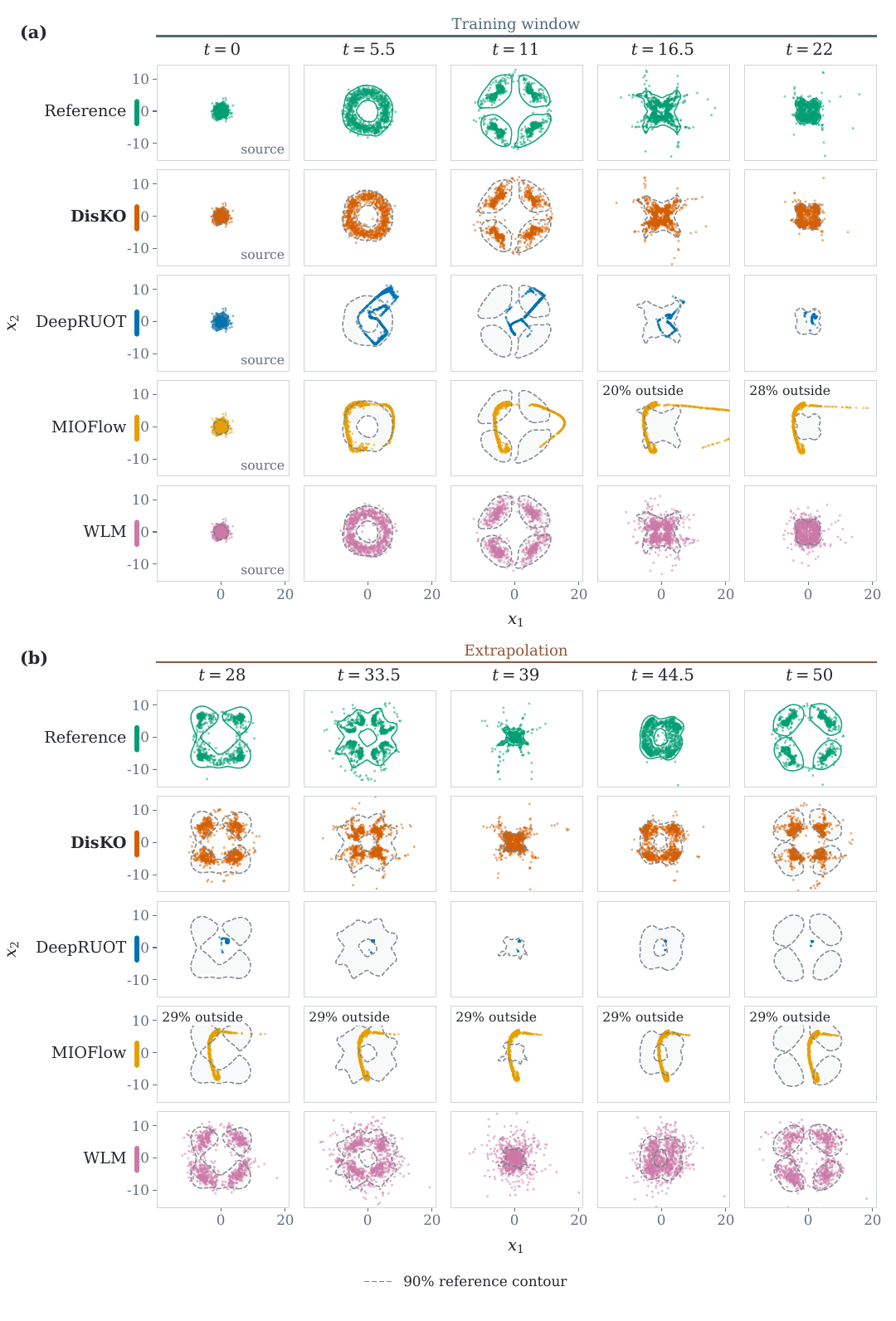}
    \caption{Boids prediction distributions in the observation coordinates $x_1$ and $x_2$. Rows show the reference distribution, DisKO, DeepRUOT, MIOFlow, and WLM. Panel (a) covers the training window and panel (b) covers extrapolation, at the times shown above the columns. Dashed contours indicate the 90\% reference contour.}
    \label{fig:extrapolation-distribution-boids}
\end{figure}

\begin{figure}[!htbp]
    \centering
    \includegraphics[width=\linewidth,height=0.86\textheight,keepaspectratio]{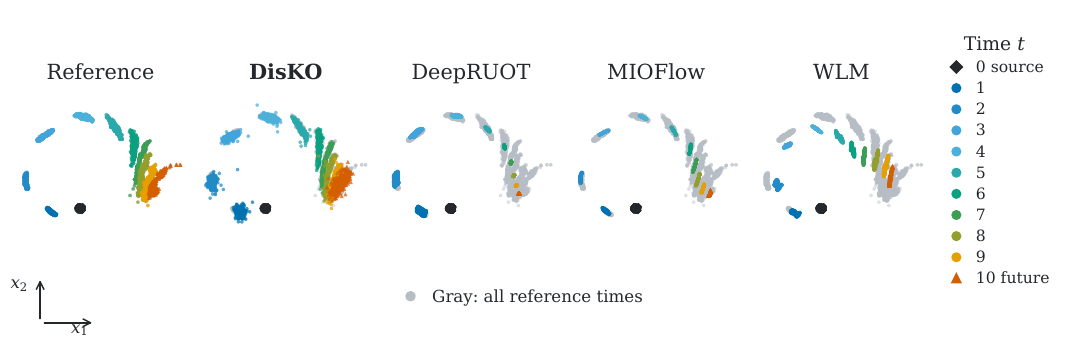}
    \caption{Ocean prediction distributions in the observation coordinates $x_1$ and $x_2$. Panels show the reference distribution, DisKO, DeepRUOT, MIOFlow, and WLM from left to right. Colors indicate time, with the source at $t=0$ and the future observation at $t=10$. Gray points show reference samples from all times.}
    \label{fig:extrapolation-distribution-ocean}
\end{figure}

\begin{figure}[p]
    \centering
    \includegraphics[width=\linewidth,height=0.86\textheight,keepaspectratio]{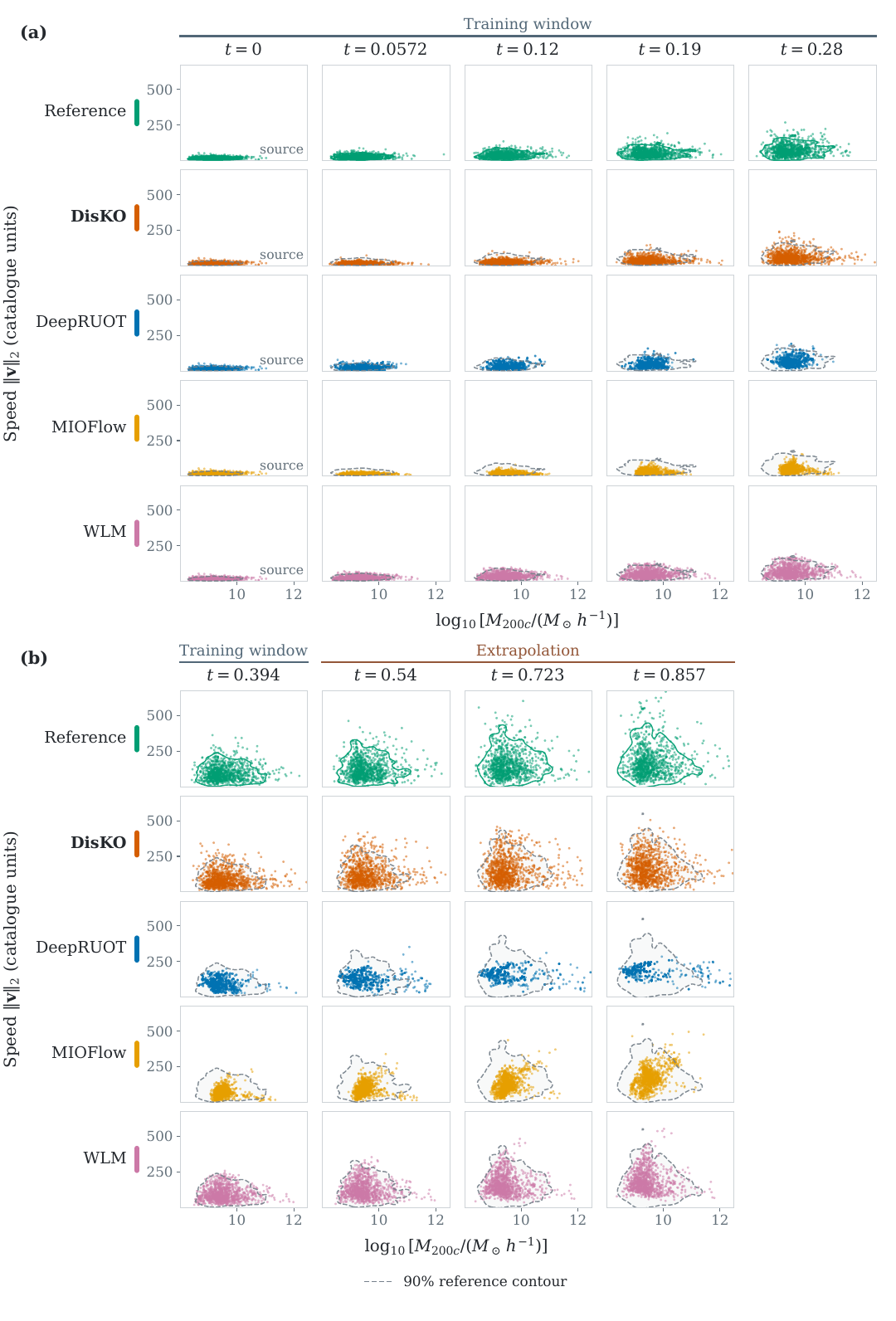}
    \caption{CAMELS prediction distributions over halo mass and speed, with coordinates $\log_{10}[M_{200c}/(M_\odot h^{-1})]$ and $\lVert v\rVert_2$ in catalogue units. Rows show the reference distribution and the four methods in the same order as Figure~\ref{fig:extrapolation-distribution-ou}. The bars above the columns identify training and extrapolation times. Dashed contours indicate the 90\% reference contour.}
    \label{fig:extrapolation-distribution-camels}
\end{figure}

\begin{figure}[p]
    \centering
    \includegraphics[width=\linewidth,height=0.86\textheight,keepaspectratio]{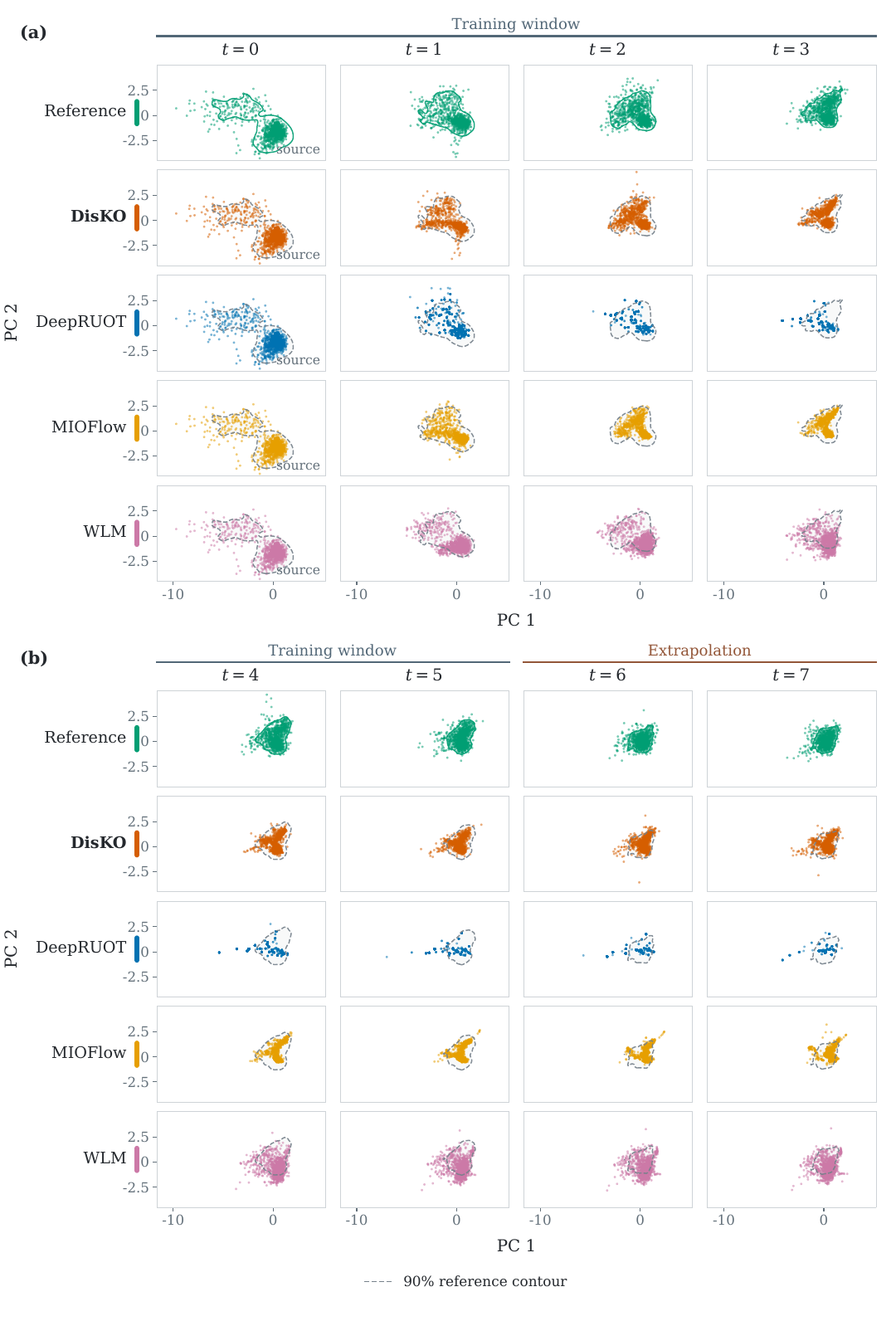}
    \caption{Pancreas prediction distributions in the first two principal component coordinates. Rows show the reference distribution, DisKO, DeepRUOT, MIOFlow, and WLM. Columns show the indicated times; the bars identify the training window and extrapolation at $t=6$ and $t=7$. Dashed contours indicate the 90\% reference contour.}
    \label{fig:extrapolation-distribution-pancreas}
\end{figure}
\FloatBarrier

\begin{figure}[!htbp]
    \centering
    \includegraphics[width=\linewidth,height=0.80\textheight,keepaspectratio]{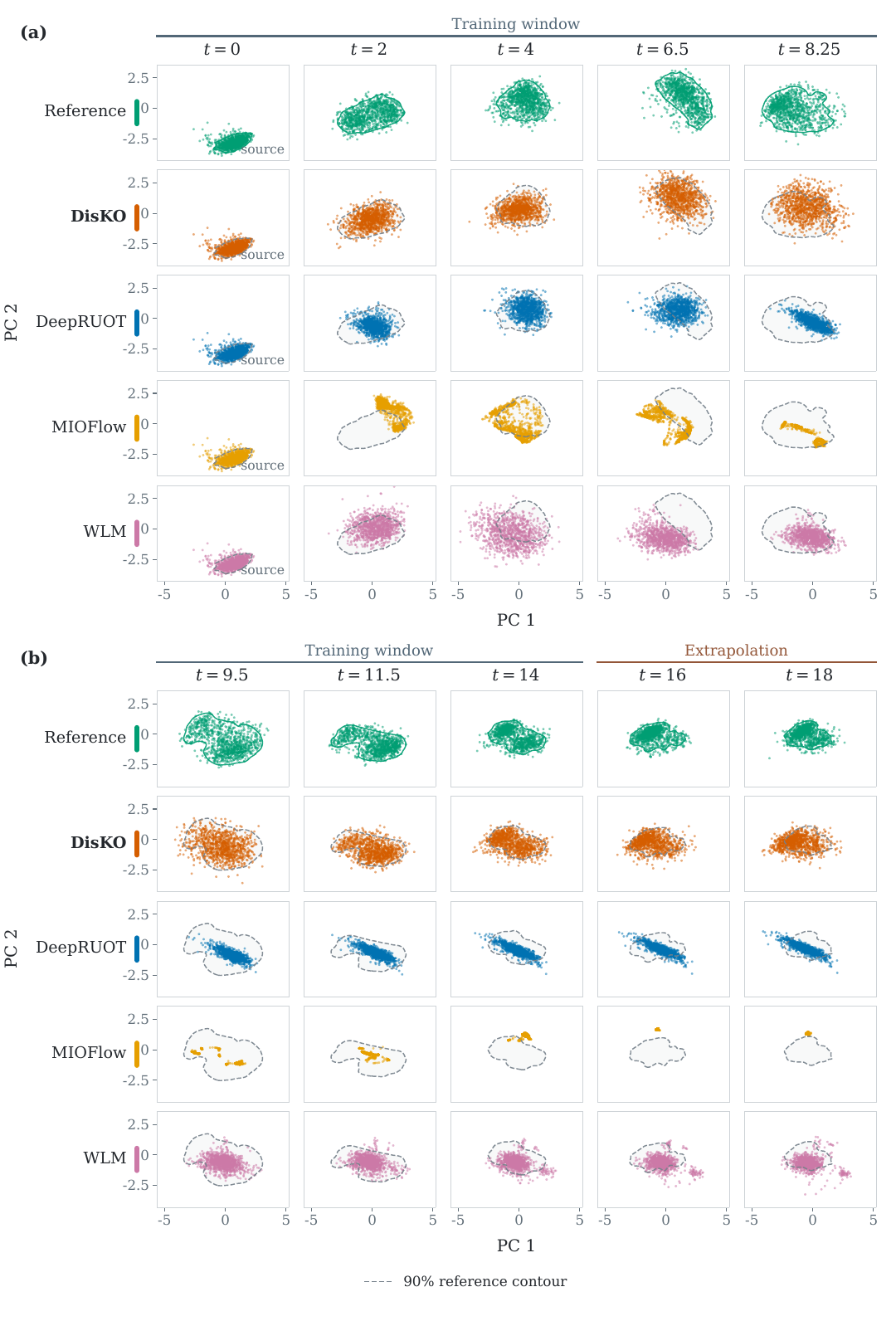}
    \caption{WOT iPSC prediction distributions in the first two principal component coordinates. Rows show the reference distribution, DisKO, DeepRUOT, MIOFlow, and WLM. Columns show the indicated times, with training and extrapolation intervals identified by the bars above them. Dashed contours indicate the 90\% reference contour.}
    \label{fig:extrapolation-distribution-wot-ipsc}
\end{figure}

\FloatBarrier
}

\clearpage

\begin{table}[!htbp]
\centering
\begingroup
\definecolor{resultFirst}{RGB}{190,25,35}
\definecolor{resultSecond}{RGB}{25,80,185}
\caption{Prediction within the training time window for training initial distributions and single sequences. Boids, Ocean, Pancreas, and WOT iPSC each contain a single population sequence and use only a temporal partition.
Means and standard deviations are rounded to three decimal places.
\textcolor{resultFirst}{Red} and \textcolor{resultSecond}{blue} mark the lowest and second lowest means before rounding.}
\label{tab:train-time-train-initial}
\fontsize{8}{10.5}\selectfont
\setlength{\tabcolsep}{1pt}
\renewcommand{\arraystretch}{1.05}
\setlength{\aboverulesep}{1pt}
\setlength{\belowrulesep}{1pt}
\newcommand{\resultmetric}[2]{\mbox{#1\textsubscript{\fontsize{6}{7}\selectfont$\pm$#2}}}
\begin{tabular*}{\linewidth}{@{}l@{\hspace{4pt}}c@{\hspace{4pt}\extracolsep{\fill}}ccccccc@{}}
\toprule
Method & Metric & OU & Duffing & Boids & Ocean & CAMELS & Pancreas & WOT iPSC \\
\midrule
\multirow{4}{*}[-3pt]{TrajectoryNet} & $W_1$ & \resultmetric{1.536}{0.026} & \resultmetric{0.788}{0.067} & \resultmetric{4.715}{0.131} & \resultmetric{1.508}{0.098} & \resultmetric{46.276}{0.186} & \resultmetric{20.423}{0.497} & \resultmetric{35.166}{1.357} \\
\arrayrulecolor{black!45}
\cmidrule[0.2pt](l{1pt}r{1pt}){3-9}
\arrayrulecolor{black}
 & SW1 & \resultmetric{0.956}{0.029} & \resultmetric{0.493}{0.053} & \resultmetric{2.758}{0.308} & \resultmetric{0.979}{0.076} & \resultmetric{11.758}{0.333} & \resultmetric{2.488}{0.201} & \resultmetric{2.847}{0.291} \\
\arrayrulecolor{black!45}
\cmidrule[0.2pt](l{1pt}r{1pt}){3-9}
\arrayrulecolor{black}
 & $W_2$ & \resultmetric{1.592}{0.027} & \resultmetric{0.843}{0.057} & \resultmetric{5.595}{0.135} & \resultmetric{1.513}{0.097} & \resultmetric{52.473}{0.268} & \resultmetric{20.984}{0.539} & \resultmetric{54.721}{2.050} \\
\arrayrulecolor{black!45}
\cmidrule[0.2pt](l{1pt}r{1pt}){3-9}
\arrayrulecolor{black}
 & MMD & \resultmetric{0.813}{0.015} & \resultmetric{0.621}{0.102} & \resultmetric{0.264}{0.044} & \resultmetric{1.002}{0.038} & \resultmetric{0.175}{0.008} & \resultmetric{0.329}{0.054} & \resultmetric{0.666}{0.099} \\
\specialrule{0.25pt}{2pt}{2pt}
\multirow{4}{*}[-3pt]{DeepRUOT} & $W_1$ & \resultmetric{1.221}{0.020} & \resultmetric{0.773}{0.145} & \resultmetric{4.342}{0.546} & \resultmetric{0.102}{0.031} & \resultmetric{20.784}{0.084} & \resultmetric{13.669}{1.096} & \textcolor{resultSecond}{\resultmetric{18.648}{1.053}} \\
\arrayrulecolor{black!45}
\cmidrule[0.2pt](l{1pt}r{1pt}){3-9}
\arrayrulecolor{black}
 & SW1 & \resultmetric{0.768}{0.024} & \resultmetric{0.484}{0.091} & \resultmetric{2.715}{0.358} & \resultmetric{0.064}{0.020} & \resultmetric{3.683}{0.103} & \resultmetric{1.340}{0.319} & \textcolor{resultSecond}{\resultmetric{1.441}{0.178}} \\
\arrayrulecolor{black!45}
\cmidrule[0.2pt](l{1pt}r{1pt}){3-9}
\arrayrulecolor{black}
 & $W_2$ & \resultmetric{1.282}{0.021} & \resultmetric{0.827}{0.147} & \resultmetric{4.816}{0.488} & \resultmetric{0.117}{0.028} & \resultmetric{25.479}{0.135} & \resultmetric{14.714}{2.030} & \textcolor{resultSecond}{\resultmetric{18.847}{1.069}} \\
\arrayrulecolor{black!45}
\cmidrule[0.2pt](l{1pt}r{1pt}){3-9}
\arrayrulecolor{black}
 & MMD & \resultmetric{0.638}{0.013} & \resultmetric{0.510}{0.137} & \resultmetric{0.245}{0.047} & \resultmetric{0.305}{0.118} & \resultmetric{0.016}{0.001} & \resultmetric{0.079}{0.043} & \resultmetric{0.157}{0.041} \\
\specialrule{0.25pt}{2pt}{2pt}
\multirow{4}{*}[-3pt]{VGFM} & $W_1$ & \resultmetric{1.184}{0.099} & \resultmetric{0.431}{0.184} & \resultmetric{2.736}{0.667} & \textcolor{resultSecond}{\resultmetric{0.055}{0.006}} & \resultmetric{29.751}{1.353} & \resultmetric{12.183}{0.845} & \resultmetric{33.836}{1.578} \\
\arrayrulecolor{black!45}
\cmidrule[0.2pt](l{1pt}r{1pt}){3-9}
\arrayrulecolor{black}
 & SW1 & \resultmetric{0.733}{0.064} & \resultmetric{0.258}{0.114} & \resultmetric{1.601}{0.510} & \textcolor{resultSecond}{\resultmetric{0.032}{0.003}} & \resultmetric{5.864}{0.719} & \resultmetric{0.796}{0.072} & \resultmetric{2.591}{0.958} \\
\arrayrulecolor{black!45}
\cmidrule[0.2pt](l{1pt}r{1pt}){3-9}
\arrayrulecolor{black}
 & $W_2$ & \resultmetric{1.237}{0.099} & \resultmetric{0.542}{0.201} & \resultmetric{3.270}{0.746} & \textcolor{resultSecond}{\resultmetric{0.066}{0.006}} & \resultmetric{35.337}{1.433} & \resultmetric{14.099}{1.294} & \resultmetric{190.213}{1.063} \\
\arrayrulecolor{black!45}
\cmidrule[0.2pt](l{1pt}r{1pt}){3-9}
\arrayrulecolor{black}
 & MMD & \resultmetric{0.582}{0.042} & \resultmetric{0.108}{0.044} & \resultmetric{0.084}{0.047} & \textcolor{resultSecond}{\resultmetric{0.072}{0.012}} & \resultmetric{0.028}{0.020} & \resultmetric{0.033}{0.014} & \textcolor{resultSecond}{\resultmetric{0.077}{0.029}} \\
\specialrule{0.25pt}{2pt}{2pt}
\multirow{4}{*}[-3pt]{MIOFlow} & $W_1$ & \resultmetric{1.462}{0.047} & \resultmetric{0.412}{0.078} & \resultmetric{4.703}{0.988} & \resultmetric{0.099}{0.021} & \resultmetric{32.147}{0.733} & \textcolor{resultSecond}{\resultmetric{7.765}{0.166}} & \resultmetric{25.994}{0.673} \\
\arrayrulecolor{black!45}
\cmidrule[0.2pt](l{1pt}r{1pt}){3-9}
\arrayrulecolor{black}
 & SW1 & \resultmetric{0.911}{0.030} & \resultmetric{0.252}{0.051} & \resultmetric{2.811}{0.404} & \resultmetric{0.063}{0.014} & \resultmetric{7.527}{0.215} & \textcolor{resultSecond}{\resultmetric{0.462}{0.038}} & \resultmetric{2.533}{0.108} \\
\arrayrulecolor{black!45}
\cmidrule[0.2pt](l{1pt}r{1pt}){3-9}
\arrayrulecolor{black}
 & $W_2$ & \resultmetric{1.522}{0.050} & \resultmetric{0.467}{0.085} & \resultmetric{5.723}{0.579} & \resultmetric{0.116}{0.024} & \resultmetric{37.471}{0.599} & \textcolor{resultSecond}{\resultmetric{8.321}{0.187}} & \resultmetric{26.271}{0.691} \\
\arrayrulecolor{black!45}
\cmidrule[0.2pt](l{1pt}r{1pt}){3-9}
\arrayrulecolor{black}
 & MMD & \resultmetric{0.692}{0.007} & \resultmetric{0.188}{0.066} & \resultmetric{0.156}{0.063} & \resultmetric{0.278}{0.085} & \resultmetric{0.077}{0.002} & \textcolor{resultSecond}{\resultmetric{0.009}{0.002}} & \resultmetric{0.358}{0.035} \\
\specialrule{0.25pt}{2pt}{2pt}
\multirow{4}{*}[-3pt]{scNODE} & $W_1$ & \resultmetric{1.191}{0.043} & \resultmetric{0.834}{0.097} & \resultmetric{4.796}{0.057} & \resultmetric{0.782}{0.105} & \resultmetric{53.869}{2.635} & \resultmetric{20.511}{1.862} & \resultmetric{43.246}{1.129} \\
\arrayrulecolor{black!45}
\cmidrule[0.2pt](l{1pt}r{1pt}){3-9}
\arrayrulecolor{black}
 & SW1 & \resultmetric{0.750}{0.028} & \resultmetric{0.520}{0.072} & \resultmetric{2.750}{0.033} & \resultmetric{0.508}{0.063} & \resultmetric{13.103}{0.891} & \resultmetric{2.505}{0.509} & \resultmetric{4.715}{0.759} \\
\arrayrulecolor{black!45}
\cmidrule[0.2pt](l{1pt}r{1pt}){3-9}
\arrayrulecolor{black}
 & $W_2$ & \resultmetric{1.253}{0.042} & \resultmetric{0.888}{0.078} & \resultmetric{5.577}{0.107} & \resultmetric{0.787}{0.104} & \resultmetric{60.330}{2.823} & \resultmetric{21.064}{2.162} & \resultmetric{43.377}{1.057} \\
\arrayrulecolor{black!45}
\cmidrule[0.2pt](l{1pt}r{1pt}){3-9}
\arrayrulecolor{black}
 & MMD & \resultmetric{0.582}{0.014} & \resultmetric{0.449}{0.089} & \resultmetric{0.091}{0.006} & \resultmetric{0.971}{0.062} & \resultmetric{0.175}{0.026} & \resultmetric{0.336}{0.064} & \resultmetric{0.707}{0.214} \\
\specialrule{0.25pt}{2pt}{2pt}
\multirow{4}{*}[-3pt]{PRESCIENT} & $W_1$ & \resultmetric{1.452}{0.024} & \resultmetric{0.259}{0.004} & \resultmetric{3.100}{0.017} & \resultmetric{0.962}{0.012} & \resultmetric{45.111}{0.095} & \resultmetric{11.523}{0.217} & \resultmetric{25.992}{0.046} \\
\arrayrulecolor{black!45}
\cmidrule[0.2pt](l{1pt}r{1pt}){3-9}
\arrayrulecolor{black}
 & SW1 & \resultmetric{0.902}{0.015} & \resultmetric{0.141}{0.003} & \resultmetric{1.871}{0.018} & \resultmetric{0.599}{0.007} & \resultmetric{11.546}{0.024} & \resultmetric{0.989}{0.039} & \resultmetric{2.379}{0.008} \\
\arrayrulecolor{black!45}
\cmidrule[0.2pt](l{1pt}r{1pt}){3-9}
\arrayrulecolor{black}
 & $W_2$ & \resultmetric{1.502}{0.025} & \resultmetric{0.332}{0.003} & \resultmetric{3.318}{0.022} & \resultmetric{0.976}{0.012} & \resultmetric{51.258}{0.090} & \resultmetric{12.234}{0.137} & \resultmetric{26.084}{0.046} \\
\arrayrulecolor{black!45}
\cmidrule[0.2pt](l{1pt}r{1pt}){3-9}
\arrayrulecolor{black}
 & MMD & \resultmetric{0.683}{0.002} & \resultmetric{0.074}{0.002} & \resultmetric{0.123}{0.003} & \resultmetric{0.894}{0.004} & \resultmetric{0.169}{0.001} & \resultmetric{0.079}{0.008} & \resultmetric{0.373}{0.002} \\
\specialrule{0.25pt}{2pt}{2pt}
\multirow{4}{*}[-3pt]{JKOnet*} & $W_1$ & \resultmetric{1.655}{0.027} & \resultmetric{0.405}{0.035} & \resultmetric{3.080}{0.086} & \resultmetric{1.838}{0.305} & \resultmetric{37.951}{0.151} & \resultmetric{47.086}{1.003} & \resultmetric{29.582}{1.154} \\
\arrayrulecolor{black!45}
\cmidrule[0.2pt](l{1pt}r{1pt}){3-9}
\arrayrulecolor{black}
 & SW1 & \resultmetric{1.028}{0.031} & \resultmetric{0.221}{0.024} & \resultmetric{1.844}{0.047} & \resultmetric{1.166}{0.196} & \resultmetric{9.282}{0.265} & \resultmetric{7.248}{0.599} & \resultmetric{2.561}{0.258} \\
\arrayrulecolor{black!45}
\cmidrule[0.2pt](l{1pt}r{1pt}){3-9}
\arrayrulecolor{black}
 & $W_2$ & \resultmetric{1.707}{0.028} & \resultmetric{0.723}{0.048} & \resultmetric{3.363}{0.096} & \resultmetric{1.946}{0.374} & \resultmetric{51.292}{0.274} & \resultmetric{59.232}{1.185} & \resultmetric{43.680}{1.353} \\
\arrayrulecolor{black!45}
\cmidrule[0.2pt](l{1pt}r{1pt}){3-9}
\arrayrulecolor{black}
 & MMD & \resultmetric{0.703}{0.014} & \resultmetric{0.047}{0.009} & \resultmetric{0.093}{0.001} & \resultmetric{0.913}{0.050} & \resultmetric{0.078}{0.004} & \resultmetric{0.290}{0.048} & \resultmetric{0.339}{0.051} \\
\specialrule{0.25pt}{2pt}{2pt}
\multirow{4}{*}[-3pt]{WLM} & $W_1$ & \textcolor{resultSecond}{\resultmetric{0.709}{0.006}} & \textcolor{resultSecond}{\resultmetric{0.116}{0.004}} & \textcolor{resultSecond}{\resultmetric{0.934}{0.059}} & \resultmetric{0.443}{0.118} & \textcolor{resultFirst}{\resultmetric{16.601}{0.066}} & \resultmetric{12.053}{0.330} & \resultmetric{23.056}{1.109} \\
\arrayrulecolor{black!45}
\cmidrule[0.2pt](l{1pt}r{1pt}){3-9}
\arrayrulecolor{black}
 & SW1 & \textcolor{resultSecond}{\resultmetric{0.428}{0.003}} & \textcolor{resultSecond}{\resultmetric{0.042}{0.002}} & \textcolor{resultSecond}{\resultmetric{0.219}{0.064}} & \resultmetric{0.284}{0.262} & \textcolor{resultFirst}{\resultmetric{1.251}{0.023}} & \resultmetric{0.840}{0.057} & \resultmetric{1.680}{0.129} \\
\arrayrulecolor{black!45}
\cmidrule[0.2pt](l{1pt}r{1pt}){3-9}
\arrayrulecolor{black}
 & $W_2$ & \textcolor{resultSecond}{\resultmetric{0.762}{0.006}} & \textcolor{resultSecond}{\resultmetric{0.165}{0.006}} & \textcolor{resultSecond}{\resultmetric{1.300}{0.072}} & \resultmetric{0.824}{0.435} & \textcolor{resultFirst}{\resultmetric{20.246}{0.106}} & \resultmetric{12.852}{0.332} & \resultmetric{23.401}{1.094} \\
\arrayrulecolor{black!45}
\cmidrule[0.2pt](l{1pt}r{1pt}){3-9}
\arrayrulecolor{black}
 & MMD & \textcolor{resultSecond}{\resultmetric{0.350}{0.006}} & \textcolor{resultSecond}{\resultmetric{0.002}{0.000}} & \textcolor{resultSecond}{\resultmetric{0.002}{0.001}} & \resultmetric{0.480}{0.241} & \textcolor{resultFirst}{\resultmetric{0.002}{0.000}} & \resultmetric{0.048}{0.006} & \resultmetric{0.186}{0.023} \\
\midrule
\multirow{4}{*}[-3pt]{\textbf{DisKO}} & $W_1$ & \textcolor{resultFirst}{\resultmetric{0.279}{0.005}} & \textcolor{resultFirst}{\resultmetric{0.097}{0.001}} & \textcolor{resultFirst}{\resultmetric{0.816}{0.011}} & \textcolor{resultFirst}{\resultmetric{0.029}{0.004}} & \textcolor{resultSecond}{\resultmetric{16.634}{0.035}} & \textcolor{resultFirst}{\resultmetric{7.268}{0.046}} & \textcolor{resultFirst}{\resultmetric{16.700}{0.217}} \\
\arrayrulecolor{black!45}
\cmidrule[0.2pt](l{1pt}r{1pt}){3-9}
\arrayrulecolor{black}
 & SW1 & \textcolor{resultFirst}{\resultmetric{0.135}{0.004}} & \textcolor{resultFirst}{\resultmetric{0.033}{0.001}} & \textcolor{resultFirst}{\resultmetric{0.176}{0.016}} & \textcolor{resultFirst}{\resultmetric{0.015}{0.003}} & \textcolor{resultSecond}{\resultmetric{1.312}{0.072}} & \textcolor{resultFirst}{\resultmetric{0.345}{0.011}} & \textcolor{resultFirst}{\resultmetric{0.614}{0.073}} \\
\arrayrulecolor{black!45}
\cmidrule[0.2pt](l{1pt}r{1pt}){3-9}
\arrayrulecolor{black}
 & $W_2$ & \textcolor{resultFirst}{\resultmetric{0.337}{0.004}} & \textcolor{resultFirst}{\resultmetric{0.139}{0.001}} & \textcolor{resultFirst}{\resultmetric{1.261}{0.014}} & \textcolor{resultFirst}{\resultmetric{0.034}{0.006}} & \textcolor{resultSecond}{\resultmetric{20.622}{0.065}} & \textcolor{resultFirst}{\resultmetric{7.737}{0.037}} & \textcolor{resultFirst}{\resultmetric{16.874}{0.203}} \\
\arrayrulecolor{black!45}
\cmidrule[0.2pt](l{1pt}r{1pt}){3-9}
\arrayrulecolor{black}
 & MMD & \textcolor{resultFirst}{\resultmetric{0.025}{0.002}} & \textcolor{resultFirst}{\resultmetric{0.002}{0.000}} & \textcolor{resultFirst}{\resultmetric{0.001}{0.000}} & \textcolor{resultFirst}{\resultmetric{0.017}{0.011}} & \textcolor{resultSecond}{\resultmetric{0.002}{0.000}} & \textcolor{resultFirst}{\resultmetric{0.006}{0.000}} & \textcolor{resultFirst}{\resultmetric{0.032}{0.005}} \\
\bottomrule
\end{tabular*}
\endgroup
\end{table}

\clearpage
\begin{table}[!htbp]
\centering
\begingroup
\definecolor{resultFirst}{RGB}{190,25,35}
\definecolor{resultSecond}{RGB}{25,80,185}
\caption{Prediction within the training time window for test initial distributions. A hyphen (-) denotes the absence of a separate set of test initial distributions for a single sequence.
Means and standard deviations are rounded to three decimal places.
\textcolor{resultFirst}{Red} and \textcolor{resultSecond}{blue} mark the lowest and second lowest means before rounding.}
\label{tab:train-time-test-initial}
\fontsize{8}{10.5}\selectfont
\setlength{\tabcolsep}{1pt}
\renewcommand{\arraystretch}{1.05}
\setlength{\aboverulesep}{1pt}
\setlength{\belowrulesep}{1pt}
\newcommand{\resultmetric}[2]{\mbox{#1\textsubscript{\fontsize{6}{7}\selectfont$\pm$#2}}}
\begin{tabular*}{\linewidth}{@{}l@{\hspace{4pt}}c@{\hspace{4pt}\extracolsep{\fill}}ccccccc@{}}
\toprule
Method & Metric & OU & Duffing & Boids & Ocean & CAMELS & Pancreas & WOT iPSC \\
\midrule
\multirow{4}{*}[-3pt]{TrajectoryNet} & $W_1$ & \resultmetric{1.585}{0.044} & \resultmetric{0.751}{0.064} & \textcolor{black!55}{-} & \textcolor{black!55}{-} & \resultmetric{45.835}{0.506} & \textcolor{black!55}{-} & \textcolor{black!55}{-} \\
\arrayrulecolor{black!45}
\cmidrule[0.2pt](l{1pt}r{1pt}){3-9}
\arrayrulecolor{black}
 & SW1 & \resultmetric{0.985}{0.034} & \resultmetric{0.469}{0.050} & \textcolor{black!55}{-} & \textcolor{black!55}{-} & \resultmetric{11.567}{0.342} & \textcolor{black!55}{-} & \textcolor{black!55}{-} \\
\arrayrulecolor{black!45}
\cmidrule[0.2pt](l{1pt}r{1pt}){3-9}
\arrayrulecolor{black}
 & $W_2$ & \resultmetric{1.648}{0.031} & \resultmetric{0.820}{0.055} & \textcolor{black!55}{-} & \textcolor{black!55}{-} & \resultmetric{53.090}{0.311} & \textcolor{black!55}{-} & \textcolor{black!55}{-} \\
\arrayrulecolor{black!45}
\cmidrule[0.2pt](l{1pt}r{1pt}){3-9}
\arrayrulecolor{black}
 & MMD & \resultmetric{0.706}{0.022} & \resultmetric{0.499}{0.098} & \textcolor{black!55}{-} & \textcolor{black!55}{-} & \resultmetric{0.186}{0.015} & \textcolor{black!55}{-} & \textcolor{black!55}{-} \\
\specialrule{0.25pt}{2pt}{2pt}
\multirow{4}{*}[-3pt]{DeepRUOT} & $W_1$ & \resultmetric{1.263}{0.034} & \resultmetric{0.772}{0.146} & \textcolor{black!55}{-} & \textcolor{black!55}{-} & \textcolor{resultSecond}{\resultmetric{21.261}{0.725}} & \textcolor{black!55}{-} & \textcolor{black!55}{-} \\
\arrayrulecolor{black!45}
\cmidrule[0.2pt](l{1pt}r{1pt}){3-9}
\arrayrulecolor{black}
 & SW1 & \resultmetric{0.797}{0.027} & \resultmetric{0.484}{0.092} & \textcolor{black!55}{-} & \textcolor{black!55}{-} & \textcolor{resultSecond}{\resultmetric{3.622}{0.106}} & \textcolor{black!55}{-} & \textcolor{black!55}{-} \\
\arrayrulecolor{black!45}
\cmidrule[0.2pt](l{1pt}r{1pt}){3-9}
\arrayrulecolor{black}
 & $W_2$ & \resultmetric{1.322}{0.025} & \resultmetric{0.827}{0.149} & \textcolor{black!55}{-} & \textcolor{black!55}{-} & \textcolor{resultSecond}{\resultmetric{27.025}{0.165}} & \textcolor{black!55}{-} & \textcolor{black!55}{-} \\
\arrayrulecolor{black!45}
\cmidrule[0.2pt](l{1pt}r{1pt}){3-9}
\arrayrulecolor{black}
 & MMD & \resultmetric{0.625}{0.020} & \resultmetric{0.510}{0.137} & \textcolor{black!55}{-} & \textcolor{black!55}{-} & \textcolor{resultSecond}{\resultmetric{0.014}{0.001}} & \textcolor{black!55}{-} & \textcolor{black!55}{-} \\
\specialrule{0.25pt}{2pt}{2pt}
\multirow{4}{*}[-3pt]{VGFM} & $W_1$ & \resultmetric{1.226}{0.127} & \resultmetric{0.429}{0.183} & \textcolor{black!55}{-} & \textcolor{black!55}{-} & \resultmetric{31.046}{1.957} & \textcolor{black!55}{-} & \textcolor{black!55}{-} \\
\arrayrulecolor{black!45}
\cmidrule[0.2pt](l{1pt}r{1pt}){3-9}
\arrayrulecolor{black}
 & SW1 & \resultmetric{0.758}{0.082} & \resultmetric{0.256}{0.113} & \textcolor{black!55}{-} & \textcolor{black!55}{-} & \resultmetric{6.050}{1.064} & \textcolor{black!55}{-} & \textcolor{black!55}{-} \\
\arrayrulecolor{black!45}
\cmidrule[0.2pt](l{1pt}r{1pt}){3-9}
\arrayrulecolor{black}
 & $W_2$ & \resultmetric{1.281}{0.127} & \resultmetric{0.536}{0.200} & \textcolor{black!55}{-} & \textcolor{black!55}{-} & \resultmetric{37.239}{2.114} & \textcolor{black!55}{-} & \textcolor{black!55}{-} \\
\arrayrulecolor{black!45}
\cmidrule[0.2pt](l{1pt}r{1pt}){3-9}
\arrayrulecolor{black}
 & MMD & \resultmetric{0.581}{0.060} & \resultmetric{0.108}{0.044} & \textcolor{black!55}{-} & \textcolor{black!55}{-} & \resultmetric{0.031}{0.026} & \textcolor{black!55}{-} & \textcolor{black!55}{-} \\
\specialrule{0.25pt}{2pt}{2pt}
\multirow{4}{*}[-3pt]{MIOFlow} & $W_1$ & \resultmetric{1.547}{0.088} & \resultmetric{0.411}{0.078} & \textcolor{black!55}{-} & \textcolor{black!55}{-} & \resultmetric{32.735}{0.356} & \textcolor{black!55}{-} & \textcolor{black!55}{-} \\
\arrayrulecolor{black!45}
\cmidrule[0.2pt](l{1pt}r{1pt}){3-9}
\arrayrulecolor{black}
 & SW1 & \resultmetric{0.963}{0.057} & \resultmetric{0.251}{0.051} & \textcolor{black!55}{-} & \textcolor{black!55}{-} & \resultmetric{7.662}{0.227} & \textcolor{black!55}{-} & \textcolor{black!55}{-} \\
\arrayrulecolor{black!45}
\cmidrule[0.2pt](l{1pt}r{1pt}){3-9}
\arrayrulecolor{black}
 & $W_2$ & \resultmetric{1.613}{0.099} & \resultmetric{0.466}{0.084} & \textcolor{black!55}{-} & \textcolor{black!55}{-} & \resultmetric{39.520}{0.231} & \textcolor{black!55}{-} & \textcolor{black!55}{-} \\
\arrayrulecolor{black!45}
\cmidrule[0.2pt](l{1pt}r{1pt}){3-9}
\arrayrulecolor{black}
 & MMD & \resultmetric{0.707}{0.021} & \resultmetric{0.188}{0.067} & \textcolor{black!55}{-} & \textcolor{black!55}{-} & \resultmetric{0.081}{0.003} & \textcolor{black!55}{-} & \textcolor{black!55}{-} \\
\specialrule{0.25pt}{2pt}{2pt}
\multirow{4}{*}[-3pt]{scNODE} & $W_1$ & \resultmetric{1.266}{0.046} & \resultmetric{0.834}{0.096} & \textcolor{black!55}{-} & \textcolor{black!55}{-} & \resultmetric{54.432}{2.139} & \textcolor{black!55}{-} & \textcolor{black!55}{-} \\
\arrayrulecolor{black!45}
\cmidrule[0.2pt](l{1pt}r{1pt}){3-9}
\arrayrulecolor{black}
 & SW1 & \resultmetric{0.800}{0.027} & \resultmetric{0.521}{0.071} & \textcolor{black!55}{-} & \textcolor{black!55}{-} & \resultmetric{13.294}{0.570} & \textcolor{black!55}{-} & \textcolor{black!55}{-} \\
\arrayrulecolor{black!45}
\cmidrule[0.2pt](l{1pt}r{1pt}){3-9}
\arrayrulecolor{black}
 & $W_2$ & \resultmetric{1.325}{0.046} & \resultmetric{0.888}{0.077} & \textcolor{black!55}{-} & \textcolor{black!55}{-} & \resultmetric{61.763}{2.481} & \textcolor{black!55}{-} & \textcolor{black!55}{-} \\
\arrayrulecolor{black!45}
\cmidrule[0.2pt](l{1pt}r{1pt}){3-9}
\arrayrulecolor{black}
 & MMD & \resultmetric{0.589}{0.019} & \resultmetric{0.450}{0.087} & \textcolor{black!55}{-} & \textcolor{black!55}{-} & \resultmetric{0.190}{0.019} & \textcolor{black!55}{-} & \textcolor{black!55}{-} \\
\specialrule{0.25pt}{2pt}{2pt}
\multirow{4}{*}[-3pt]{PRESCIENT} & $W_1$ & \resultmetric{1.542}{0.027} & \resultmetric{0.258}{0.004} & \textcolor{black!55}{-} & \textcolor{black!55}{-} & \resultmetric{44.631}{0.115} & \textcolor{black!55}{-} & \textcolor{black!55}{-} \\
\arrayrulecolor{black!45}
\cmidrule[0.2pt](l{1pt}r{1pt}){3-9}
\arrayrulecolor{black}
 & SW1 & \resultmetric{0.956}{0.016} & \resultmetric{0.141}{0.004} & \textcolor{black!55}{-} & \textcolor{black!55}{-} & \resultmetric{11.321}{0.027} & \textcolor{black!55}{-} & \textcolor{black!55}{-} \\
\arrayrulecolor{black!45}
\cmidrule[0.2pt](l{1pt}r{1pt}){3-9}
\arrayrulecolor{black}
 & $W_2$ & \resultmetric{1.595}{0.027} & \resultmetric{0.331}{0.003} & \textcolor{black!55}{-} & \textcolor{black!55}{-} & \resultmetric{51.860}{0.128} & \textcolor{black!55}{-} & \textcolor{black!55}{-} \\
\arrayrulecolor{black!45}
\cmidrule[0.2pt](l{1pt}r{1pt}){3-9}
\arrayrulecolor{black}
 & MMD & \resultmetric{0.690}{0.003} & \resultmetric{0.074}{0.003} & \textcolor{black!55}{-} & \textcolor{black!55}{-} & \resultmetric{0.163}{0.001} & \textcolor{black!55}{-} & \textcolor{black!55}{-} \\
\specialrule{0.25pt}{2pt}{2pt}
\multirow{4}{*}[-3pt]{JKOnet*} & $W_1$ & \resultmetric{1.728}{0.046} & \resultmetric{0.401}{0.035} & \textcolor{black!55}{-} & \textcolor{black!55}{-} & \resultmetric{39.020}{0.434} & \textcolor{black!55}{-} & \textcolor{black!55}{-} \\
\arrayrulecolor{black!45}
\cmidrule[0.2pt](l{1pt}r{1pt}){3-9}
\arrayrulecolor{black}
 & SW1 & \resultmetric{1.071}{0.037} & \resultmetric{0.218}{0.023} & \textcolor{black!55}{-} & \textcolor{black!55}{-} & \resultmetric{9.559}{0.282} & \textcolor{black!55}{-} & \textcolor{black!55}{-} \\
\arrayrulecolor{black!45}
\cmidrule[0.2pt](l{1pt}r{1pt}){3-9}
\arrayrulecolor{black}
 & $W_2$ & \resultmetric{1.783}{0.034} & \resultmetric{0.703}{0.047} & \textcolor{black!55}{-} & \textcolor{black!55}{-} & \resultmetric{52.598}{0.322} & \textcolor{black!55}{-} & \textcolor{black!55}{-} \\
\arrayrulecolor{black!45}
\cmidrule[0.2pt](l{1pt}r{1pt}){3-9}
\arrayrulecolor{black}
 & MMD & \resultmetric{0.696}{0.022} & \resultmetric{0.047}{0.009} & \textcolor{black!55}{-} & \textcolor{black!55}{-} & \resultmetric{0.074}{0.006} & \textcolor{black!55}{-} & \textcolor{black!55}{-} \\
\specialrule{0.25pt}{2pt}{2pt}
\multirow{4}{*}[-3pt]{WLM} & $W_1$ & \textcolor{resultSecond}{\resultmetric{0.843}{0.006}} & \textcolor{resultSecond}{\resultmetric{0.116}{0.004}} & \textcolor{black!55}{-} & \textcolor{black!55}{-} & \textcolor{resultFirst}{\resultmetric{18.274}{0.102}} & \textcolor{black!55}{-} & \textcolor{black!55}{-} \\
\arrayrulecolor{black!45}
\cmidrule[0.2pt](l{1pt}r{1pt}){3-9}
\arrayrulecolor{black}
 & SW1 & \textcolor{resultSecond}{\resultmetric{0.510}{0.003}} & \textcolor{resultSecond}{\resultmetric{0.041}{0.002}} & \textcolor{black!55}{-} & \textcolor{black!55}{-} & \textcolor{resultFirst}{\resultmetric{1.895}{0.055}} & \textcolor{black!55}{-} & \textcolor{black!55}{-} \\
\arrayrulecolor{black!45}
\cmidrule[0.2pt](l{1pt}r{1pt}){3-9}
\arrayrulecolor{black}
 & $W_2$ & \textcolor{resultSecond}{\resultmetric{0.900}{0.006}} & \textcolor{resultSecond}{\resultmetric{0.165}{0.006}} & \textcolor{black!55}{-} & \textcolor{black!55}{-} & \textcolor{resultFirst}{\resultmetric{23.334}{0.140}} & \textcolor{black!55}{-} & \textcolor{black!55}{-} \\
\arrayrulecolor{black!45}
\cmidrule[0.2pt](l{1pt}r{1pt}){3-9}
\arrayrulecolor{black}
 & MMD & \textcolor{resultSecond}{\resultmetric{0.397}{0.004}} & \textcolor{resultSecond}{\resultmetric{0.002}{0.000}} & \textcolor{black!55}{-} & \textcolor{black!55}{-} & \textcolor{resultFirst}{\resultmetric{0.004}{0.000}} & \textcolor{black!55}{-} & \textcolor{black!55}{-} \\
\midrule
\multirow{4}{*}[-3pt]{\textbf{DisKO}} & $W_1$ & \textcolor{resultFirst}{\resultmetric{0.315}{0.009}} & \textcolor{resultFirst}{\resultmetric{0.098}{0.001}} & \textcolor{black!55}{-} & \textcolor{black!55}{-} & \resultmetric{22.710}{0.466} & \textcolor{black!55}{-} & \textcolor{black!55}{-} \\
\arrayrulecolor{black!45}
\cmidrule[0.2pt](l{1pt}r{1pt}){3-9}
\arrayrulecolor{black}
 & SW1 & \textcolor{resultFirst}{\resultmetric{0.154}{0.006}} & \textcolor{resultFirst}{\resultmetric{0.033}{0.001}} & \textcolor{black!55}{-} & \textcolor{black!55}{-} & \resultmetric{4.100}{0.239} & \textcolor{black!55}{-} & \textcolor{black!55}{-} \\
\arrayrulecolor{black!45}
\cmidrule[0.2pt](l{1pt}r{1pt}){3-9}
\arrayrulecolor{black}
 & $W_2$ & \textcolor{resultFirst}{\resultmetric{0.372}{0.007}} & \textcolor{resultFirst}{\resultmetric{0.139}{0.001}} & \textcolor{black!55}{-} & \textcolor{black!55}{-} & \resultmetric{28.157}{0.463} & \textcolor{black!55}{-} & \textcolor{black!55}{-} \\
\arrayrulecolor{black!45}
\cmidrule[0.2pt](l{1pt}r{1pt}){3-9}
\arrayrulecolor{black}
 & MMD & \textcolor{resultFirst}{\resultmetric{0.031}{0.003}} & \textcolor{resultFirst}{\resultmetric{0.002}{0.000}} & \textcolor{black!55}{-} & \textcolor{black!55}{-} & \resultmetric{0.024}{0.003} & \textcolor{black!55}{-} & \textcolor{black!55}{-} \\
\bottomrule
\end{tabular*}
\endgroup
\end{table}

\clearpage
\begin{table}[!htbp]
\centering
\begingroup
\definecolor{resultFirst}{RGB}{190,25,35}
\definecolor{resultSecond}{RGB}{25,80,185}
\caption{Temporal extrapolation from training initial distributions in datasets with multiple sequences. Hyphens (-) mark single sequences, whose temporal extrapolation results are reported in Table~\ref{tab:test-time-test-initial}.
Means and standard deviations are rounded to three decimal places.
\textcolor{resultFirst}{Red} and \textcolor{resultSecond}{blue} mark the lowest and second lowest means before rounding.}
\label{tab:test-time-train-initial}
\fontsize{8}{10.5}\selectfont
\setlength{\tabcolsep}{1pt}
\renewcommand{\arraystretch}{1.05}
\setlength{\aboverulesep}{1pt}
\setlength{\belowrulesep}{1pt}
\newcommand{\resultmetric}[2]{\mbox{#1\textsubscript{\fontsize{6}{7}\selectfont$\pm$#2}}}
\begin{tabular*}{\linewidth}{@{}l@{\hspace{4pt}}c@{\hspace{4pt}\extracolsep{\fill}}ccccccc@{}}
\toprule
Method & Metric & OU & Duffing & Boids & Ocean & CAMELS & Pancreas & WOT iPSC \\
\midrule
\multirow{4}{*}[-3pt]{TrajectoryNet} & $W_1$ & \resultmetric{1.279}{0.055} & \resultmetric{12.544}{1.069} & \textcolor{black!55}{-} & \textcolor{black!55}{-} & \resultmetric{145.995}{1.209} & \textcolor{black!55}{-} & \textcolor{black!55}{-} \\
\arrayrulecolor{black!45}
\cmidrule[0.2pt](l{1pt}r{1pt}){3-9}
\arrayrulecolor{black}
 & SW1 & \resultmetric{0.800}{0.033} & \resultmetric{7.876}{0.975} & \textcolor{black!55}{-} & \textcolor{black!55}{-} & \resultmetric{36.488}{1.768} & \textcolor{black!55}{-} & \textcolor{black!55}{-} \\
\arrayrulecolor{black!45}
\cmidrule[0.2pt](l{1pt}r{1pt}){3-9}
\arrayrulecolor{black}
 & $W_2$ & \resultmetric{1.339}{0.052} & \resultmetric{26.279}{1.501} & \textcolor{black!55}{-} & \textcolor{black!55}{-} & \resultmetric{167.079}{1.926} & \textcolor{black!55}{-} & \textcolor{black!55}{-} \\
\arrayrulecolor{black!45}
\cmidrule[0.2pt](l{1pt}r{1pt}){3-9}
\arrayrulecolor{black}
 & MMD & \resultmetric{0.656}{0.019} & \resultmetric{0.523}{0.138} & \textcolor{black!55}{-} & \textcolor{black!55}{-} & \resultmetric{0.212}{0.020} & \textcolor{black!55}{-} & \textcolor{black!55}{-} \\
\specialrule{0.25pt}{2pt}{2pt}
\multirow{4}{*}[-3pt]{DeepRUOT} & $W_1$ & \resultmetric{0.765}{0.033} & \resultmetric{1.033}{0.290} & \textcolor{black!55}{-} & \textcolor{black!55}{-} & \resultmetric{78.561}{0.676} & \textcolor{black!55}{-} & \textcolor{black!55}{-} \\
\arrayrulecolor{black!45}
\cmidrule[0.2pt](l{1pt}r{1pt}){3-9}
\arrayrulecolor{black}
 & SW1 & \resultmetric{0.492}{0.020} & \resultmetric{0.651}{0.187} & \textcolor{black!55}{-} & \textcolor{black!55}{-} & \resultmetric{14.119}{0.687} & \textcolor{black!55}{-} & \textcolor{black!55}{-} \\
\arrayrulecolor{black!45}
\cmidrule[0.2pt](l{1pt}r{1pt}){3-9}
\arrayrulecolor{black}
 & $W_2$ & \resultmetric{0.814}{0.032} & \resultmetric{1.100}{0.281} & \textcolor{black!55}{-} & \textcolor{black!55}{-} & \resultmetric{93.868}{1.043} & \textcolor{black!55}{-} & \textcolor{black!55}{-} \\
\arrayrulecolor{black!45}
\cmidrule[0.2pt](l{1pt}r{1pt}){3-9}
\arrayrulecolor{black}
 & MMD & \resultmetric{0.575}{0.017} & \resultmetric{0.522}{0.140} & \textcolor{black!55}{-} & \textcolor{black!55}{-} & \resultmetric{0.019}{0.002} & \textcolor{black!55}{-} & \textcolor{black!55}{-} \\
\specialrule{0.25pt}{2pt}{2pt}
\multirow{4}{*}[-3pt]{VGFM} & $W_1$ & \resultmetric{1.247}{0.477} & \resultmetric{11.542}{0.528} & \textcolor{black!55}{-} & \textcolor{black!55}{-} & \resultmetric{117.983}{2.040} & \textcolor{black!55}{-} & \textcolor{black!55}{-} \\
\arrayrulecolor{black!45}
\cmidrule[0.2pt](l{1pt}r{1pt}){3-9}
\arrayrulecolor{black}
 & SW1 & \resultmetric{0.780}{0.309} & \resultmetric{8.395}{1.109} & \textcolor{black!55}{-} & \textcolor{black!55}{-} & \resultmetric{24.104}{1.296} & \textcolor{black!55}{-} & \textcolor{black!55}{-} \\
\arrayrulecolor{black!45}
\cmidrule[0.2pt](l{1pt}r{1pt}){3-9}
\arrayrulecolor{black}
 & $W_2$ & \resultmetric{1.303}{0.469} & \resultmetric{50.033}{1.136} & \textcolor{black!55}{-} & \textcolor{black!55}{-} & \resultmetric{138.186}{2.053} & \textcolor{black!55}{-} & \textcolor{black!55}{-} \\
\arrayrulecolor{black!45}
\cmidrule[0.2pt](l{1pt}r{1pt}){3-9}
\arrayrulecolor{black}
 & MMD & \resultmetric{0.653}{0.234} & \resultmetric{0.525}{0.107} & \textcolor{black!55}{-} & \textcolor{black!55}{-} & \resultmetric{0.060}{0.057} & \textcolor{black!55}{-} & \textcolor{black!55}{-} \\
\specialrule{0.25pt}{2pt}{2pt}
\multirow{4}{*}[-3pt]{MIOFlow} & $W_1$ & \resultmetric{1.424}{0.057} & \resultmetric{0.692}{0.109} & \textcolor{black!55}{-} & \textcolor{black!55}{-} & \resultmetric{98.199}{1.144} & \textcolor{black!55}{-} & \textcolor{black!55}{-} \\
\arrayrulecolor{black!45}
\cmidrule[0.2pt](l{1pt}r{1pt}){3-9}
\arrayrulecolor{black}
 & SW1 & \resultmetric{0.895}{0.034} & \resultmetric{0.432}{0.071} & \textcolor{black!55}{-} & \textcolor{black!55}{-} & \resultmetric{20.917}{1.353} & \textcolor{black!55}{-} & \textcolor{black!55}{-} \\
\arrayrulecolor{black!45}
\cmidrule[0.2pt](l{1pt}r{1pt}){3-9}
\arrayrulecolor{black}
 & $W_2$ & \resultmetric{1.507}{0.044} & \resultmetric{0.790}{0.142} & \textcolor{black!55}{-} & \textcolor{black!55}{-} & \resultmetric{115.136}{1.277} & \textcolor{black!55}{-} & \textcolor{black!55}{-} \\
\arrayrulecolor{black!45}
\cmidrule[0.2pt](l{1pt}r{1pt}){3-9}
\arrayrulecolor{black}
 & MMD & \resultmetric{0.669}{0.014} & \resultmetric{0.318}{0.085} & \textcolor{black!55}{-} & \textcolor{black!55}{-} & \resultmetric{0.058}{0.023} & \textcolor{black!55}{-} & \textcolor{black!55}{-} \\
\specialrule{0.25pt}{2pt}{2pt}
\multirow{4}{*}[-3pt]{scNODE} & $W_1$ & \resultmetric{0.738}{0.008} & \resultmetric{0.886}{0.075} & \textcolor{black!55}{-} & \textcolor{black!55}{-} & \resultmetric{145.715}{1.770} & \textcolor{black!55}{-} & \textcolor{black!55}{-} \\
\arrayrulecolor{black!45}
\cmidrule[0.2pt](l{1pt}r{1pt}){3-9}
\arrayrulecolor{black}
 & SW1 & \resultmetric{0.468}{0.006} & \resultmetric{0.541}{0.071} & \textcolor{black!55}{-} & \textcolor{black!55}{-} & \resultmetric{36.200}{1.752} & \textcolor{black!55}{-} & \textcolor{black!55}{-} \\
\arrayrulecolor{black!45}
\cmidrule[0.2pt](l{1pt}r{1pt}){3-9}
\arrayrulecolor{black}
 & $W_2$ & \resultmetric{0.806}{0.008} & \resultmetric{0.952}{0.056} & \textcolor{black!55}{-} & \textcolor{black!55}{-} & \resultmetric{166.598}{1.898} & \textcolor{black!55}{-} & \textcolor{black!55}{-} \\
\arrayrulecolor{black!45}
\cmidrule[0.2pt](l{1pt}r{1pt}){3-9}
\arrayrulecolor{black}
 & MMD & \resultmetric{0.399}{0.012} & \resultmetric{0.321}{0.093} & \textcolor{black!55}{-} & \textcolor{black!55}{-} & \resultmetric{0.202}{0.019} & \textcolor{black!55}{-} & \textcolor{black!55}{-} \\
\specialrule{0.25pt}{2pt}{2pt}
\multirow{4}{*}[-3pt]{PRESCIENT} & $W_1$ & \resultmetric{0.768}{0.032} & \resultmetric{0.534}{0.041} & \textcolor{black!55}{-} & \textcolor{black!55}{-} & \resultmetric{147.606}{0.199} & \textcolor{black!55}{-} & \textcolor{black!55}{-} \\
\arrayrulecolor{black!45}
\cmidrule[0.2pt](l{1pt}r{1pt}){3-9}
\arrayrulecolor{black}
 & SW1 & \resultmetric{0.474}{0.019} & \resultmetric{0.282}{0.025} & \textcolor{black!55}{-} & \textcolor{black!55}{-} & \resultmetric{38.188}{0.053} & \textcolor{black!55}{-} & \textcolor{black!55}{-} \\
\arrayrulecolor{black!45}
\cmidrule[0.2pt](l{1pt}r{1pt}){3-9}
\arrayrulecolor{black}
 & $W_2$ & \resultmetric{0.814}{0.031} & \resultmetric{0.663}{0.053} & \textcolor{black!55}{-} & \textcolor{black!55}{-} & \resultmetric{169.867}{0.200} & \textcolor{black!55}{-} & \textcolor{black!55}{-} \\
\arrayrulecolor{black!45}
\cmidrule[0.2pt](l{1pt}r{1pt}){3-9}
\arrayrulecolor{black}
 & MMD & \resultmetric{0.446}{0.009} & \resultmetric{0.093}{0.007} & \textcolor{black!55}{-} & \textcolor{black!55}{-} & \resultmetric{0.282}{0.001} & \textcolor{black!55}{-} & \textcolor{black!55}{-} \\
\specialrule{0.25pt}{2pt}{2pt}
\multirow{4}{*}[-3pt]{JKOnet*} & $W_1$ & \resultmetric{1.223}{0.051} & \resultmetric{5.267}{0.452} & \textcolor{black!55}{-} & \textcolor{black!55}{-} & \resultmetric{130.758}{1.101} & \textcolor{black!55}{-} & \textcolor{black!55}{-} \\
\arrayrulecolor{black!45}
\cmidrule[0.2pt](l{1pt}r{1pt}){3-9}
\arrayrulecolor{black}
 & SW1 & \resultmetric{0.761}{0.032} & \resultmetric{3.300}{0.403} & \textcolor{black!55}{-} & \textcolor{black!55}{-} & \resultmetric{29.383}{1.434} & \textcolor{black!55}{-} & \textcolor{black!55}{-} \\
\arrayrulecolor{black!45}
\cmidrule[0.2pt](l{1pt}r{1pt}){3-9}
\arrayrulecolor{black}
 & $W_2$ & \resultmetric{1.263}{0.048} & \resultmetric{8.796}{0.707} & \textcolor{black!55}{-} & \textcolor{black!55}{-} & \resultmetric{156.289}{1.761} & \textcolor{black!55}{-} & \textcolor{black!55}{-} \\
\arrayrulecolor{black!45}
\cmidrule[0.2pt](l{1pt}r{1pt}){3-9}
\arrayrulecolor{black}
 & MMD & \resultmetric{0.584}{0.017} & \resultmetric{0.179}{0.048} & \textcolor{black!55}{-} & \textcolor{black!55}{-} & \resultmetric{0.059}{0.005} & \textcolor{black!55}{-} & \textcolor{black!55}{-} \\
\specialrule{0.25pt}{2pt}{2pt}
\multirow{4}{*}[-3pt]{WLM} & $W_1$ & \textcolor{resultSecond}{\resultmetric{0.534}{0.033}} & \textcolor{resultSecond}{\resultmetric{0.316}{0.027}} & \textcolor{black!55}{-} & \textcolor{black!55}{-} & \textcolor{resultSecond}{\resultmetric{52.000}{0.323}} & \textcolor{black!55}{-} & \textcolor{black!55}{-} \\
\arrayrulecolor{black!45}
\cmidrule[0.2pt](l{1pt}r{1pt}){3-9}
\arrayrulecolor{black}
 & SW1 & \textcolor{resultSecond}{\resultmetric{0.318}{0.021}} & \textcolor{resultSecond}{\resultmetric{0.159}{0.019}} & \textcolor{black!55}{-} & \textcolor{black!55}{-} & \textcolor{resultFirst}{\resultmetric{6.013}{0.212}} & \textcolor{black!55}{-} & \textcolor{black!55}{-} \\
\arrayrulecolor{black!45}
\cmidrule[0.2pt](l{1pt}r{1pt}){3-9}
\arrayrulecolor{black}
 & $W_2$ & \textcolor{resultSecond}{\resultmetric{0.599}{0.033}} & \textcolor{resultSecond}{\resultmetric{0.411}{0.033}} & \textcolor{black!55}{-} & \textcolor{black!55}{-} & \textcolor{resultSecond}{\resultmetric{65.203}{0.326}} & \textcolor{black!55}{-} & \textcolor{black!55}{-} \\
\arrayrulecolor{black!45}
\cmidrule[0.2pt](l{1pt}r{1pt}){3-9}
\arrayrulecolor{black}
 & MMD & \textcolor{resultSecond}{\resultmetric{0.223}{0.020}} & \textcolor{resultSecond}{\resultmetric{0.028}{0.010}} & \textcolor{black!55}{-} & \textcolor{black!55}{-} & \textcolor{resultFirst}{\resultmetric{0.004}{0.000}} & \textcolor{black!55}{-} & \textcolor{black!55}{-} \\
\midrule
\multirow{4}{*}[-3pt]{\textbf{DisKO}} & $W_1$ & \textcolor{resultFirst}{\resultmetric{0.238}{0.014}} & \textcolor{resultFirst}{\resultmetric{0.145}{0.005}} & \textcolor{black!55}{-} & \textcolor{black!55}{-} & \textcolor{resultFirst}{\resultmetric{47.995}{0.404}} & \textcolor{black!55}{-} & \textcolor{black!55}{-} \\
\arrayrulecolor{black!45}
\cmidrule[0.2pt](l{1pt}r{1pt}){3-9}
\arrayrulecolor{black}
 & SW1 & \textcolor{resultFirst}{\resultmetric{0.129}{0.011}} & \textcolor{resultFirst}{\resultmetric{0.056}{0.004}} & \textcolor{black!55}{-} & \textcolor{black!55}{-} & \textcolor{resultSecond}{\resultmetric{6.142}{0.401}} & \textcolor{black!55}{-} & \textcolor{black!55}{-} \\
\arrayrulecolor{black!45}
\cmidrule[0.2pt](l{1pt}r{1pt}){3-9}
\arrayrulecolor{black}
 & $W_2$ & \textcolor{resultFirst}{\resultmetric{0.279}{0.014}} & \textcolor{resultFirst}{\resultmetric{0.195}{0.008}} & \textcolor{black!55}{-} & \textcolor{black!55}{-} & \textcolor{resultFirst}{\resultmetric{62.773}{0.711}} & \textcolor{black!55}{-} & \textcolor{black!55}{-} \\
\arrayrulecolor{black!45}
\cmidrule[0.2pt](l{1pt}r{1pt}){3-9}
\arrayrulecolor{black}
 & MMD & \textcolor{resultFirst}{\resultmetric{0.029}{0.006}} & \textcolor{resultFirst}{\resultmetric{0.003}{0.001}} & \textcolor{black!55}{-} & \textcolor{black!55}{-} & \textcolor{resultSecond}{\resultmetric{0.005}{0.001}} & \textcolor{black!55}{-} & \textcolor{black!55}{-} \\
\bottomrule
\end{tabular*}
\endgroup
\end{table}

\clearpage
\begin{table}[!htbp]
\centering
\begingroup
\definecolor{resultFirst}{RGB}{190,25,35}
\definecolor{resultSecond}{RGB}{25,80,185}
\caption{Temporal extrapolation from test initial distributions and single sequences. Boids, Ocean, Pancreas, and WOT iPSC each contain a single population sequence and use only a temporal partition.
Means and standard deviations are rounded to three decimal places.
\textcolor{resultFirst}{Red} and \textcolor{resultSecond}{blue} mark the lowest and second lowest means before rounding.}
\label{tab:test-time-test-initial}
\fontsize{8}{10.5}\selectfont
\setlength{\tabcolsep}{1pt}
\renewcommand{\arraystretch}{1.05}
\setlength{\aboverulesep}{1pt}
\setlength{\belowrulesep}{1pt}
\newcommand{\resultmetric}[2]{\mbox{#1\textsubscript{\fontsize{6}{7}\selectfont$\pm$#2}}}
\begin{tabular*}{\linewidth}{@{}l@{\hspace{4pt}}c@{\hspace{4pt}\extracolsep{\fill}}ccccccc@{}}
\toprule
Method & Metric & OU & Duffing & Boids & Ocean & CAMELS & Pancreas & WOT iPSC \\
\midrule
\multirow{4}{*}[-3pt]{TrajectoryNet} & $W_1$ & \resultmetric{1.369}{0.071} & \resultmetric{9.222}{0.783} & \resultmetric{11.843}{0.560} & \resultmetric{2.198}{0.900} & \resultmetric{143.767}{2.170} & \resultmetric{47.376}{2.950} & \resultmetric{118.524}{2.509} \\
\arrayrulecolor{black!45}
\cmidrule[0.2pt](l{1pt}r{1pt}){3-9}
\arrayrulecolor{black}
 & SW1 & \resultmetric{0.857}{0.047} & \resultmetric{5.815}{0.712} & \resultmetric{7.267}{0.365} & \resultmetric{1.426}{0.556} & \resultmetric{35.364}{1.128} & \resultmetric{6.765}{0.826} & \resultmetric{14.502}{0.647} \\
\arrayrulecolor{black!45}
\cmidrule[0.2pt](l{1pt}r{1pt}){3-9}
\arrayrulecolor{black}
 & $W_2$ & \resultmetric{1.443}{0.067} & \resultmetric{17.953}{1.433} & \resultmetric{15.612}{0.973} & \resultmetric{2.204}{0.893} & \resultmetric{166.789}{2.871} & \resultmetric{53.487}{1.372} & \resultmetric{321.936}{2.588} \\
\arrayrulecolor{black!45}
\cmidrule[0.2pt](l{1pt}r{1pt}){3-9}
\arrayrulecolor{black}
 & MMD & \resultmetric{0.666}{0.052} & \resultmetric{0.523}{0.140} & \resultmetric{0.348}{0.078} & \resultmetric{0.957}{0.074} & \resultmetric{0.233}{0.024} & \resultmetric{0.372}{0.101} & \resultmetric{0.628}{0.086} \\
\specialrule{0.25pt}{2pt}{2pt}
\multirow{4}{*}[-3pt]{DeepRUOT} & $W_1$ & \resultmetric{0.846}{0.044} & \resultmetric{1.032}{0.287} & \resultmetric{5.962}{0.283} & \resultmetric{0.214}{0.020} & \resultmetric{79.989}{1.252} & \resultmetric{17.286}{1.572} & \resultmetric{23.635}{1.623} \\
\arrayrulecolor{black!45}
\cmidrule[0.2pt](l{1pt}r{1pt}){3-9}
\arrayrulecolor{black}
 & SW1 & \resultmetric{0.554}{0.030} & \resultmetric{0.650}{0.186} & \resultmetric{3.796}{0.172} & \resultmetric{0.149}{0.014} & \resultmetric{14.362}{0.466} & \resultmetric{2.113}{0.221} & \resultmetric{2.108}{0.220} \\
\arrayrulecolor{black!45}
\cmidrule[0.2pt](l{1pt}r{1pt}){3-9}
\arrayrulecolor{black}
 & $W_2$ & \resultmetric{0.892}{0.042} & \resultmetric{1.099}{0.278} & \resultmetric{6.356}{0.349} & \resultmetric{0.274}{0.020} & \resultmetric{96.028}{1.647} & \resultmetric{18.801}{2.136} & \resultmetric{23.822}{1.575} \\
\arrayrulecolor{black!45}
\cmidrule[0.2pt](l{1pt}r{1pt}){3-9}
\arrayrulecolor{black}
 & MMD & \resultmetric{0.593}{0.046} & \resultmetric{0.523}{0.141} & \resultmetric{0.331}{0.090} & \resultmetric{0.598}{0.062} & \resultmetric{0.022}{0.002} & \resultmetric{0.122}{0.050} & \resultmetric{0.227}{0.046} \\
\specialrule{0.25pt}{2pt}{2pt}
\multirow{4}{*}[-3pt]{VGFM} & $W_1$ & \resultmetric{1.290}{0.486} & \resultmetric{12.834}{0.976} & \resultmetric{31.008}{1.709} & \resultmetric{0.195}{0.047} & \resultmetric{123.071}{2.075} & \resultmetric{27.497}{1.337} & \resultmetric{187.154}{2.894} \\
\arrayrulecolor{black!45}
\cmidrule[0.2pt](l{1pt}r{1pt}){3-9}
\arrayrulecolor{black}
 & SW1 & \resultmetric{0.807}{0.315} & \resultmetric{9.358}{0.352} & \resultmetric{19.604}{1.428} & \resultmetric{0.128}{0.034} & \resultmetric{24.995}{17.602} & \resultmetric{2.978}{2.126} & \resultmetric{27.389}{1.762} \\
\arrayrulecolor{black!45}
\cmidrule[0.2pt](l{1pt}r{1pt}){3-9}
\arrayrulecolor{black}
 & $W_2$ & \resultmetric{1.352}{0.478} & \resultmetric{62.176}{1.946} & \resultmetric{32.798}{1.546} & \resultmetric{0.210}{0.047} & \resultmetric{145.189}{2.565} & \resultmetric{49.009}{1.408} & \resultmetric{370.921}{2.025} \\
\arrayrulecolor{black!45}
\cmidrule[0.2pt](l{1pt}r{1pt}){3-9}
\arrayrulecolor{black}
 & MMD & \resultmetric{0.651}{0.242} & \resultmetric{0.526}{0.109} & \resultmetric{0.440}{0.251} & \textcolor{resultSecond}{\resultmetric{0.248}{0.100}} & \resultmetric{0.065}{0.066} & \resultmetric{0.091}{0.048} & \textcolor{resultSecond}{\resultmetric{0.166}{0.032}} \\
\specialrule{0.25pt}{2pt}{2pt}
\multirow{4}{*}[-3pt]{MIOFlow} & $W_1$ & \resultmetric{1.514}{0.135} & \resultmetric{0.692}{0.109} & \resultmetric{9.642}{0.390} & \resultmetric{0.187}{0.044} & \resultmetric{98.173}{1.421} & \textcolor{resultSecond}{\resultmetric{11.111}{0.902}} & \resultmetric{34.854}{0.627} \\
\arrayrulecolor{black!45}
\cmidrule[0.2pt](l{1pt}r{1pt}){3-9}
\arrayrulecolor{black}
 & SW1 & \resultmetric{0.950}{0.087} & \resultmetric{0.432}{0.070} & \resultmetric{5.879}{1.216} & \resultmetric{0.130}{0.033} & \resultmetric{20.381}{1.104} & \textcolor{resultSecond}{\resultmetric{0.895}{0.171}} & \resultmetric{3.931}{0.053} \\
\arrayrulecolor{black!45}
\cmidrule[0.2pt](l{1pt}r{1pt}){3-9}
\arrayrulecolor{black}
 & $W_2$ & \resultmetric{1.612}{0.161} & \resultmetric{0.804}{0.166} & \resultmetric{13.322}{1.066} & \resultmetric{0.233}{0.048} & \resultmetric{115.750}{2.881} & \textcolor{resultSecond}{\resultmetric{12.305}{1.198}} & \resultmetric{34.916}{0.620} \\
\arrayrulecolor{black!45}
\cmidrule[0.2pt](l{1pt}r{1pt}){3-9}
\arrayrulecolor{black}
 & MMD & \resultmetric{0.693}{0.062} & \resultmetric{0.320}{0.084} & \resultmetric{0.221}{0.109} & \resultmetric{0.492}{0.163} & \resultmetric{0.057}{0.023} & \textcolor{resultSecond}{\resultmetric{0.027}{0.013}} & \resultmetric{0.479}{0.012} \\
\specialrule{0.25pt}{2pt}{2pt}
\multirow{4}{*}[-3pt]{scNODE} & $W_1$ & \resultmetric{0.759}{0.010} & \resultmetric{0.887}{0.075} & \resultmetric{7.822}{0.359} & \resultmetric{0.783}{0.246} & \resultmetric{143.751}{1.692} & \resultmetric{43.047}{1.030} & \resultmetric{94.268}{1.536} \\
\arrayrulecolor{black!45}
\cmidrule[0.2pt](l{1pt}r{1pt}){3-9}
\arrayrulecolor{black}
 & SW1 & \resultmetric{0.481}{0.008} & \resultmetric{0.541}{0.070} & \resultmetric{4.381}{0.233} & \resultmetric{0.515}{0.154} & \resultmetric{34.896}{1.137} & \resultmetric{6.168}{0.831} & \resultmetric{10.544}{1.150} \\
\arrayrulecolor{black!45}
\cmidrule[0.2pt](l{1pt}r{1pt}){3-9}
\arrayrulecolor{black}
 & $W_2$ & \resultmetric{0.827}{0.010} & \resultmetric{0.952}{0.056} & \resultmetric{9.633}{0.649} & \resultmetric{0.789}{0.239} & \resultmetric{166.307}{1.709} & \resultmetric{44.610}{2.552} & \resultmetric{94.623}{1.388} \\
\arrayrulecolor{black!45}
\cmidrule[0.2pt](l{1pt}r{1pt}){3-9}
\arrayrulecolor{black}
 & MMD & \resultmetric{0.403}{0.015} & \resultmetric{0.324}{0.091} & \resultmetric{0.060}{0.004} & \resultmetric{1.302}{0.104} & \resultmetric{0.211}{0.018} & \resultmetric{0.659}{0.139} & \resultmetric{1.077}{0.250} \\
\specialrule{0.25pt}{2pt}{2pt}
\multirow{4}{*}[-3pt]{PRESCIENT} & $W_1$ & \resultmetric{0.849}{0.034} & \resultmetric{0.532}{0.040} & \resultmetric{3.498}{0.136} & \textcolor{resultSecond}{\resultmetric{0.153}{0.009}} & \resultmetric{143.797}{0.232} & \resultmetric{12.169}{0.325} & \textcolor{resultSecond}{\resultmetric{23.328}{0.086}} \\
\arrayrulecolor{black!45}
\cmidrule[0.2pt](l{1pt}r{1pt}){3-9}
\arrayrulecolor{black}
 & SW1 & \resultmetric{0.525}{0.020} & \resultmetric{0.280}{0.025} & \resultmetric{1.951}{0.087} & \textcolor{resultSecond}{\resultmetric{0.097}{0.006}} & \resultmetric{36.244}{0.062} & \resultmetric{0.965}{0.068} & \textcolor{resultSecond}{\resultmetric{1.883}{0.013}} \\
\arrayrulecolor{black!45}
\cmidrule[0.2pt](l{1pt}r{1pt}){3-9}
\arrayrulecolor{black}
 & $W_2$ & \resultmetric{0.894}{0.032} & \resultmetric{0.660}{0.052} & \resultmetric{3.845}{0.148} & \textcolor{resultSecond}{\resultmetric{0.186}{0.007}} & \resultmetric{167.681}{0.273} & \resultmetric{12.873}{0.393} & \textcolor{resultSecond}{\resultmetric{23.397}{0.086}} \\
\arrayrulecolor{black!45}
\cmidrule[0.2pt](l{1pt}r{1pt}){3-9}
\arrayrulecolor{black}
 & MMD & \resultmetric{0.483}{0.007} & \resultmetric{0.093}{0.007} & \resultmetric{0.077}{0.005} & \resultmetric{0.400}{0.055} & \resultmetric{0.281}{0.001} & \resultmetric{0.054}{0.008} & \resultmetric{0.177}{0.002} \\
\specialrule{0.25pt}{2pt}{2pt}
\multirow{4}{*}[-3pt]{JKOnet*} & $W_1$ & \resultmetric{1.288}{0.067} & \resultmetric{5.246}{0.447} & \resultmetric{3.282}{0.187} & \resultmetric{2.332}{0.876} & \resultmetric{134.105}{2.073} & \resultmetric{52.395}{1.119} & \resultmetric{63.051}{1.367} \\
\arrayrulecolor{black!45}
\cmidrule[0.2pt](l{1pt}r{1pt}){3-9}
\arrayrulecolor{black}
 & SW1 & \resultmetric{0.797}{0.044} & \resultmetric{3.290}{0.413} & \resultmetric{1.770}{0.086} & \resultmetric{1.561}{0.587} & \resultmetric{30.058}{0.976} & \resultmetric{8.221}{0.713} & \resultmetric{7.076}{0.325} \\
\arrayrulecolor{black!45}
\cmidrule[0.2pt](l{1pt}r{1pt}){3-9}
\arrayrulecolor{black}
 & $W_2$ & \resultmetric{1.328}{0.060} & \resultmetric{8.726}{0.704} & \resultmetric{3.857}{0.286} & \resultmetric{2.743}{0.891} & \resultmetric{160.195}{2.727} & \resultmetric{76.620}{1.496} & \resultmetric{75.024}{2.428} \\
\arrayrulecolor{black!45}
\cmidrule[0.2pt](l{1pt}r{1pt}){3-9}
\arrayrulecolor{black}
 & MMD & \resultmetric{0.594}{0.046} & \resultmetric{0.180}{0.048} & \resultmetric{0.053}{0.007} & \resultmetric{0.868}{0.089} & \resultmetric{0.062}{0.006} & \resultmetric{0.303}{0.095} & \resultmetric{0.206}{0.028} \\
\specialrule{0.25pt}{2pt}{2pt}
\multirow{4}{*}[-3pt]{WLM} & $W_1$ & \textcolor{resultSecond}{\resultmetric{0.602}{0.032}} & \textcolor{resultSecond}{\resultmetric{0.318}{0.028}} & \textcolor{resultSecond}{\resultmetric{1.595}{0.047}} & \resultmetric{0.941}{0.264} & \textcolor{resultSecond}{\resultmetric{55.019}{0.370}} & \resultmetric{13.345}{0.517} & \resultmetric{25.805}{0.638} \\
\arrayrulecolor{black!45}
\cmidrule[0.2pt](l{1pt}r{1pt}){3-9}
\arrayrulecolor{black}
 & SW1 & \textcolor{resultSecond}{\resultmetric{0.362}{0.020}} & \textcolor{resultSecond}{\resultmetric{0.160}{0.019}} & \textcolor{resultSecond}{\resultmetric{0.656}{0.050}} & \resultmetric{0.520}{0.246} & \textcolor{resultFirst}{\resultmetric{6.699}{0.206}} & \resultmetric{0.988}{0.140} & \resultmetric{1.929}{0.093} \\
\arrayrulecolor{black!45}
\cmidrule[0.2pt](l{1pt}r{1pt}){3-9}
\arrayrulecolor{black}
 & $W_2$ & \textcolor{resultSecond}{\resultmetric{0.688}{0.037}} & \textcolor{resultSecond}{\resultmetric{0.413}{0.034}} & \textcolor{resultSecond}{\resultmetric{2.152}{0.165}} & \resultmetric{1.845}{0.848} & \textcolor{resultFirst}{\resultmetric{71.177}{0.327}} & \resultmetric{13.969}{0.611} & \resultmetric{26.921}{1.159} \\
\arrayrulecolor{black!45}
\cmidrule[0.2pt](l{1pt}r{1pt}){3-9}
\arrayrulecolor{black}
 & MMD & \textcolor{resultSecond}{\resultmetric{0.255}{0.019}} & \textcolor{resultSecond}{\resultmetric{0.028}{0.010}} & \textcolor{resultSecond}{\resultmetric{0.011}{0.002}} & \resultmetric{0.366}{0.127} & \textcolor{resultFirst}{\resultmetric{0.005}{0.000}} & \resultmetric{0.055}{0.015} & \resultmetric{0.169}{0.023} \\
\midrule
\multirow{4}{*}[-3pt]{\textbf{DisKO}} & $W_1$ & \textcolor{resultFirst}{\resultmetric{0.249}{0.013}} & \textcolor{resultFirst}{\resultmetric{0.146}{0.005}} & \textcolor{resultFirst}{\resultmetric{1.454}{0.110}} & \textcolor{resultFirst}{\resultmetric{0.097}{0.024}} & \textcolor{resultFirst}{\resultmetric{54.241}{0.830}} & \textcolor{resultFirst}{\resultmetric{9.114}{0.143}} & \textcolor{resultFirst}{\resultmetric{18.973}{0.330}} \\
\arrayrulecolor{black!45}
\cmidrule[0.2pt](l{1pt}r{1pt}){3-9}
\arrayrulecolor{black}
 & SW1 & \textcolor{resultFirst}{\resultmetric{0.135}{0.010}} & \textcolor{resultFirst}{\resultmetric{0.056}{0.004}} & \textcolor{resultFirst}{\resultmetric{0.656}{0.087}} & \textcolor{resultFirst}{\resultmetric{0.062}{0.019}} & \textcolor{resultSecond}{\resultmetric{7.039}{0.500}} & \textcolor{resultFirst}{\resultmetric{0.559}{0.007}} & \textcolor{resultFirst}{\resultmetric{0.940}{0.039}} \\
\arrayrulecolor{black!45}
\cmidrule[0.2pt](l{1pt}r{1pt}){3-9}
\arrayrulecolor{black}
 & $W_2$ & \textcolor{resultFirst}{\resultmetric{0.294}{0.014}} & \textcolor{resultFirst}{\resultmetric{0.195}{0.008}} & \textcolor{resultFirst}{\resultmetric{1.827}{0.104}} & \textcolor{resultFirst}{\resultmetric{0.115}{0.023}} & \textcolor{resultSecond}{\resultmetric{71.260}{1.239}} & \textcolor{resultFirst}{\resultmetric{9.372}{0.109}} & \textcolor{resultFirst}{\resultmetric{19.356}{0.324}} \\
\arrayrulecolor{black!45}
\cmidrule[0.2pt](l{1pt}r{1pt}){3-9}
\arrayrulecolor{black}
 & MMD & \textcolor{resultFirst}{\resultmetric{0.030}{0.005}} & \textcolor{resultFirst}{\resultmetric{0.003}{0.001}} & \textcolor{resultFirst}{\resultmetric{0.011}{0.004}} & \textcolor{resultFirst}{\resultmetric{0.085}{0.046}} & \textcolor{resultSecond}{\resultmetric{0.007}{0.001}} & \textcolor{resultFirst}{\resultmetric{0.013}{0.000}} & \textcolor{resultFirst}{\resultmetric{0.036}{0.004}} \\
\bottomrule
\end{tabular*}
\endgroup
\end{table}

\clearpage

\section{Koopman Spectral Approximation}
\label{app:spectral-approximation}
\setcounter{table}{0}
\setcounter{figure}{0}
\renewcommand{\thefigure}{\thesection.\arabic{figure}}

We examine the approximation of Koopman eigenvalues and eigenfunctions on Circle and Torus using discrete and continuous parameterizations of DisKO.
Following the terminology of \citet{brunton2022modern,klus2020generator}, an eigenfunction is a scalar observable satisfying an eigenvalue relation with the operator or its generator.
Here its argument is a probability measure, so it is a distribution observable as defined in Section~\ref{sec:distributional-koopman}.
We reserve the term Koopman mode for a coefficient vector in the eigenfunction expansion of a vector-valued observable.
The figures below display eigenvalue estimates and evaluations of learned eigenfunctions on distributions.

\subsection{Analytical Eigenvalues and Eigenfunctions}
\label{app:spectral-reference}

Write $\vartheta_t\in\mathbb T^q$ for the angular state, where $q=1$ for Circle and $q=2$ for Torus.
The dynamics and the generator acting on smooth periodic state observables are
\begin{equation}
    d\vartheta_{j,t}=\omega_j\,dt+\sqrt{2\kappa_j}\,dW_{j,t}
    \pmod{2\pi},
    \qquad
    \mathcal G f=\sum_{j=1}^q\left(\omega_j\partial_j f+\kappa_j\partial_{jj}f\right),
    \label{eq:spectral-angular-generator}
\end{equation}
with independent Brownian motions.
Circle uses $(\omega_1,\kappa_1)=(1,0.01)$; Torus uses $\omega=(1,1.7)$ and $\kappa=(0.01,0.014)$.
For $k\in\mathbb Z^q$, the Fourier function $\phi_k(\vartheta)=e^{\mathrm{i}k\cdot\vartheta}$ satisfies
\begin{equation}
    \mathcal G\phi_k=\lambda_k\phi_k,
    \qquad
    \lambda_k=-\sum_{j=1}^q\kappa_j k_j^2+\mathrm{i}\sum_{j=1}^q\omega_j k_j.
    \label{eq:spectral-generator-eigenvalue}
\end{equation}
This follows directly from $\partial_j\phi_k=\mathrm{i}k_j\phi_k$ and $\partial_{jj}\phi_k=-k_j^2\phi_k$.
The real and imaginary parts of $\lambda_k$ determine the decay rate and angular frequency, respectively.

Let $\mu_t=T_t\mu_0$ be the angular state distribution and define $h_k(\mu)=\int\phi_k\,d\mu$.
Taking expectations in It\^o's formula yields
\begin{equation}
    \frac{d}{dt}h_k(\mu_t)=\int\mathcal G\phi_k\,d\mu_t
    =\lambda_k h_k(\mu_t),
    \qquad
    (\mathcal K_t h_k)(\mu)=e^{t\lambda_k}h_k(\mu).
    \label{eq:spectral-distribution-eigenfunction}
\end{equation}
Thus $h_k$ is an eigenfunction of the distributional Koopman generator $\mathcal L$ with eigenvalue $\lambda_k$, and of $\mathcal K_{\Delta t}$ with eigenvalue $\rho_k=e^{\Delta t\lambda_k}$.
The observation maps are $P(\vartheta)=(\cos\vartheta,\sin\vartheta)$ on Circle and the concatenated sine and cosine pairs on Torus.
They are injective on the respective periodic state spaces, so the observed distribution $\nu=P_\#\mu$ determines $h_k(\mu)$.

All spectral comparisons use $\Delta t=0.1$.
Circle has $\lambda_k=-0.01k^2+\mathrm{i}k$, with target indices $k=1,\ldots,6$.
Torus has $\lambda_{(m,n)}=-(0.01m^2+0.014n^2)+\mathrm{i}(m+1.7n)$, with target indices $(1,0)$, $(0,1)$, $(1,1)$, $(1,-1)$, $(2,0)$, and $(0,2)$.
These prescribe a finite set of reference eigenpairs.
The constant eigenfunction has generator eigenvalue zero and operator eigenvalue one; the index $-k$ gives the complex conjugate of the eigenpair for $k$.

For an initial mixture of wrapped Gaussians with weights $\pi_\ell$, centers $a_\ell$, and angular covariance matrices $\Sigma_\ell$, the reference values at every time are available analytically,
\begin{equation}
    h_k(\mu_t)=e^{t\lambda_k}\sum_\ell\pi_\ell
    \exp\!\left(\mathrm{i}k\cdot a_\ell-\tfrac12 k^\top\Sigma_\ell k\right).
    \label{eq:spectral-mixture-moment}
\end{equation}
Reference values use the initial mixture parameters and Equation~\ref{eq:spectral-mixture-moment}; learned values are computed from finite empirical snapshots.

\subsection{Discrete and Continuous DisKO}
\label{app:spectral-parameterizations}

The two models share the encoder and conditional flow decoder architecture, training distributions, particle subsets, and pretrained weights within each dataset.
The encoder has a Deep Sets construction of depth three and width 256, with latent dimension 64; the decoder velocity network has depth five and width 256.
Both systems use 128 training population sequences with 1,024 particles per snapshot.
Circle has 16 validation and 32 test sequences, and Torus has 24 validation and 48 test sequences.
Training uses 81 snapshots over $0\le t\le8$; the subsequent 40 snapshots cover $8.1\le t\le12$.
We report the final models after 10,000 joint optimization steps, with seed zero for both parameterizations.

For the discrete model, $z_{n+1}=Kz_n+b$ at interval $\Delta t$.
For the continuous model, $\dot z=Az+c$ and propagation uses an exact matrix exponential.
With $\bar z=[z^\top,1]^\top$, these become
\begin{equation}
    \bar z_{n+1}=M\bar z_n,
    \quad M=\begin{pmatrix}K&b\\0&1\end{pmatrix},
    \qquad
    \bar z(t+\Delta t)=e^{\Delta t B}\bar z(t),
    \quad B=\begin{pmatrix}A&c\\0&0\end{pmatrix}.
    \label{eq:spectral-augmented-models}
\end{equation}
If $M^\top w=\widehat\rho w$, the scalar function $u(\nu)=w^\top[E(\nu)^\top,1]^\top$ evolves by $\widehat\rho$ under the discrete latent model.
If $B^\top w=\widehat\lambda w$, the analogous continuous latent observable evolves by $e^{t\widehat\lambda}$.
Under exact encoder closure these are Koopman eigenfunctions on the represented distribution family; in the learned models they are candidate approximations, evaluated below against $h_k$.
The transpose is the ordinary transpose, consistent with the complex bilinear expression $w^\top\bar z$.
Only the known constant eigenfunction introduced by augmentation is removed from the computed eigenpairs.

The discrete operator is refitted by ridge regression after re-encoding all 10,240 adjacent training snapshot pairs.
Let $X$ contain augmented source encodings as columns, $Y$ the corresponding target encodings, and $\Theta=[K\ b]$.
The update is
\begin{equation}
    \widehat\Theta
    =\underset{\Theta}{\operatorname{argmin}}\,
    \|Y-\Theta X\|_F^2+\eta\|\Theta\|_F^2
    =YX^\top(XX^\top+\eta I)^{-1},
    \qquad \eta=10^{-3}.
    \label{eq:spectral-ridge}
\end{equation}
The residual term is a sum over snapshot pairs, and the penalty includes the affine offset.
The implementation evaluates this solution by a singular value decomposition in double precision.
The refitted dynamics are held fixed during each subsequent gradient update of the encoder and decoder.
The continuous model optimizes $A,c$ by differentiating through the matrix exponential.
Each ridge refit minimizes the regularized one-step latent residual for the current encoder.
This may help preserve harmonic structure as the learned observable space changes, although the different initialization procedures and dynamics update budgets prevent isolating its contribution from this comparison.
Table~\ref{tab:spectral-training} summarizes the optimization differences.

\begin{table}[!ht]
\centering
\caption{Training configurations for the spectral experiments. Pretraining weights are shared within each dataset. The initialization procedures and dynamics update budgets differ between the two parameterizations.}
\label{tab:spectral-training}
\small
\begingroup
\setlength{\tabcolsep}{4pt}
\renewcommand{\arraystretch}{1.12}
\begin{tabular*}{\linewidth}{@{\extracolsep{\fill}}>{\raggedright\arraybackslash}p{.28\linewidth}>{\raggedright\arraybackslash}p{.31\linewidth}>{\raggedright\arraybackslash}p{.31\linewidth}}
\toprule
Setting & DisKO (discrete) & DisKO (continuous) \\
\midrule
Encoder/decoder pretraining & 40,000 shared steps & 40,000 shared steps \\
Dynamics initialization & Full ridge fit & 6,000 gradient steps \\
Joint optimization & 10,000 steps & 10,000 steps \\
Dynamics update & Full ridge refit at each step & AdamW, learning rate $3\times10^{-4}$ \\
Dynamics regularization & Ridge $10^{-3}$ on $K,b$ & Zero weight decay on $A,c$ \\
Encoder/decoder optimizer & AdamW, learning rate $3\times10^{-4}$, weight decay $10^{-5}$ & Same \\
Joint training horizons & $1,2,4,8,16,32$ steps & Same \\
Training batch & 8 snapshot pairs, 1,024 particles per snapshot & Same \\
\bottomrule
\end{tabular*}
\endgroup
\end{table}

Continuous dynamics initialization uses horizons $1,2,4$ for its first 3,000 updates and $1,2,4,8,16,32$ for its remaining 3,000 updates.
The discrete run performs 10,002 closed-form fits including initialization; the continuous run performs 16,000 gradient updates to the dynamics.
The joint loss in Equation~\ref{eq:joint-objective} uses $(\lambda_{\mathrm{pred}},\lambda_{\mathrm{rec}},\lambda_{\mathrm{lat}},\lambda_{\mathrm{dist}})=(1,0.5,0.05,0.2)$, with $\beta=0.25$ and $\gamma=0.5$ in Equation~\ref{eq:distribution-training-loss}.
The distribution loss uses 128 generated samples, eight sampling steps, and 32 sliced projections; gradients are clipped at norm five.
These results compare the two complete training configurations and provide evidence from a single seed.

\subsection{Eigenvalue Approximation}
\label{app:eigenvalue-approximation}

We compare the eigenvalues of $K$ with those of $e^{\Delta t A}$ in the common operator domain,
\begin{equation}
    \widehat\rho_j^{\mathrm{disc}}\in\operatorname{spec}(K),
    \qquad
    \widehat\rho_j^{\mathrm{cont}}=e^{\Delta t\widehat\lambda_j},
    \quad\widehat\lambda_j\in\operatorname{spec}(A).
    \label{eq:spectral-common-domain}
\end{equation}
This comparison avoids the frequency ambiguity associated with taking a logarithm of a discrete eigenvalue.
For each system and model, a one-to-one assignment minimizes the total complex-plane distance between the six reference eigenvalues and the learned candidates.
Assignment uses eigenvalue distances alone.
The absolute error for a matched pair is $e_k=|\widehat\rho_k-\rho_k|$; its mean gives equal weight to each of the six target indices.

\begin{figure}[p]
\centering
\includegraphics[width=\linewidth]{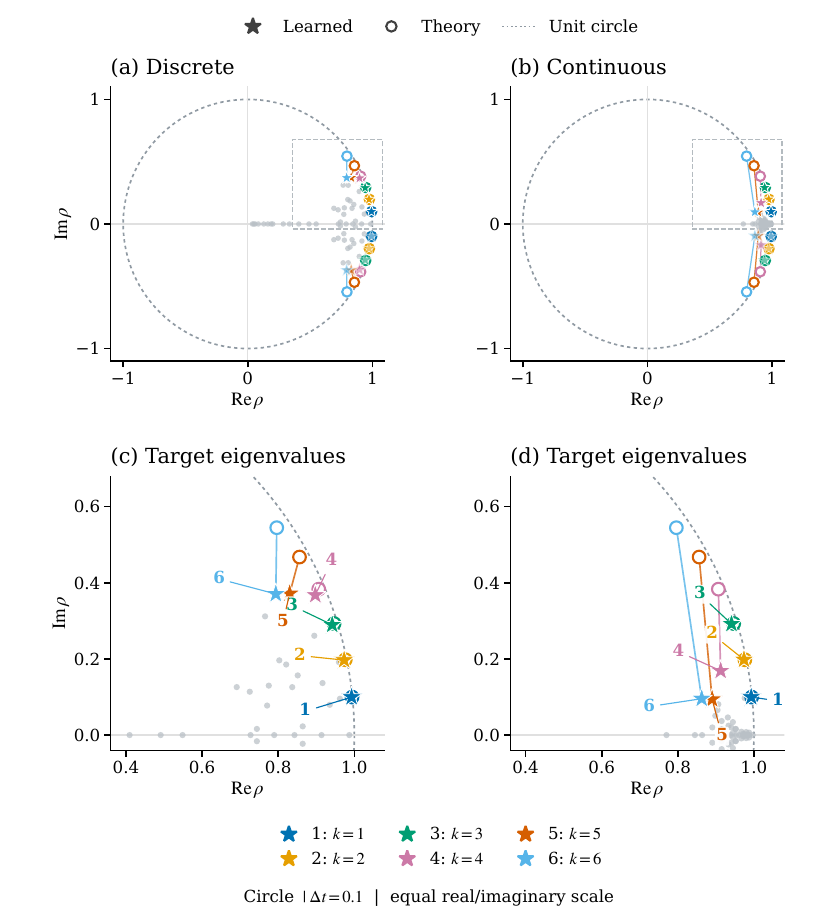}
\caption{Circle eigenvalue approximation at $\Delta t=0.1$. Discrete and Continuous denote the two DisKO parameterizations. Circles show the analytical eigenvalues, stars the matched learned values, and gray points the remaining learned eigenvalues. Upper panels include conjugate partners; lower panels enlarge the selected representatives. Colors index $k=1,\ldots,6$. Both axes have equal scale. A matched star denotes the assigned estimate for a reference value.}
\label{fig:spectrum-circle}
\end{figure}

\begin{figure}[p]
\centering
\includegraphics[width=\linewidth]{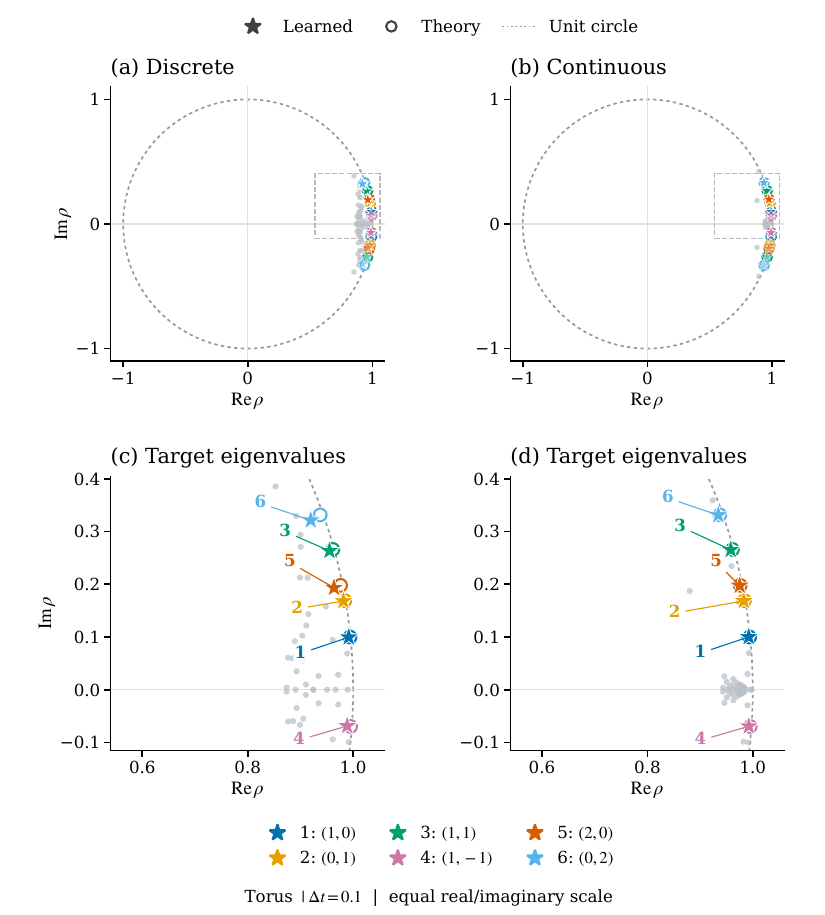}
\caption{Torus eigenvalue approximation at $\Delta t=0.1$, with the same conventions as Figure~\ref{fig:spectrum-circle}. The six Fourier indices identify the reference eigenpairs selected for evaluation. The gray points show the additional eigenvalues of the learned finite-dimensional operators.}
\label{fig:spectrum-torus}
\end{figure}

Table~\ref{tab:spectral-eigenvalues} reports all reference and estimated values.
Figures~\ref{fig:spectrum-circle} and~\ref{fig:spectrum-torus} show the full learned spectra together with the six reference pairs and detailed views of the selected representatives.
On Circle, both models approximate the first three eigenvalues with absolute errors below $0.006$.
At $k=4$, the errors are $0.0179$ for the discrete model and $0.2145$ for the continuous model; both models have larger errors at $k=5,6$.
The mean errors over all six indices are $0.050311$ and $0.175543$, respectively.
On Torus, the continuous model has smaller errors at every selected index, with mean error $0.002256$ compared with $0.008816$ for the discrete model.

\begin{table}[!ht]
\centering
\caption{Eigenvalues of $\mathcal K_{0.1}$ and their approximations. Each column corresponds to one prescribed Fourier index; conjugate partners are omitted. For the continuous model, $\widehat\rho_k=\exp(0.1\widehat\lambda_k)$. Complex values are rounded to four decimals, with real and imaginary parts on separate lines. Absolute errors are computed before rounding. All estimates use the final model from seed zero.}
\label{tab:spectral-eigenvalues}
\begingroup
\small
\setlength{\tabcolsep}{3pt}
\renewcommand{\arraystretch}{1.15}
\begin{tabular*}{\linewidth}{@{}l@{\extracolsep{\fill}}cccccc@{}}
\toprule
\multicolumn{7}{l}{\textbf{Circle}}\\
 & $k=1$ & $k=2$ & $k=3$ & $k=4$ & $k=5$ & $k=6$ \\
\midrule
Ground truth & \shortstack{$0.9940$\\$+0.0997\mathrm{i}$} & \shortstack{$0.9762$\\$+0.1979\mathrm{i}$} & \shortstack{$0.9468$\\$+0.2929\mathrm{i}$} & \shortstack{$0.9064$\\$+0.3832\mathrm{i}$} & \shortstack{$0.8559$\\$+0.4676\mathrm{i}$} & \shortstack{$0.7962$\\$+0.5447\mathrm{i}$} \\[3pt]
DisKO (discrete) & \shortstack{$0.9927$\\$+0.0996\mathrm{i}$} & \shortstack{$0.9731$\\$+0.1969\mathrm{i}$} & \shortstack{$0.9419$\\$+0.2895\mathrm{i}$} & \shortstack{$0.8974$\\$+0.3678\mathrm{i}$} & \shortstack{$0.8301$\\$+0.3715\mathrm{i}$} & \shortstack{$0.7937$\\$+0.3706\mathrm{i}$} \\[3pt]
DisKO (continuous) & \shortstack{$0.9927$\\$+0.0997\mathrm{i}$} & \shortstack{$0.9738$\\$+0.1981\mathrm{i}$} & \shortstack{$0.9410$\\$+0.2919\mathrm{i}$} & \shortstack{$0.9125$\\$+0.1688\mathrm{i}$} & \shortstack{$0.8901$\\$+0.0941\mathrm{i}$} & \shortstack{$0.8627$\\$+0.0956\mathrm{i}$} \\[3pt]
\midrule
Abs. error (discrete) & 0.001315 & 0.003191 & 0.005928 & 0.017900 & 0.099453 & 0.174081 \\
Abs. error (continuous) & 0.001314 & 0.002410 & 0.005868 & 0.214545 & 0.375084 & 0.454035 \\
\midrule
\multicolumn{7}{l}{\textbf{Torus}}\\
 & $k=(1,0)$ & $k=(0,1)$ & $k=(1,1)$ & $k=(1,-1)$ & $k=(2,0)$ & $k=(0,2)$ \\
\midrule
Ground truth & \shortstack{$0.9940$\\$+0.0997\mathrm{i}$} & \shortstack{$0.9842$\\$+0.1689\mathrm{i}$} & \shortstack{$0.9615$\\$+0.2661\mathrm{i}$} & \shortstack{$0.9952$\\$-0.0698\mathrm{i}$} & \shortstack{$0.9762$\\$+0.1979\mathrm{i}$} & \shortstack{$0.9375$\\$+0.3316\mathrm{i}$} \\[3pt]
DisKO (discrete) & \shortstack{$0.9916$\\$+0.0995\mathrm{i}$} & \shortstack{$0.9814$\\$+0.1682\mathrm{i}$} & \shortstack{$0.9550$\\$+0.2634\mathrm{i}$} & \shortstack{$0.9886$\\$-0.0689\mathrm{i}$} & \shortstack{$0.9636$\\$+0.1933\mathrm{i}$} & \shortstack{$0.9196$\\$+0.3212\mathrm{i}$} \\[3pt]
DisKO (continuous) & \shortstack{$0.9923$\\$+0.1001\mathrm{i}$} & \shortstack{$0.9830$\\$+0.1683\mathrm{i}$} & \shortstack{$0.9581$\\$+0.2653\mathrm{i}$} & \shortstack{$0.9925$\\$-0.0694\mathrm{i}$} & \shortstack{$0.9750$\\$+0.1974\mathrm{i}$} & \shortstack{$0.9345$\\$+0.3311\mathrm{i}$} \\[3pt]
\midrule
Abs. error (discrete) & 0.002404 & 0.002892 & 0.006962 & 0.006614 & 0.013348 & 0.020678 \\
Abs. error (continuous) & 0.001781 & 0.001377 & 0.003430 & 0.002690 & 0.001239 & 0.003020 \\
\bottomrule
\end{tabular*}
\endgroup
\end{table}

\FloatBarrier

\clearpage

\subsection{Eigenfunction Evaluation on Distributions}
\label{app:eigenfunction-evaluation}

The learned function associated with a matched eigenvalue is evaluated as $u_k(\widehat\nu)=w_k^\top[E(\widehat\nu)^\top,1]^\top$.
Its eigenvector has an arbitrary nonzero complex scale.
For visual comparison, we fit one complex scalar per eigenfunction using 640 training snapshots, comprising all 128 training sequences at times $0,2,4,6,8$,
\begin{equation}
    \alpha_k=\frac{\sum_{s\in\mathcal C}\overline{u_k(\widehat\nu_s)}h_k(\mu_s)}
                       {\sum_{s\in\mathcal C}|u_k(\widehat\nu_s)|^2},
    \qquad \widehat h_k=\alpha_k u_k.
    \label{eq:spectral-alignment}
\end{equation}
Here $\mathcal C$ is the calibration set; the fit has no intercept and is fixed for all subsequent plots.
This post-training alignment resolves eigenvector amplitude and phase for evaluation.
Analytical values are used only for this alignment and the spectral diagnostics.

\begin{figure}[p]
\centering
\includegraphics[width=\linewidth]{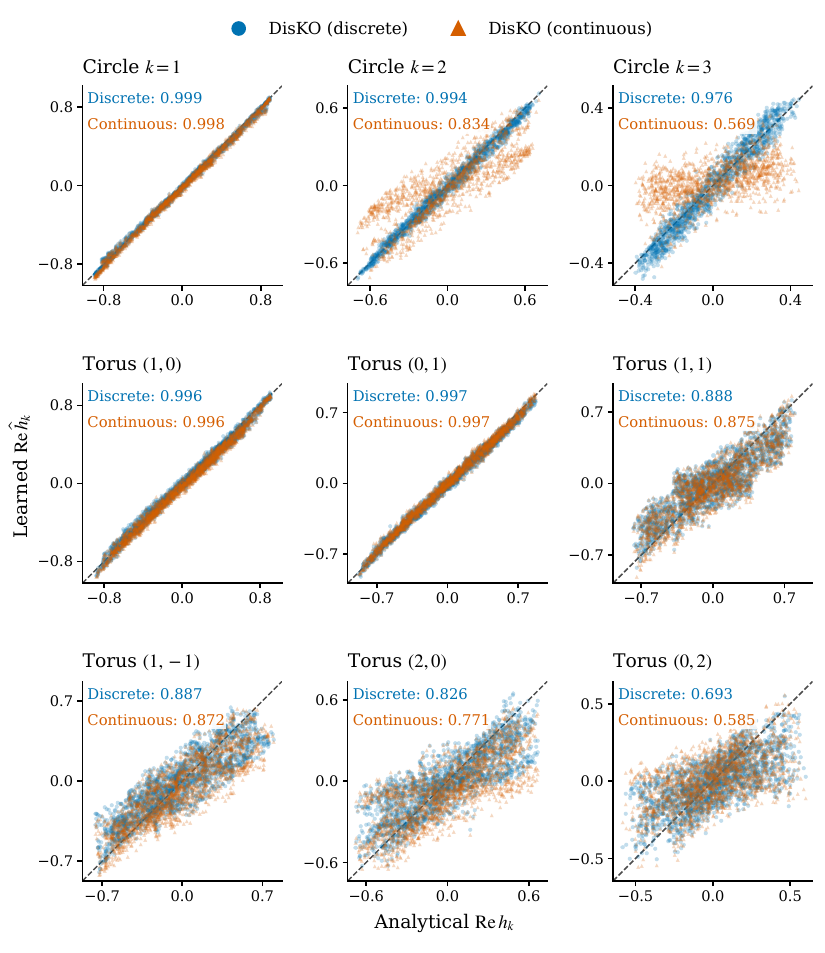}
\caption{Eigenfunction values on observed future snapshots from test sequences. The top row shows Circle indices $k=1,2,3$; the remaining rows show all six Torus indices. Each panel overlays DisKO (discrete) and DisKO (continuous), with the equality line dashed. Coordinates display real parts after the training calibration; annotations give centered complex correlations. Each point is obtained by encoding the observed snapshot at that time, so the figure assesses the learned functions on future distributions.}
\label{fig:spectral-eigenfunction-values}
\end{figure}

Figure~\ref{fig:spectral-eigenfunction-values} compares the real parts of $\widehat h_k(\widehat\nu_t)$ and $h_k(\mu_t)$ on test sequences at $8<t\le12$.
Each Circle panel contains $32\times40=1,280$ snapshots and each Torus panel contains $48\times40=1,920$ snapshots per method.
The displayed correlation is computed from the full complex values,
\begin{equation}
    C_k=\frac{\left|\sum_s\overline{(u_{k,s}-\bar u_k)}(h_{k,s}-\bar h_k)\right|}
                  {\left(\sum_s|u_{k,s}-\bar u_k|^2\right)^{1/2}
                   \left(\sum_s|h_{k,s}-\bar h_k|^2\right)^{1/2}}.
    \label{eq:spectral-complex-correlation}
\end{equation}
The plotted points use independently encoded observed snapshots at each time and assess eigenfunction approximation on these distributions.
For Circle, future correlations for the discrete model are $0.999$, $0.994$, and $0.976$ at $k=1,2,3$, compared with $0.998$, $0.834$, and $0.569$ for the continuous model.
For Torus, both models have correlations above $0.995$ for the two fundamental indices, while higher and mixed indices have lower correlations.
Across all six Torus indices, the mean future correlations are $0.881$ and $0.849$, respectively.
Together with the eigenvalue errors, these results show that the quality of the scalar eigenfunctions and the accuracy of their eigenvalues can differ between parameterizations.

The remaining figures evaluate the same learned functions on a controlled family of distributions.
For a wrapped Gaussian $\mu_{\vartheta,\sigma}$ with center $\vartheta$ and angular covariance $\sigma^2 I$, its Fourier moment is
\begin{equation}
    h_k(\mu_{\vartheta,\sigma})
    =\exp\!\left(-\tfrac12\sigma^2\|k\|_2^2\right)e^{\mathrm{i}k\cdot\vartheta},
    \qquad \sigma=0.15.
    \label{eq:spectral-probe-reference}
\end{equation}
Each center in the plot therefore indexes a probability distribution supplied to the encoder.
Probes use 1,024 common Gaussian offsets obtained from Sobol normal nodes and their negatives, with 256 Circle centers and a $64\times64$ uniform angular grid on Torus.
The training calibration in Equation~\ref{eq:spectral-alignment} is reused without fitting to the probes.
The normalized root mean square error on the probe grid $\mathcal P$ is
\begin{equation}
    \operatorname{NRMSE}_k=
    \left(\frac{\sum_{\vartheta\in\mathcal P}
    |\widehat h_k(\widehat\nu_{\vartheta,\sigma})-h_k(\mu_{\vartheta,\sigma})|^2}
    {\sum_{\vartheta\in\mathcal P}|h_k(\mu_{\vartheta,\sigma})|^2}\right)^{1/2}.
    \label{eq:spectral-probe-nrmse}
\end{equation}

\begin{figure}[p]
\centering
\includegraphics[width=\linewidth]{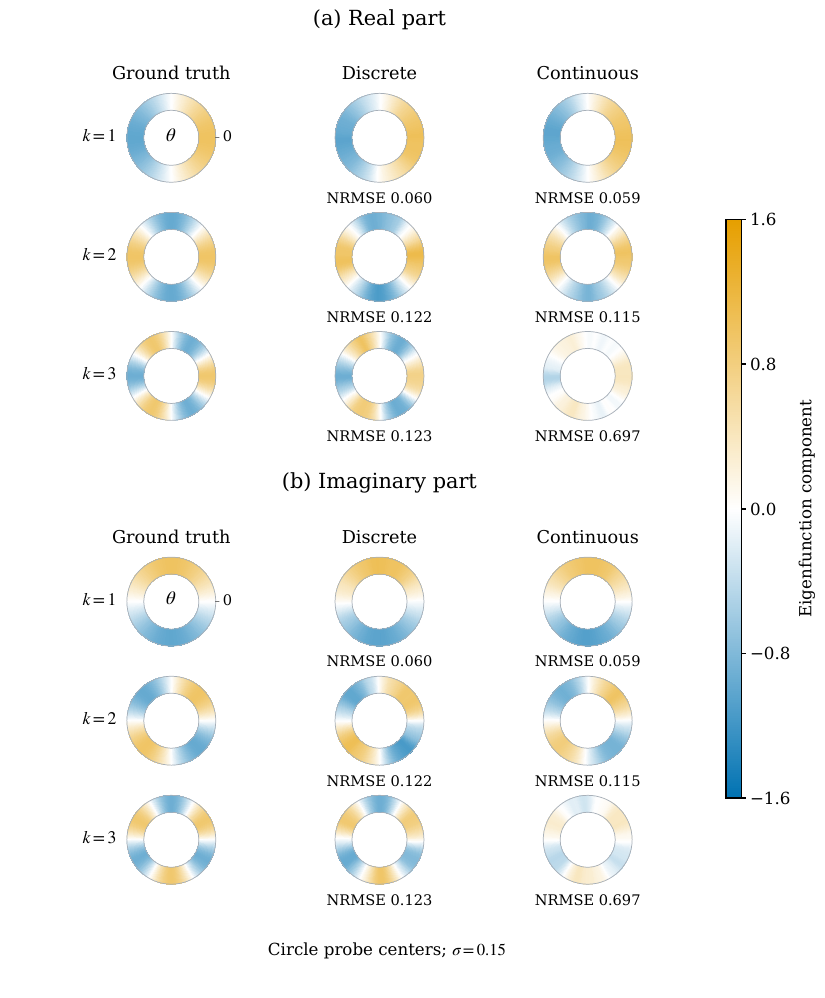}
\caption{Circle eigenfunctions evaluated on translated wrapped Gaussian probes. Panels (a) and (b) display real and imaginary components using the same color scale. Each angle is the center of a distribution of width $\sigma=0.15$; ring radius and thickness are fixed. Columns compare the analytical reference with the discrete and continuous DisKO functions. NRMSE uses the full complex error and is consequently the same in both component panels.}
\label{fig:spectral-circle-probes}
\end{figure}

\begin{figure}[p]
\centering
\textbf{(a) Real part}\par
\includegraphics[width=\linewidth]{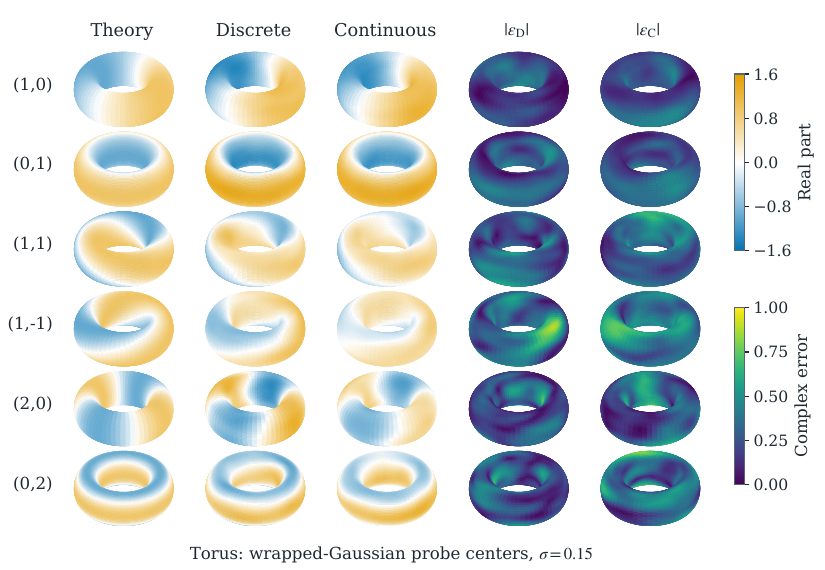}
\textbf{(b) Imaginary part}\par
\includegraphics[width=\linewidth]{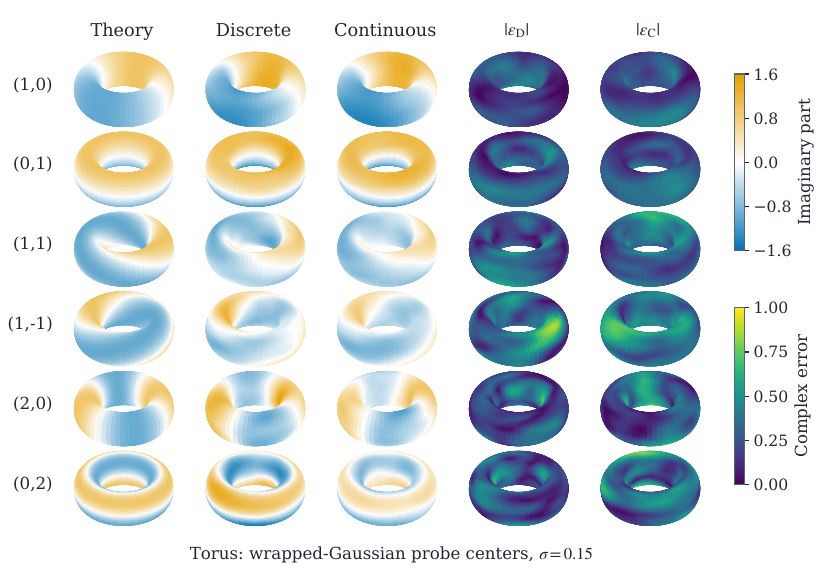}
\caption{Torus eigenfunctions on wrapped Gaussian probes. Rows are the six Fourier indices; the first three columns show the reference and the two DisKO approximations. The final two columns show $|\widehat h_k-h_k|$ for the discrete and continuous models. These complex error fields are identical in panels (a) and (b). All color ranges are shared, and surface coordinates index probe centers.}
\label{fig:spectral-torus-probes}
\end{figure}

Figure~\ref{fig:spectral-circle-probes} displays both components for Circle indices $k=1,2,3$.
At $k=3$, the probe NRMSE is $0.123$ for the discrete model and $0.697$ for the continuous model, despite their similar eigenvalue errors.
Figure~\ref{fig:spectral-torus-probes} displays all six Torus indices, including the pointwise complex absolute errors.
The real and imaginary components use a common color range across both systems and methods.
Torus geometry is used only to display the two angular center coordinates; the encoder receives four-dimensional point clouds.
Errors are evaluated uniformly over the angular grid.
These probes characterize the restrictions of learned distribution observables to the chosen Gaussian family.
Diffusion increases the Gaussian width over time, so a static comparison on the fixed-width family supports conclusions about function approximation on that family.
The nonlinear set encoder permits more general distribution observables than linear Fourier moments, and accurate approximation across all probability measures is outside the scope of these finite evaluations.
\FloatBarrier

\clearpage

\clearpage
\section{Latent Dynamical Consistency}
\label{app:latent-dynamics}
\setcounter{figure}{0}

We assess approximate closure in Equation~\ref{eq:observable-propagation} by comparing latent forecasts with encodings of observed snapshots.
Exact closure with the corresponding affine generator implies agreement at the population level.

We display the first test population sequence of the OU benchmark (Appendix~\ref{app:experimental-setup}), starting from a Gaussian distribution.
Observations are spaced by $0.05$, training covers $0\le t\le2.5$, and extrapolation covers $2.5<t\le5$.
At each time, 512 particles are sampled without replacement from the available 1,024 particles and standardized using statistics fitted to training populations within the training interval.
The same sampled snapshots are supplied to five models with seeds $s=0,\ldots,4$.
Each Deep Sets encoder produces a 32-dimensional representation.
After training, the encoder and continuous affine dynamics are fixed, and we compute
\begin{equation}
    z_s^{\mathrm{enc}}(t)=E_s(\widehat\nu_t),
    \qquad
    \widetilde z_s(t)=F_t^{(s)}\!\left(E_s(\widehat\nu_0)\right).
    \label{eq:latent-consistency-trajectories}
\end{equation}
Here $F_t^{(s)}$ is evaluated through the matrix exponential in Equation~\ref{eq:augmented-latent-flow}.
Every forecast originates at $t=0$ and continues through the training boundary; subsequent snapshot encodings serve solely as references.

\begin{figure}[!ht]
    \centering
    \includegraphics[width=\linewidth]{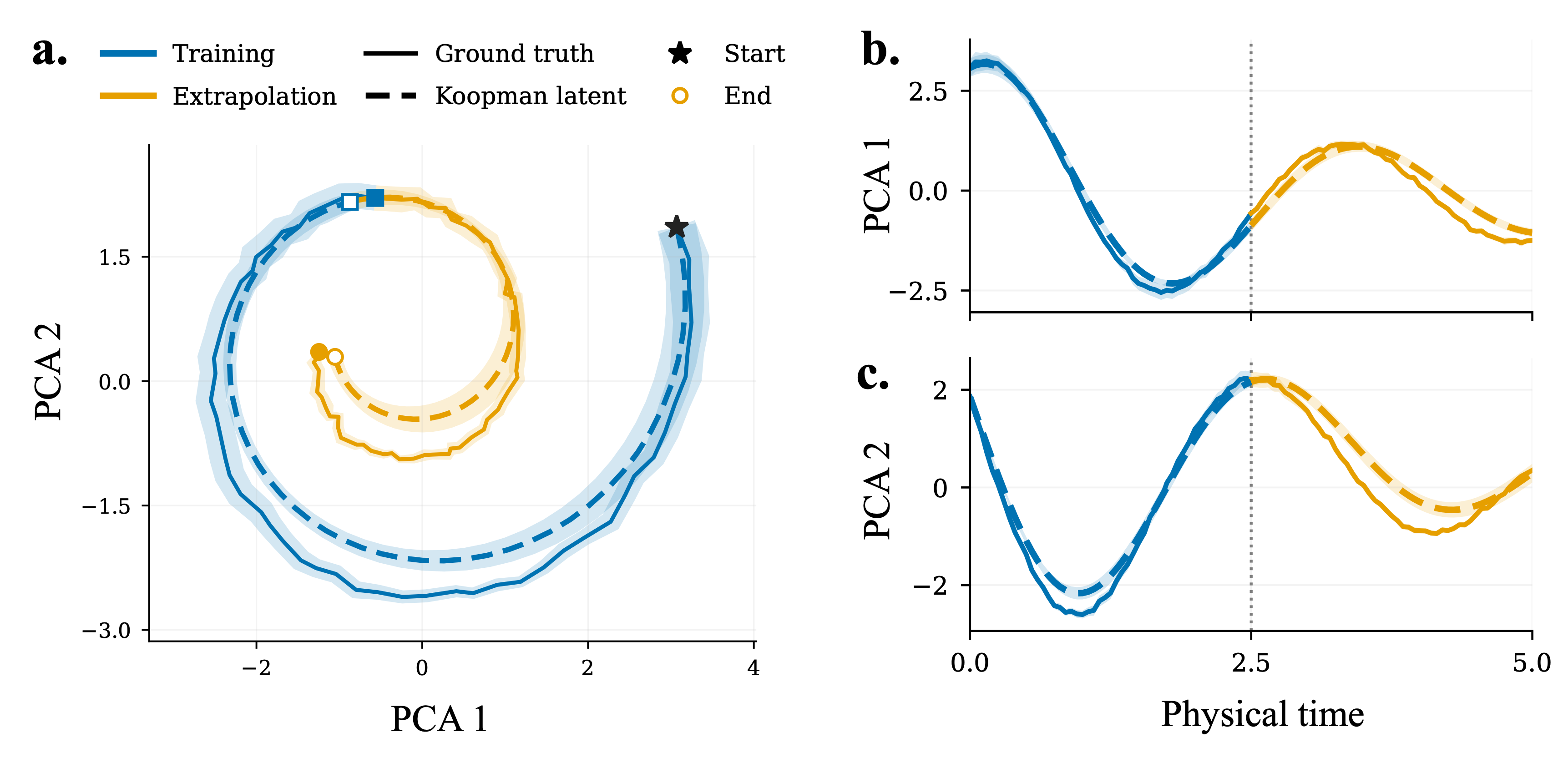}
    \caption{OU latent evolution. Solid curves (Ground truth) show reference encodings; dashed curves (Koopman latent) show forecasts. Blue and gold indicate training and future time ranges; all snapshots belong to the test population. Curves are means over five models after PCA and rigid alignment. Shading is one sample standard deviation, normal to each mean curve in (a) and coordinatewise in (b,c). Stars, squares, and circles mark $t=0$, $2.5$, and $5$; filled and open markers denote reference and forecast.}
    \label{fig:ou-latent-dynamics}
\end{figure}

For each model, we jointly fit PCA to the centered reference and forecast trajectories, retaining two components without whitening.
Reference trajectories are aligned to seed zero by an orthogonal transformation and translation, with the same transformation applied to forecasts and no rescaling.
Both visualization steps use the full interval after prediction.
The retained components explain $77.66\%$ to $83.51\%$ of joint variance, depending on the model.
Figure~\ref{fig:ou-latent-dynamics} shows means and sample standard deviations in these aligned coordinates.

The forecast follows the reference's rotation and inward progression, reproducing the main oscillation within the training interval and continuing the contracting trajectory beyond $t=2.5$.
Amplitude differences are visible near the troughs of the second coordinate.
The clearest future discrepancy occurs near $t=4$, where the predicted minimum is shallower and slightly later.
These observations support approximate dynamical consistency in the dominant projected coordinates for this test sequence, with residual amplitude and timing errors.
Projection and averaging restrict this evidence to the displayed dynamics.

\clearpage
\section{Ablation Study}
\label{app:ablation}
\setcounter{figure}{0}

We examine how the form of latent dynamics affects distribution extrapolation on the OU benchmark.
The continuous DisKO model uses the affine generator $\dot z=Az+c$, evaluated through the matrix exponential in Equation~\ref{eq:augmented-latent-flow}.
For comparison, we replace this generator with an autonomous Neural ODE,
\begin{equation}
    \dot z=W_2\tanh(W_1z+b_1)+b_2,
    \label{eq:ablation-neural-ode}
\end{equation}
with latent dimension 64 and hidden width 32.
The two dynamics modules have similar parameter counts, 4,160 for the affine generator and 4,192 for the Neural ODE.
Both models use the same Deep Sets encoder and conditional flow matching decoder architectures, with width 256 and depths three and five, respectively.

We use 128 training initial distributions and all 64 test initial distributions from the OU dataset in Appendix~\ref{app:datasets}; 16 further initial distributions are reserved for validation.
Training covers $0\le t\le2.5$, with snapshots spaced by $0.05$ and 1,024 particles per snapshot.
Standardization is fitted to the training distributions within this interval.
For each of five training seeds, the models share encoder and decoder weights obtained from 10,000 pretraining updates.
Each then receives 1,000 updates of the dynamics alone, followed by 10,000 joint updates of all components.
Both dynamics modules are optimized by gradient descent with the same losses, minibatch sampling, and learning rate $3\times10^{-4}$.
We evaluate the final checkpoints.

Each test rollout encodes only the initial snapshot and propagates recursively to $t=5$, without incorporating subsequent observations.
The Neural ODE uses fourth-order Runge--Kutta integration with maximum step $0.05$.
Halving this step changes latent trajectories by at most $3.1\times10^{-6}$ in relative Euclidean norm across the five models.
Both decoders generate 512 samples with 32 sampling steps, and errors are computed against 512 reference samples in the original observation coordinates.
We use 256 samples per distribution for $W_1$ and $W_2$, 128 projection directions for $\mathrm{SW}_1$, and the biased RBF estimate of $\mathrm{MMD}^2$ defined in Appendix~\ref{app:evaluation-metrics}.
Reference samples and evaluation random seeds are matched between methods.

\begin{figure}[!ht]
    \centering
    \includegraphics[width=\linewidth]{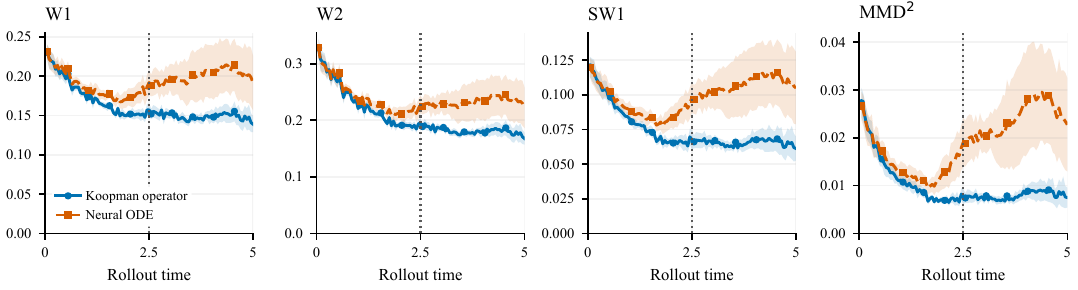}
    \caption{OU distribution prediction errors for the continuous Koopman model and the Neural ODE. At each time, errors are averaged over 64 test initial distributions within each run, then over five training seeds. Shading denotes one sample standard deviation across seeds. The dotted vertical line marks the end of training at $t=2.5$. All predictions originate at $t=0$.}
    \label{fig:koopman-node}
\end{figure}

Figure~\ref{fig:koopman-node} shows lower future prediction errors and smaller variation across training seeds for the Koopman model in all four metrics.
Averaged over $2.5<t\le5$, its $W_1$ error is $0.149\pm0.007$, compared with $0.201\pm0.026$ for the Neural ODE; the corresponding $\mathrm{SW}_1$ errors are $0.065\pm0.005$ and $0.107\pm0.018$.
The Neural ODE exhibits a more pronounced increase in $\mathrm{SW}_1$ and $\mathrm{MMD}^2$ near the training boundary and into the future interval.
These results support the affine Koopman parameterization as an effective inductive bias for accurate and reproducible extrapolation under this training protocol.

\clearpage
\section{Data Efficiency}
\label{app:data-efficiency}
\setcounter{figure}{0}

We examine the data requirements of continuous DisKO on the OU benchmark by varying two aspects of the training data.
Let $R=|\mathcal I_{\mathrm{train}}|$ denote the number of training initial distributions, each providing a population sequence of snapshots, and let $N$ denote the number of available particle samples per training snapshot.
Thus, $R$ controls the number of distribution evolutions represented in training, while $N$ controls the sampling budget within each snapshot.

We vary $R\in\{16,32,64,128\}$ using 1,024 particles per snapshot input during training and evaluation.
Separately, we vary $N\in\{128,256,512,1024\}$ with $R=128$, keeping each training and evaluation input at 128 particles.
In the latter experiment, $N$ specifies the available training sample pool from which inputs are drawn.
The two experiments are interpreted within their respective input budgets.
Each setting uses three training seeds, and the reported scores average prediction errors over the future time window.

\begin{figure}[!ht]
    \centering
    \includegraphics[width=\linewidth]{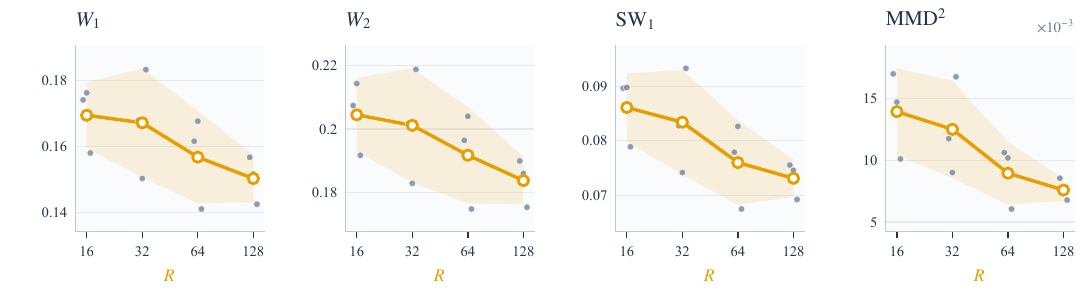}
    \caption{Effect of the number of training initial distributions $R$ on future OU prediction errors. Colored circles and lines show means over three training seeds; shading denotes one sample standard deviation. Gray points show individual seeds, slightly displaced horizontally for visibility. Inputs contain 1,024 particles. The horizontal axis is logarithmic with base two.}
    \label{fig:data-efficiency-r}
\end{figure}

Figure~\ref{fig:data-efficiency-r} shows decreasing mean errors in all four metrics as $R$ increases.
From $R=16$ to $R=128$, $W_1$ decreases from $0.169\pm0.010$ to $0.150\pm0.007$, and $\mathrm{SW}_1$ decreases from $0.086\pm0.006$ to $0.073\pm0.003$.
This trend supports the benefit of exposing the model to more initial distributions.

\begin{figure}[!ht]
    \centering
    \includegraphics[width=\linewidth]{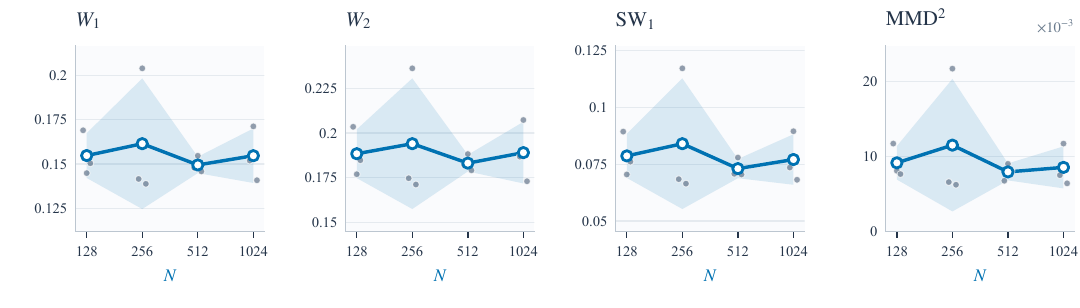}
    \caption{Effect of the available samples per training snapshot $N$, with $R=128$ training initial distributions and 128 particles per input. Means, sample standard deviations, and individual seeds follow Figure~\ref{fig:data-efficiency-r}. Errors summarize the future time window. The horizontal axis is logarithmic with base two; the $\mathrm{MMD}^2$ axis in both figures uses units of $10^{-3}$.}
    \label{fig:data-efficiency-n}
\end{figure}

Figure~\ref{fig:data-efficiency-n} shows modest, nonmonotonic changes with $N$ under the fixed input budget.
Mean $W_1$ ranges from $0.149$ to $0.161$, with the lowest mean error in all four metrics at $N=512$.
Variation across seeds is largest at $N=256$.
Over the tested range, increasing the available sample pool yields limited additional accuracy with 128 particles per input.

\clearpage
\begingroup
\providecommand{\Dom}{\operatorname{Dom}}
\providecommand{\spanof}{\operatorname{span}}
\newtheorem{theoryproposition}{Proposition}[section]
\newenvironment{theoryproof}[1][Proof]{%
  \par\addvspace{\topsep}\noindent\textit{#1.}\enspace\ignorespaces
}{%
  \unskip\nobreak\hfill\ensuremath{\square}\par\addvspace{\topsep}
}
\setcounter{equation}{0}
\renewcommand{\theequation}{\thesection.\arabic{equation}}
\renewcommand{\theHequation}{\thesection.\arabic{equation}}

\section{Distributional Koopman Representation}
\label{app:revised-theory}

We describe the distributional Koopman representation underlying DisKO, the
role of learned distribution observables, and the information required for
predictive closure. The results separate the expressiveness of an observable
family from the dynamical consistency of its learned finite-dimensional
representation.

\subsection{Distribution Observables and Koopman Operators}
\label{app:revised-operators}

Let $\mathcal X$ be a state space and $\mathfrak M\subseteq\mathcal P(\mathcal X)$
an invariant family of probability distributions. Distribution evolution is
described by maps $T_t:\mathfrak M\to\mathfrak M$ satisfying
\begin{equation}
    T_0=\operatorname{Id},\qquad
    T_{t+s}=T_t\circ T_s,\qquad t,s\geq0.
    \label{eq:rev-law-semigroup}
\end{equation}
Autonomous ODEs induce such evolution by transporting an initial distribution
through their solution maps. Time-homogeneous Markov SDEs induce it through
their transition probabilities. In both cases, the initial law determines
the subsequent law, even when individual state trajectories are stochastic.

A distribution observable is a scalar functional $h:\mathfrak M\to\mathbb C$.
Complex values accommodate oscillatory eigenfunctions; encoder coordinates
are real valued. The distributional Koopman operator acts by composition
\citep{oprea2025distributional},
\begin{equation}
    (\mathcal K_t h)(\mu)=h(T_t\mu).
    \label{eq:rev-koopman}
\end{equation}
It is linear in $h$ and satisfies
$\mathcal K_{t+s}=\mathcal K_t\mathcal K_s$.
Thus, nonlinear distribution evolution admits a linear description on a
space of distribution observables.
On an invariant Banach space $\mathcal H$ where $(\mathcal K_t)_{t\geq0}$ is
strongly continuous, its generator is
\begin{equation}
    \mathcal Lh=\lim_{t\downarrow0}\frac{\mathcal K_t h-h}{t},
    \qquad h\in\Dom(\mathcal L),
    \label{eq:rev-generator}
\end{equation}
with the limit taken in the norm of $\mathcal H$.
For example, when $\mathfrak M$ is compact and $(t,\mu)\mapsto T_t\mu$ is
continuous, $\mathcal H=C(\mathfrak M;\mathbb C)$ with the uniform norm gives
a strongly continuous contraction semigroup.

The connection to state observables is explicit. Let
$(S_t\phi)(x)=\mathbb E[\phi(X_t)\mid X_0=x]$ be a Markov semigroup on state
functions. For $h_\phi(\mu)=\int\phi\,d\mu$, its induced law evolution obeys
\begin{equation}
    (\mathcal K_t h_\phi)(\mu)
    =\int S_t\phi\,d\mu=h_{S_t\phi}(\mu).
    \label{eq:rev-moment-intertwining}
\end{equation}
This identity relates state observables to linear functionals of a
distribution \citep{oprea2025distributional}.
Distribution observables also include nonlinear statistics, such as
$\Var_\mu[\phi]=h_{\phi^2}(\mu)-h_\phi(\mu)^2$ whenever these moments exist.

\subsection{Learned Observable Spaces and Approximate Closure}
\label{app:revised-observables}

\paragraph{Nonlinear moment representations.}
Consider functionals formed from finitely many moments,
\begin{equation}
    h(\mu)=g\bigl(a_1(\mu),\ldots,a_q(\mu)\bigr),
    \qquad a_\ell(\mu)=\int\phi_\ell\,d\mu.
    \label{eq:rev-cylindrical}
\end{equation}
The readout $g$ permits nonlinear interactions between statistics of the
entire distribution. Their expressive power has a precise interpretation in
a compact setting. Assume that $\mathcal X$ is a compact metric space,
$\mathfrak M=\mathcal P(\mathcal X)$ carries the weak topology, and $S_t$ is a
strongly continuous Markov semigroup on $C(\mathcal X)$ with generator
$\mathcal A$. Its dual law evolution induces a strongly continuous
$\mathcal K_t$ on $C(\mathfrak M)$.
Let $\mathfrak P$ contain all functionals in
Equation~\ref{eq:rev-cylindrical} with arbitrary finite $q$, polynomial $g$,
and real-valued $\phi_\ell\in\Dom(\mathcal A)$, together with their complex
linear combinations.

\begin{theoryproposition}[Richness of polynomial moment observables]
\label{prop:rev-moment-core}
Under these assumptions, $\mathfrak P$ is dense in $C(\mathfrak M;\mathbb C)$
and is a core for $\mathcal L$. In particular, every
$h\in\Dom(\mathcal L)$ admits $p_n\in\mathfrak P$ with
$\|p_n-h\|_\infty\to0$ and
$\|\mathcal Lp_n-\mathcal Lh\|_\infty\to0$.
\end{theoryproposition}

\begin{theoryproof}[Proof sketch]
(Adapted from the core criterion in
\citealp[Proposition~II.1.7]{engel2000semigroups}.)
The moment algebra contains constants and separates probability measures
because $\Dom(\mathcal A)$ is dense in $C(\mathcal X)$; the
Stone--Weierstrass theorem gives density.
For a polynomial $g$, the uniform generator limit and the chain rule give
\begin{equation}
    (\mathcal Lh)(\mu)=
    \sum_{\ell=1}^q
    \partial_\ell g\bigl(a_1(\mu),\ldots,a_q(\mu)\bigr)
    \int\mathcal A\phi_\ell\,d\mu.
    \label{eq:rev-cylindrical-generator}
\end{equation}
Thus $\mathfrak P\subseteq\Dom(\mathcal L)$.
Equation~\ref{eq:rev-moment-intertwining} and
$S_t\Dom(\mathcal A)\subseteq\Dom(\mathcal A)$ imply
$\mathcal K_t\mathfrak P\subseteq\mathfrak P$.
The cited criterion then establishes the core property.
\end{theoryproof}

This result makes nonlinear moment functionals a rich starting point for
distribution representations. Its scope is the complete moment family under
the compact Markov assumptions above. A fixed finite dictionary, or a
trained neural encoder, requires a separate assessment of approximation and
closure.

\paragraph{Learned distribution observables.}
An encoder $E$ defines each observable by $h_j(\mu)=E_j(\mu)$ and collects
them in $\mathbf h=(h_1,\ldots,h_m)^\top$.
For an empirical snapshot
$\widehat\nu=N^{-1}\sum_{i=1}^N\delta_{y_i}$, a moment encoder evaluates
Equation~\ref{eq:rev-cylindrical} by replacing each integral with
$N^{-1}\sum_i\phi_\ell(y_i)$.
Neural mean aggregation admits approximation results for continuous
functionals on compact families of probability measures
\citep{pevny2019approximation}. DisKO parameterizes its coordinates with
permutation-invariant set networks \citep{zaheer2017deep,lee2019set} and
learns them jointly with latent propagation and distribution reconstruction.
This gives a data-driven finite-dimensional observable space
\begin{equation}
    V_E=\spanof\{1,h_1,\ldots,h_m\}.
    \label{eq:rev-observable-space}
\end{equation}
The approximation results motivate expressive parameterizations.
The dynamical interpretation of the learned space depends additionally on
its behavior under $\mathcal K_t$.

\paragraph{Affine propagation and closure.}
The continuous DisKO model uses $\dot z=Az+c$ with flow $F_t$.
Including the constant observable yields
\begin{equation}
    \overline{\mathbf h}(\mu)=
    \begin{pmatrix}\mathbf h(\mu)\\1\end{pmatrix},
    \qquad B=\begin{pmatrix}A&c\\0&0\end{pmatrix},
    \qquad
    \begin{pmatrix}F_t(z)\\1\end{pmatrix}
    =e^{tB}\begin{pmatrix}z\\1\end{pmatrix}.
    \label{eq:rev-augmented}
\end{equation}
If $h_j\in\Dom(\mathcal L)$ and
$\mathcal L\mathbf h=A\mathbf h+c$ on $\mathfrak M$, uniqueness of the
resulting linear ODE gives
$\overline{\mathbf h}(T_t\mu)=e^{tB}\overline{\mathbf h}(\mu)$.
The space $V_E$ is then Koopman invariant, realizing exact closure in the
chosen coordinates \citep{brunton2016invariant}.

For learned coordinates, define the finite-time closure residual
\begin{equation}
    R_t(\mu)=\overline{\mathbf h}(T_t\mu)
                -e^{tB}\overline{\mathbf h}(\mu).
    \label{eq:rev-closure-residual}
\end{equation}
This definition only requires observable values and can be evaluated from
distribution snapshots. Small residuals across relevant distributions and
prediction times express approximate closure under the learned propagation.
Latent consistency training estimates this discrepancy at observed time
pairs using empirical encodings. Its behavior on future snapshots provides
a separate test of dynamical consistency.

\paragraph{Eigenfunctions within the learned space.}
Let $w^\top B=\lambda w^\top$ be a left eigenvector relation over
$\mathbb C$, and define $h_w(\mu)=w^\top\overline{\mathbf h}(\mu)$, assuming
$h_w\not\equiv0$. Multiplying Equation~\ref{eq:rev-closure-residual} by
$w^\top$ gives
\begin{equation}
    (\mathcal K_t h_w)(\mu)-e^{t\lambda}h_w(\mu)
    =w^\top R_t(\mu).
    \label{eq:rev-eigenfunction-residual}
\end{equation}
Here $\top$ denotes ordinary transpose, so $w$ specifies the coefficients
of a complex linear combination of real observables.
Exact closure therefore produces distributional Koopman eigenfunctions.
Under approximate closure, these combinations are candidate eigenfunctions
whose evolution can be checked through Equation~\ref{eq:rev-eigenfunction-residual}.
For $\lambda=-\alpha+i\omega$, $\alpha>0$ gives exponential decay and
$\omega$ gives angular oscillation frequency.

The discrete model has the same interpretation with
$\overline K=\bigl(\begin{smallmatrix}K&b\\0&1\end{smallmatrix}\bigr)$,
$\overline K^n$ replacing $e^{tB}$ at $t=n\Delta t$, and an eigenvalue
$\rho$ replacing $e^{\Delta t\lambda}$.
Eigenvectors select combinations within $V_E$; individual encoder
coordinates generally combine several dynamical components.
Empirical agreement with analytic eigenfunctions consequently tests the
dynamical content of the learned observable space.
The residual identity alone supplies no bound on the distance to an exact
eigenvalue.

\paragraph{Connection to distribution prediction.}
The decoder $D$ is trained to reconstruct the modeled distribution from
$E(\mu)$ and to generate its future law from $F_t(E(\mu))$.
Exact reconstruction on a family requires $E$ to be injective there.
A continuous injective encoder on a compact family has a continuous inverse
on its image. This explains the complementary roles of reconstruction and
closure in retaining distributional information and propagating it
consistently; practical decoder accuracy remains an empirical property.

\subsection{Observation Sufficiency and Identifiability}
\label{app:revised-identifiability}

Suppose that a measurable observation map $P:\mathcal X\to\mathcal Y$
provides the law $\nu=P_\#\mu$, where $P_\#$ denotes pushforward.
The observed family is $\mathfrak N=P_\#\mathfrak M$.
Full observation corresponds to $P=\operatorname{Id}$; partial observation
can merge distinct full distributions into the same observed law.

\begin{theoryproposition}[Sufficiency for autonomous observed evolution]
\label{prop:rev-observation}
There is a semigroup $\overline T_t:\mathfrak N\to\mathfrak N$ satisfying
$\overline T_t(P_\#\mu)=P_\#T_t\mu$ for all $\mu\in\mathfrak M$ and $t\geq0$
if and only if
\begin{equation}
    P_\#\mu=P_\#\mu'
    \quad\Longrightarrow\quad
    P_\#T_t\mu=P_\#T_t\mu'
    \quad\text{for every }t\geq0.
    \label{eq:rev-sufficiency}
\end{equation}
\end{theoryproposition}

\begin{theoryproof}
An observed evolution must assign identical futures to identical inputs,
which proves necessity. Under Equation~\ref{eq:rev-sufficiency}, define
$\overline T_t(\nu)=P_\#T_t\mu$ using any $\mu$ with $P_\#\mu=\nu$.
The condition makes this definition independent of the chosen preimage.
The identity and composition laws follow from those of $T_t$.
\end{theoryproof}

This condition concerns autonomous evolution on observed laws.
Generator statements additionally require the strong continuity conditions
in Section~\ref{app:revised-operators}.
If Equation~\ref{eq:rev-sufficiency} fails at time $t$, any predictor $Q_t$
using only the current observed law has an unavoidable ambiguity.
For a metric $d$ on the predicted distributions and the common input
$\nu=P_\#\mu=P_\#\mu'$, the triangle inequality gives
\begin{equation}
    \max\!\left\{
      d\bigl(Q_t(\nu),P_\#T_t\mu\bigr),
      d\bigl(Q_t(\nu),P_\#T_t\mu'\bigr)
    \right\}
    \geq \frac12 d\bigl(P_\#T_t\mu,P_\#T_t\mu'\bigr).
    \label{eq:rev-predictive-ambiguity}
\end{equation}
This obstruction persists with exact population distributions and therefore
precedes finite sampling or optimization error.

For example, consider $\dot x=v$, $\dot v=0$ on $\mathbb R^2$, with
$P(x,v)=x$. The full initial laws $\delta_{(0,1)}$ and $\delta_{(0,-1)}$
both give the observed law $\delta_0$.
At time $t>0$, their observed laws are $\delta_t$ and $\delta_{-t}$,
whose $W_1$ distance is $2t$.
At least one prediction from the common current snapshot thus has $W_1$
error at least $t$. Additional measurements or temporal context can provide
information about the missing velocity.

In DisKO, $h_j(\mu)=E_j(P_\#\mu)$ factors through the observation map.
Consequently, every learned observable assigns the same value to full laws
with identical observed distributions.
Predicting the entire future observed law requires the sufficiency
condition above. Predicting a selected set of observables requires only
that those future values be determined by the current representation.
These requirements distinguish closure of selected statistics from
information sufficient for distribution reconstruction.

\paragraph{Identification from finitely sampled times.}
Even with sufficient observations, the sampling times and initial laws
constrain which dynamics can be identified. For instance, the circle
rotation $\theta_t=\theta_0+\omega t\pmod{2\pi}$ induces
$T_t^{(\omega)}\mu=(\theta\mapsto\theta+\omega t)_\#\mu$.
For observations on a uniform grid with spacing $\Delta t$,
\begin{equation}
    T_{n\Delta t}^{(\omega)}
    =T_{n\Delta t}^{(\omega+2\pi k/\Delta t)},
    \qquad n\in\mathbb N_0,\quad k\in\mathbb Z.
    \label{eq:rev-temporal-aliasing}
\end{equation}
Distinct continuous-time frequencies produce identical sampled evolution
for every initial distribution. Recovering a generator therefore requires
additional information, such as a frequency restriction or informative
sampling times. Initial distributions must also excite the observables
whose evolution is to be learned.
Together, observation sufficiency and informative sampling delimit the
identification problem. Convergence and stability of jointly learning the
encoder, operator, and decoder require further analysis within an
identifiable model class.
\par\endgroup

\end{document}